\documentclass[final]{cvpr}

\makeatletter
\@namedef{ver@everyshi.sty}{}
\makeatother
\usepackage{tikz}

\usepackage{times}
\usepackage{epsfig}
\usepackage{graphicx}
\usepackage{amsmath}
\usepackage{amssymb}

\usepackage{placeins}

\usepackage[export]{adjustbox}  
\usepackage{xcolor}
\usepackage{subcaption}

\usepackage{tikz}
\usetikzlibrary{spy}

\usepackage{wrapfig}
\usepackage{multirow}

\newcommand{\fig}[1]{Figure~\ref{fig:#1}}
\newcommand{\sect}[1]{Section~\ref{sect:#1}}
\newcommand{\tab}[1]{Table~\ref{tab:#1}}

\newcommand{\eq}[1]{(\ref{eq:#1})}

\renewcommand{\P}{\mathcal{P}}
\newcommand{\B}{\mathcal{B}}
\newcommand{\hy}{{\hat y}}

\DeclareMathOperator{\pgn}{PGN}
\DeclareMathOperator{\vgg}{VGG}
\DeclareMathOperator{\mse}{MSE}
\DeclareMathOperator{\sg}{sg}

\newcommand\makespy[1]{}

\usetikzlibrary{calc}
\newcommand*{\ExtractCoordinate}[1]{\path (#1); \pgfgetlastxy{\XCoord}{\YCoord}; }
\newcommand*{\ExtractImgDims}[1]{
    \ExtractCoordinate{$(#1.south west)$};
    \pgfmathsetmacro{\imgx}{\XCoord}
    \pgfmathsetmacro{\imgy}{\YCoord}
    \ExtractCoordinate{$(#1.north east)$};
    \pgfmathsetmacro{\imgw}{\XCoord - \imgx}
    \pgfmathsetmacro{\imgh}{\YCoord - \imgy}
}
\newcommand*{\RelativeSpy}[5]{
    \ExtractImgDims{#2};
    \begin{scope}[x=\imgw,y=\imgh,xshift=\imgx,yshift=\imgy]
        \coordinate (spyroi-#1) at #3;
        \coordinate (spypos-#1) at #4;
        \spy[anchor=center,color=#5] on (spyroi-#1) in node[anchor=center] at (spypos-#1);
    \end{scope}
}

\newcommand{\deflen}[2]{%
    \expandafter\newlength\csname #1\endcsname
    \expandafter\setlength\csname #1\endcsname{#2}%
}

\newcommand{\redeflen}[2]{%
    \expandafter\setlength\csname #1\endcsname{#2}%
}

\usepackage{array}
\newcommand{\PreserveBackslash}[1]{\let\temp=\\#1\let\\=\temp}
\newcolumntype{C}[1]{>{\PreserveBackslash\centering}p{#1}}
\newcolumntype{R}[1]{>{\PreserveBackslash\raggedleft}p{#1}}
\newcolumntype{L}[1]{>{\PreserveBackslash\raggedright}p{#1}}

\usepackage[pagebackref=true,breaklinks=true,colorlinks,bookmarks=false]{hyperref}

\def\cvprPaperID{5244} 
\def\confYear{CVPR 2021}

\begin{document}

\title{Perceptual Gradient Networks}

\author{Dmitry Nikulin${}^{1}$\\
\and
Roman Suvorov${}^{1}$\\
\and
Aleksei Ivakhnenko${}^{1}$\\
\and
Victor Lempitsky${}^{1,2}$\\
\and
${}^{1}$Samsung AI Center, Moscow, Russia\\
${}^{2}$Skolkovo Institute of Science and Technology, Moscow, Russia\\
}

\maketitle

\begin{abstract}
Many applications of deep learning for image generation use perceptual losses for either training or fine-tuning of the generator networks. The use of perceptual loss however incurs repeated forward-backward passes in a large image classification network as well as a considerable memory overhead required to store the activations of this network. It is therefore desirable or sometimes even critical to get rid of these overheads.

In this work, we propose a way to train generator networks using approximations of perceptual loss that are computed without forward-backward passes.
Instead, we use a simpler \textit{perceptual gradient network} that directly synthesizes the gradient field of a perceptual loss.
We introduce the concept of proxy targets, which stabilize the predicted gradient, meaning that learning with it does not lead to divergence or oscillations.
In addition, our method allows interpretation of the predicted gradient, providing insight into the internals of perceptual loss and suggesting potential ways to improve it in future work.
\end{abstract}

\section{Introduction}
\label{sect:intro}

Generative neural networks are rapidly developing and acquiring new applications in image processing, image editing, video editing, telepresence. Training generative networks that can process or generate photorealistic images is often heavy on memory and requires lots of time.

The usage scenarios of deep generative networks can be divided into two groups. In the first group of scenarios, a deep generative network is trained on a dedicated hardware such as a multi-GPU server, and then deployed onto more lightweight user devices such as personal computers or mobile phones. In such (train-and-deploy) scenarios, there still is a need to reduce training time and memory consumption throughout the model life cycle.

Even more importantly for our work, there is an emergent group of scenarios, in which deep generative networks need to be fitted to user image at test-time \textit{on user devices}. This group includes image editing systems \cite{Bau19,Zhu16}, a face image manipulation system \cite{Abdal19}, and a neural talking head creation system \cite{Zakharov19}. For all such applications, the ability to speed up network fitting, and even more crucially to reduce memory requirement is needed to run them on user devices. This paper was motivated by the practical need to deploy one of such systems.

Perceptual loss \cite{Johnson16} is one of the most important and widely used losses that assess the quality of images, and all methods mentioned in the previous paragraph rely heavily on it. Computing perceptual loss requires a forward-backward pass through a large convolutional network, typically one of the VGG networks~\cite{Simonyan14}. Using perceptual losses based on much smaller networks leads to inferior results \cite{Dosovitskiy16,Zhang18}. Using forward-backward passes within VGG networks however has a considerable time, memory, and storage costs that might not be feasible for many usage scenarios, and may often exceed the respective costs associated with the generator network itself. 

Perceptual loss measures a certain distance between a pair of images. In the majority of cases, one of such images is produced by the generator network while the other is considered ``ground truth'', so that only the gradient w.r.t.\ the former image is needed to perform fitting/training of the network. We propose to distill the perceptual loss into an image-to-image network (a \textit{perceptual gradient network}) that maps the input pair of images to the gradient of the perceptual loss with respect to the first image. The gradient of the perceptual loss, which is computed through the forward-backward pass in a VGG-network, can thus be approximated by a forward pass in a smaller perceptual gradient network (PGN). Unlike the intermediate activations of the VGG, the intermediate activations in the PGN need not to be stored in memory, greatly reducing the memory cost.


We investigate several training methods and
demonstrate
that PGNs have good generalization ability,
thus enabling the plug-and-play use of perceptual gradient networks. We conduct experimental evaluation of PGNs in several scenarios including inversion of a StyleGAN network and fitting of few-shot neural head avatar models. The code is available at \url{https://github.com/dniku/perceptual-gradient-networks}.

\section{Related work}
\label{sect:related}

Perceptual losses were introduced for general image processing tasks in \cite{Dosovitskiy16,Johnson16,Ulyanov16} based on the image statistics proposed in \cite{Gatys16}. Perceptual losses measure image dissimilarity in the activation space of a deep convolutional network, which allows to better resemble human notion of image similarity than pixel-level distances \cite{Zhang18}. The architecture and the size of the underlying network is very important \cite{Dosovitskiy16,Zhang18} and the VGG architecture~\cite{Simonyan14} has proven to work well and to be hard to replace.

Perceptual losses have been used extensively for such tasks as image processing \cite{Johnson16}, image stylization \cite{Johnson16,Ulyanov16}, and even more ambitious tasks such as image-to-image translation with a wide gap between domains \cite{Chen17}. Other applications include inversion of deep generators \cite{Abdal19} and fine-tuning of pre-trained generators \cite{Zakharov19}.

The idea of replacing a deep network with a smaller surrogate producing the approximation of its results is a popular one. Depending on the way the knowledge transferred from the large network to a small one, this approach is called model compression~\cite{Bucilua06,Ba14}, knowledge distillation \cite{Hinton15} or teacher-student learning \cite{Romero14}. Our work is very related to knowledge distillation and to teacher-student approaches, with an important caveat that traditional student networks are trained to approximate the forward pass in their teachers, while perceptual gradient networks are trained to approximate the results of the forward-backward pass.

Arguably, most related to ours is the idea of \textit{synthetic gradients} suggested in \cite{Jaderberg17,Czarnecki17} for asynchronous parallel training of large networks. The context and the applications of \cite{Jaderberg17,Czarnecki17} are different from ours, as they consider asynchronous training of recurrent and image classification networks, whereas networks predicting synthetic gradients replace parts of the main networks are trained in parallel with them. In contrast, we consider a different application (image processing and generation) and different usage scenarios (learn the loss gradient prediction and then apply them for tasks such as image fitting and generator network fine-tuning). We also consider learning strategies and distillation losses that are specific to our case and therefore not considered in \cite{Jaderberg17,Czarnecki17}.



\section{Method}
\label{sect:method}

\paragraph{General framework.} We denote perceptual loss as $\vgg(\hy, y)$, where $y$ is the "ground truth" image, while $\hat y$ is the "generated" image. Specific meaning may vary by context, but usually $y$ is supposed to be fixed, while $\hat y$ changes under optimization. In this notation, our goal is to construct a model $\pgn(\hy, y; \theta)$ such that
\begin{equation}
    \pgn(\hy, y; \theta) \approx \nabla_\hy \vgg(\hy, y) \,,
\end{equation}
where $\theta$ denotes the learnable parameters of $\pgn$.

\begin{figure*}
    \centering
    \includegraphics[width=\linewidth]{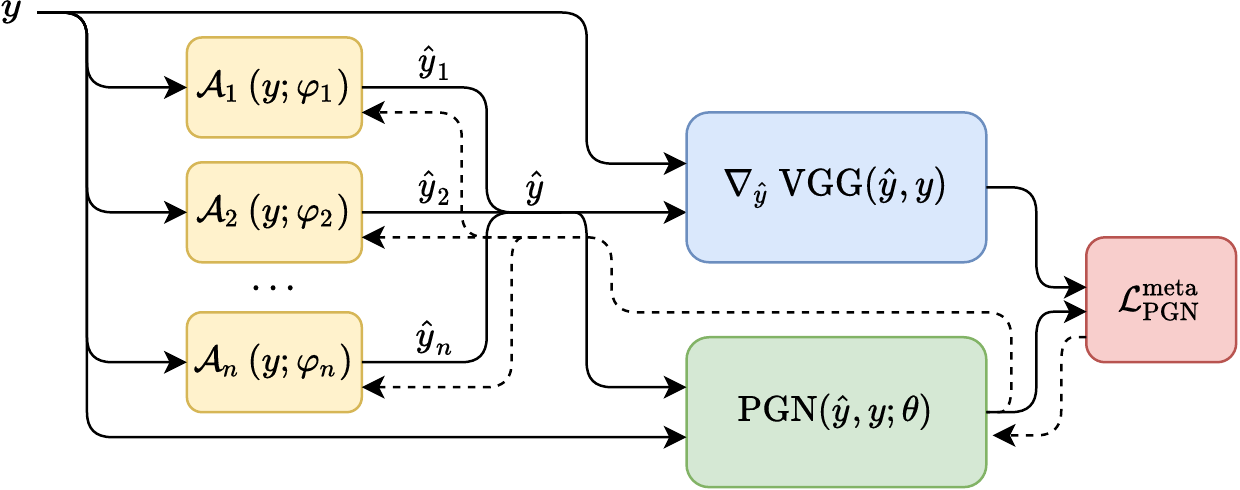}
    \caption{An overview of our training pipeline. Real images $y$ first pass through each of autoencoders $\left\{ \mathcal{A}_i \right\}_{i=1}^n$, producing images $\hy_i$, which are concatenated in the batch dimension, resulting in $\hy$. Then we compute the real gradient of the perceptual loss $\nabla_\hy \vgg \left( \hy, y \right)$ and our predicted gradient $\pgn \left( \hy, y \right)$. We use this predicted gradient to update autoencoder parameters $\varphi_i$. Finally, we compute the (meta-)loss $\mathcal{L}_{\mathrm{PGN}}^{\mathrm{meta}}$ and update $\pgn$ parameters $\theta$. We reset autoencoder parameters whenever their reconstruction quality plateaus for a fixed number of batches. A detailed description of autoencoder architecture is available in supplementary material (\fig{architectures}).}
    \label{fig:pgn-training-pipeline}
\end{figure*}

To train a $\pgn$, we employ a setup depicted in~\fig{pgn-training-pipeline}. We use a family of autoencoders $\left\{ \mathcal{A}_i \right\}_{i=1}^n$ as surrogate image generators, which take real images $y$ and produce images $\hy$. Using these images, we compute the real gradient of the perceptual loss $\nabla_{\hy} \vgg \left( \hy, y \right)$ and our synthetic gradient $\pgn \left( \hy, y; \theta \right)$. Similarly to \cite{Jaderberg17,Czarnecki17}, we update autoencoder parameters using our synthetic gradient (by performing gradient descent with Adam~\cite{Kingma15} optimizer), and then update $\pgn$ parameters using a (meta-)loss $\mathcal{L}_{\mathrm{PGN}}^{\mathrm{meta}}$, also using Adam~\cite{Kingma15}. We describe the exact form of $\mathcal{L}_{\mathrm{PGN}}^{\mathrm{meta}}$ below.

\paragraph{PGN structure.}
Let $\B$ be a neural network with parameters $\theta$ to serve as the \textit{backbone} of $\pgn$. In general, $\B$ can have any image-to-image architecture, e.g.\ a ResNet followed by upsampling \cite{Johnson16} or a UNet~\cite{Ronneberger15}, and it should take a six-channel image (two RGB images concatenated along the channel dimension) as input and produce a three-channel image. We explore multiple variations on how we can synthesize gradient approximation from the output of $\B$ and show that the exact way to do that has a large impact on the quality and stability of the resulting gradient field. In the remainder of this section, the letters $\alpha$, $\beta$, $\gamma$ denote fixed hyperparameters.

\textbf{Backbone predicts gradient directly.} In this setup, the output of the backbone is used to approximate the gradient with a scaling factor $\alpha$:
\begin{equation}
    \pgn(\hy, y; \theta) = \alpha \cdot \B(\hy, y; \theta).
\end{equation}
The scaling factor is used for matching the scale of the output of $\B$ to the scale of the real gradient and allows us to avoid tuning the initialization scheme of $\B$. We found that the models are quite robust to changes in this parameter, as long as the scale of the output of a randomly initialized PGN model is within roughly one order of magnitude from the scale of the real gradient. In our experiments, $\alpha = 1 / 512$.

We found that while models with this structure attain reasonably high performance during training, they have poor generalization. The vector field they output has erratic behavior, and trajectories in this field may diverge or otherwise fluctuate. We conduct experimental evaluation of this in \sect{experiments}.

\textbf{Backbone predicts a proxy target}. Here, rather than producing gradient approximation directly, $\B$ outputs a \textit{proxy target}, which is used as a target for optimization with mean-squared error (MSE) loss:
\begin{align}
    \P(\hy, y; \theta) &= \B(\hy, y; \theta) \\
    \pgn(\hy, y; \theta) &= \alpha \cdot \nabla_{\hy} \mse \left( \hy, \, \sg \left[ \P(\hy, y; \theta) \right] \right) = \\
        &= \alpha \frac{2}{C \times H \times W} \left( \hy - \P(\hy, y; \theta) \right),
\end{align}
where $\sg$ (``stop gradients'') indicates that the output of the backbone is regarded as constant in the computation of the MSE gradient. $C$, $H$, and $W$ are the image tensor dimensions. We used $\alpha = 40$ in our experiments.

Not surprisingly, similarly to the previous model, we found that this approach leads to an unstable vector field (see \sect{experiments}); however, it brings us to the next idea.

\textbf{Backbone predicts a constrained proxy target.} This is very similar to the previous approach, but with two small changes: we apply a constraining function $f$ to the proxy target, and we multiply its output by $\beta$.
\begin{align}
    \P(\hy, y; \theta) &= \beta \cdot f \left( \B(\hy, y; \theta) \right) \\
    \pgn(\hy, y; \theta) &= \alpha \cdot \nabla_{\hy} \mse \left( \hy, \, \sg \left[ \P(\hy, y; \theta) \right] \right),
\end{align}
In our experiments, $f$ is the sigmoid function followed by ImageNet normalization~\cite{Simonyan14} (scaling pixel values by a fixed mean and variance).
We used $\alpha = 40$ and $\beta = 1.1$.

With this change, the proxy target values are constrained, and the optimization procedure cannot diverge to infinity. We also noticed that if we keep $\beta = 1$, images obtained through optimization in this field tend to have muted colors, and would struggle to reproduce pure white or pure black color. To work around this problem, we allowed proxy targets to have pixel value range expanded by a factor $\beta$, thus allowing non-zero $\mse$ gradient when the input images are close to color range boundaries.

We study this setup in detail in \sect{experiments}. However, we briefly mention one more setup, which is a hybrid of everything introduced so far.

\textbf{Backbone predicts gradient, which is turned into a constrained proxy target.} Here, we try to find the middle ground of the previous ideas. $\B$ predicts the gradient, which is turned into the corresponding proxy target, to which we apply the constraint and turn the result back into the gradient:
\begin{align}
    \P(\hy, y; \theta) &= \beta \cdot f \left( \hy - \gamma \cdot \B(\hy, y; \theta) \right) \\
    \pgn(\hy, y; \theta) &= \alpha \cdot \nabla_{\hy} \mse \left( \hy, \, \sg \left[ \P(\hy, y; \theta) \right] \right),
\end{align}
We used $\alpha = 40$, $\beta = 1.1$, $\gamma = 1 / 512$.
We verified experimentally that this approach also leads to a stable vector field, confirming that stability is due to the constraint on the proxy, and does not depend much on $\B$ predicting the proxy instead of the gradient. However, since it requires tuning more hyperparameters than the previous method, we do not study it in detail.

\paragraph{Meta-loss.} Having introduced the concept of the proxy $\P(\hy, y; \theta)$, we are finally ready to write our (meta-)loss used to update $\theta$:
\begin{multline}\label{eq:meta-loss}
    \mathcal{L}_{\mathrm{PGN}}^{\mathrm{meta}}(\hy, y; \theta) = \\
    \lambda_{\mathrm{grad}} \|\pgn(\hy, y; \theta) - \nabla_\hy \vgg(\hy, y)\|_2^2 + \\
    \lambda_{\mathrm{VGG}} \vgg \left( \P(\hy, y; \theta), y \right) + \\
    \lambda_{\mathrm{L_1}} \|\P(\hy, y; \theta) - y\|_1
    \,.
\end{multline}
We used $\lambda_{\mathrm{grad}} = 1$, $\lambda_{\mathrm{VGG}} = 0.5$, $\lambda_{\mathrm{L_1}} = 0.2$. The second and third terms help to preserve colors in the images obtained from optimization via synthetic gradients.

\begin{figure*}[t]
    \centering
    \includegraphics[width=\textwidth]{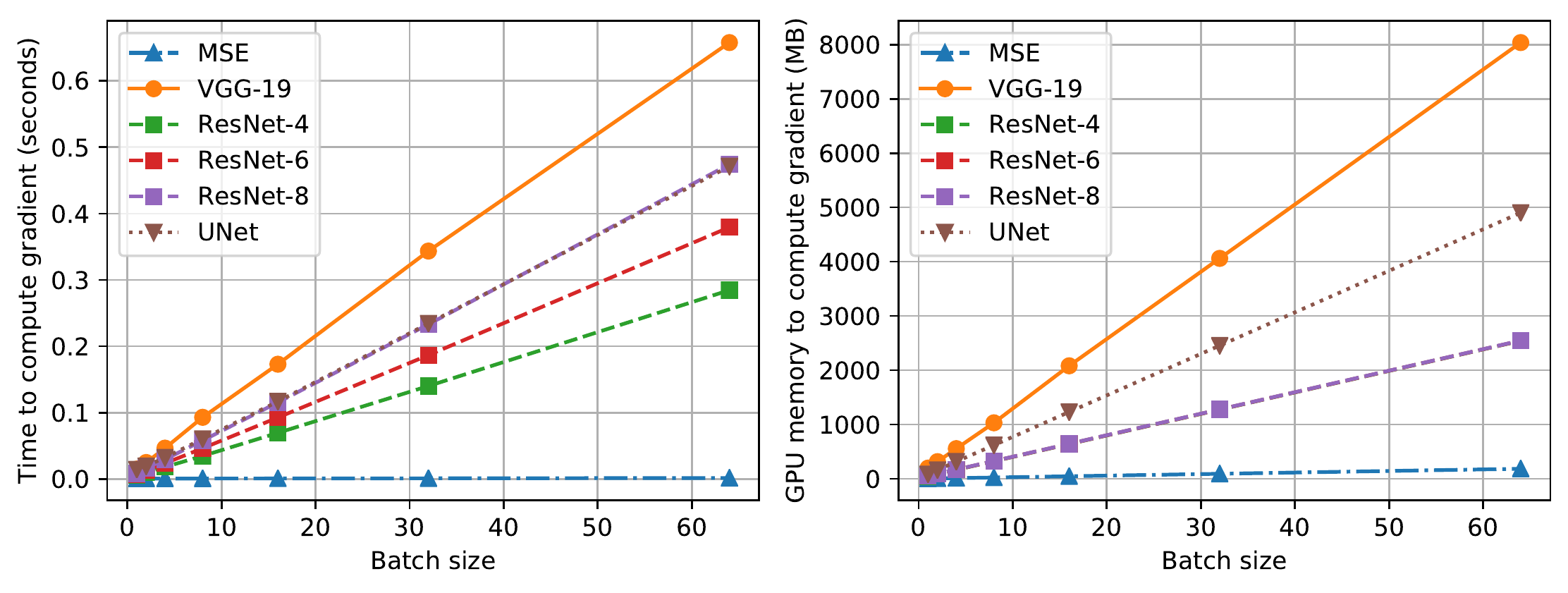}
    \caption{Time and memory comparison on an Nvidia GeForce 1080 Ti. Note that (1) despite having a drastically lower number of multiply-add operations, our UNet architecture runs in approximately the same time as the ResNet architecture with 8 blocks, and (2) all ResNet-based models use approximately the same amount of memory, since they differ only in the number of residual blocks.}
    \label{fig:performance}
\end{figure*}

\section{Experiments}
\label{sect:experiments}

    
    

\begin{table}[b]
    \centering
    \begin{tabular}{|l|ll|}
        \hline
        Model    & Params & MACs  \\ \hline
        VGG-19   & 12.94M         & 108.84B* \\
        ResNet-4 & \textbf{1.34M} & 19.79B   \\
        ResNet-6 & 1.93M          & 27.23B   \\
        ResNet-8 & 2.52M          & 34.66B   \\
        UNet     & 1.77M          & \textbf{2.10B}   \\ \hline
    \end{tabular}
    \caption[]{Size comparison. For VGG-19, we approximate the number of multiply-accumulate operations (MACs, \texttt{thop}\footnote{\url{https://github.com/Lyken17/pytorch-OpCounter}}) for gradient computation with the number of MACs for a forward pass multiplied by 3 (2 forward and 1 backward).
    }
    \label{tab:performance}
\end{table}

\paragraph{Implementation details.} We borrow specifics of the perceptual loss from the pipeline of \cite{Zakharov19} by taking a pretrained VGG-19 \cite{Simonyan14} model and replacing all of its \texttt{MaxPool} layers with \texttt{AvgPool}. Then, for both $\hat y$ and $y$, we compute activations after each ReLU for all layers up to and including \texttt{conv5\_1}. The perceptual loss is defined as the sum of $L_1$ distances between activations of corresponding layers.

For the backbone $\B$ of our perceptual gradient network, we experiment with several ResNet-like architectures followed by upsampling~\cite{Johnson16} with 4, 6, or 8 residual blocks, as well as with a UNet-like architecture~\cite{Ronneberger15} with 5 downscales and upscales. For autoencoders, we use an architecture derived from the UNet, but without skip connections and smaller. A more detailed description is available in supplemetary material (\fig{architectures}).


Since the VGG-19 model used for the perceptual loss is trained on the full ILSVRC2012 ImageNet dataset~\cite{ILSVRC15}, we also use it for training $\pgn$. Our augmentations pipeline is identical to the one that was used for training the TorchVision\footnote{\url{https://github.com/pytorch/vision}} VGG-19 model whose gradients we approximate. It consists of a random crop, resizing to $224 \times 224$, a random horizontal flip, and, finally, scaling pixel values by a fixed mean and variance. All of these are standard for training image classification networks~\cite{Simonyan14}.

\footnotetext{\url{https://github.com/Lyken17/pytorch-OpCounter}}

All of PGNs were trained with the (meta-)loss~\eq{meta-loss} using Adam~\cite{Kingma15} (meta-)optimizer in the autoencoder setup detailed in \fig{pgn-training-pipeline}. Each network was trained on a single server with 8 Nvidia P40 GPUs. PGN parameters were synchronized across all GPUs, and we spawned $n = 3$ autoencoders independently on each GPU, resulting in a total of 24 autoencoder instances.

In \tab{performance} and \fig{performance}, we compare computational performance characteristics of our models. All of our models have a smaller number of parameters, and thus smaller checkpoint size, than the VGG-19 model whose gradients they learned to imitate, as well as requiring less memory for inference.

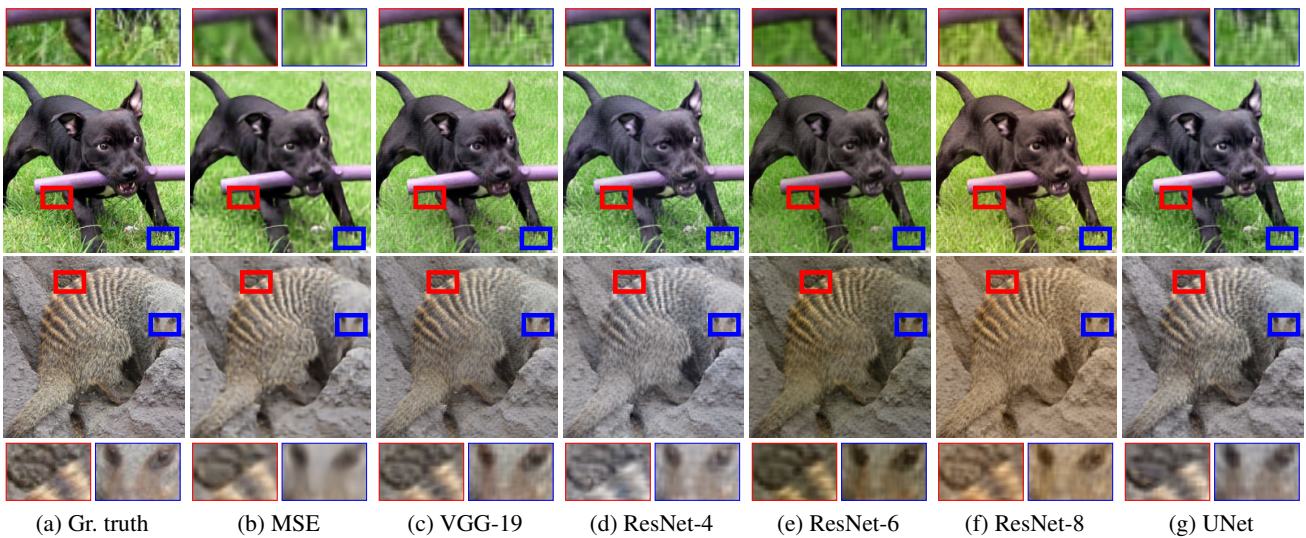
\begin{figure*}
    \centering
    \deflen{twolensplash}{0.142\linewidth}
    \renewcommand\makespy[1]{%
        \begin{tikzpicture}[spy using outlines={rectangle, magnification=3, height=2.5cm, width=3.7cm, every spy on node/.append style={line width=2mm}}]
                \node (nd1){\includegraphics{#1}};
                \RelativeSpy{nd1-spy1}{nd1}{(0.3,0.31)}{(0.255,1.17)}{red}
                \RelativeSpy{nd1-spy2}{nd1}{(0.87,0.09)}{(0.735,1.17)}{blue}
        \end{tikzpicture}%
    }
    \begin{subfigure}[b]{\twolensplash}
        \resizebox{1.02\textwidth}{!}{
            \makespy{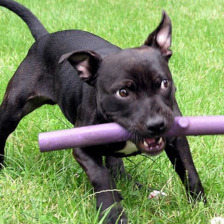}
        }
    \end{subfigure}%
    \begin{subfigure}[b]{\twolensplash}
        \resizebox{1.02\textwidth}{!}{
            \makespy{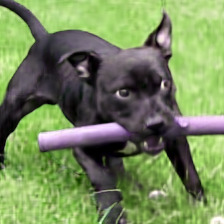}
        }
    \end{subfigure}%
    \begin{subfigure}[b]{\twolensplash}
        \resizebox{1.02\textwidth}{!}{
            \makespy{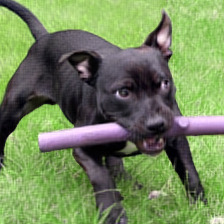}
        }
    \end{subfigure}%
    \begin{subfigure}[b]{\twolensplash}
        \resizebox{1.02\textwidth}{!}{
            \makespy{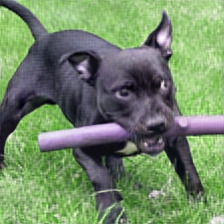}
        }
    \end{subfigure}%
    \begin{subfigure}[b]{\twolensplash}
        \resizebox{1.02\textwidth}{!}{
            \makespy{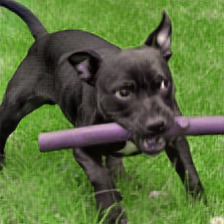}
        }
    \end{subfigure}%
    \begin{subfigure}[b]{\twolensplash}
        \resizebox{1.02\textwidth}{!}{
            \makespy{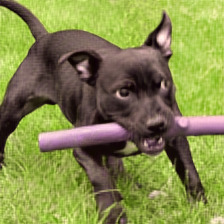}
        }
    \end{subfigure}%
    \begin{subfigure}[b]{\twolensplash}
        \resizebox{1.02\textwidth}{!}{
            \makespy{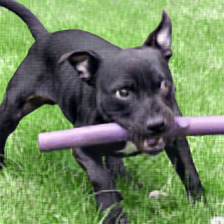}
        }
    \end{subfigure}%
    \\
    \renewcommand\makespy[1]{%
        \begin{tikzpicture}[spy using outlines={rectangle, magnification=3, height=2.5cm, width=3.7cm, every spy on node/.append style={line width=2mm}}]
                \node (nd1){\includegraphics{#1}};
                \RelativeSpy{nd1-spy1}{nd1}{(0.37,0.85)}{(0.255,-0.17)}{red}
                \RelativeSpy{nd1-spy2}{nd1}{(0.88,0.62)}{(0.735,-0.17)}{blue}
        \end{tikzpicture}%
    }
    \vspace{-0.5mm}
    \begin{subfigure}[b]{\twolensplash}
        \resizebox{1.02\textwidth}{!}{
            \makespy{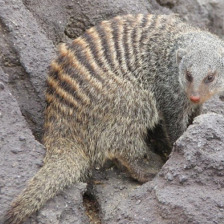}
        }
        \vspace*{-5mm}\caption{Gr.\ truth}
    \end{subfigure}%
    \begin{subfigure}[b]{\twolensplash}
        \resizebox{1.02\textwidth}{!}{
            \makespy{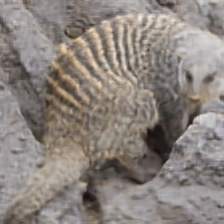}
        }
        \vspace*{-5mm}\caption{MSE}
    \end{subfigure}%
    \begin{subfigure}[b]{\twolensplash}
        \resizebox{1.02\textwidth}{!}{
            \makespy{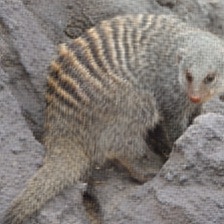}
        }
        \vspace*{-5mm}\caption{VGG-19}
    \end{subfigure}%
    \begin{subfigure}[b]{\twolensplash}
        \resizebox{1.02\textwidth}{!}{
            \makespy{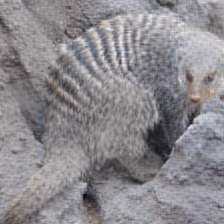}
        }
        \vspace*{-5mm}\caption{ResNet-4}
    \end{subfigure}%
    \begin{subfigure}[b]{\twolensplash}
        \resizebox{1.02\textwidth}{!}{
            \makespy{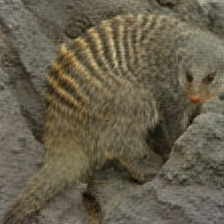}
        }
        \vspace*{-5mm}\caption{ResNet-6}
    \end{subfigure}%
    \begin{subfigure}[b]{\twolensplash}
        \resizebox{1.02\textwidth}{!}{
            \makespy{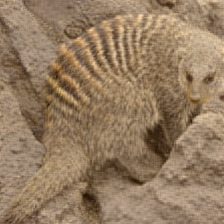}
        }
        \vspace*{-5mm}\caption{ResNet-8}
    \end{subfigure}%
    \begin{subfigure}[b]{\twolensplash}
        \resizebox{1.02\textwidth}{!}{
            \makespy{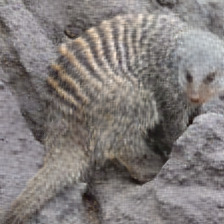}
        }
        \vspace*{-5mm}\caption{UNet}
    \end{subfigure}
    \caption{Examples of reconstruction of images from the ImageNet validation set using Deep Image Prior with different losses. Note how PGN improves fine details in grass and fur, compared to MSE.}
    \label{fig:dip-comparison}
\end{figure*}
 
{\setlength{\tabcolsep}{0.5em}
\begin{table*}[t]
    \centering
    \begin{tabular}{|l|lll|lll|}
        \hline
                    & \multicolumn{3}{c|}{Deep Image Prior} & \multicolumn{3}{c|}{Image2StyleGAN} \\
        \cline{2-7}
                    & MSE $_{\times 10^3} \downarrow$   & PL$_\textrm{VGG-19} \downarrow$ & LPIPS $\downarrow$
                    & MSE $_{\times 10^3} \downarrow$   & PL$_\textrm{VGG-19} \downarrow$ & LPIPS  $\downarrow$           \\ \hline
        MSE         & \textbf{1.43} \tiny{$\pm$ 1.06}   & 3.31 \tiny{$\pm$ 0.69}            & 0.25 \tiny{$\pm$ 0.07}          
                    & \textbf{0.83} \tiny{$\pm$ 0.48}   & 3.57 \tiny{$\pm$ 0.55}            & 0.26 \tiny{$\pm$ 0.04}            \\
        PL$_\textrm{VGG-19}$ & 2.57 \tiny{$\pm$ 2.04}            & \textbf{1.77} \tiny{$\pm$ 0.53}   & \textbf{0.11} \tiny{$\pm$ 0.04}
                    & 1.54 \tiny{$\pm$ 0.8}             & \textbf{2.72} \tiny{$\pm$ 0.47}   & \textbf{0.18} \tiny{$\pm$ 0.03}   \\
        \hline
        ResNet-4    & 9.13 \tiny{$\pm$ 21.2}            & \textbf{2.46} \tiny{$\pm$ 0.64}   & \textbf{0.16} \tiny{$\pm$ 0.06} 
                    & \textbf{1.41} \tiny{$\pm$ 0.71}   & 3.07 \tiny{$\pm$ 0.55}            & 0.22 \tiny{$\pm$ 0.04}            \\
        ResNet-6    & 60.3 \tiny{$\pm$ 92.4}            & 2.84 \tiny{$\pm$ 0.72}            & 0.21 \tiny{$\pm$ 0.09}          
                    & 1.46 \tiny{$\pm$ 0.73}            & 3.08 \tiny{$\pm$ 0.56}            & \textbf{0.22} \tiny{$\pm$ 0.05}   \\
        ResNet-8    & 39.6 \tiny{$\pm$ 84.1}            & 2.63 \tiny{$\pm$ 0.66}            & 0.18 \tiny{$\pm$ 0.07}          
                    & 1.46 \tiny{$\pm$ 0.71}            & \textbf{3.06} \tiny{$\pm$ 0.55}   & \textbf{0.22} \tiny{$\pm$ 0.04}   \\
        UNet        & \textbf{4.81} \tiny{$\pm$ 3.9}    & 2.48 \tiny{$\pm$ 0.58}            & 0.17 \tiny{$\pm$ 0.05}          
                    & 1.50 \tiny{$\pm$ 0.64}            & 3.19 \tiny{$\pm$ 0.54}            & 0.24 \tiny{$\pm$ 0.04}            \\ \hline
    \end{tabular}
    \caption{Quantitative evaluation of reconstruction in Deep Image Prior and Image2StyleGAN.}
    \label{tab:dip-stylegan}
\end{table*}
}

\paragraph{Deep Image Prior.} To ensure that our models learn a reasonable gradient approximation, we perform a simple experiment, training a Deep Image Prior~\cite{ulyanov2017deep} network with our synthetic gradients. We modify a publicly available implementation of the Deep Image Prior pipeline\footnote{\url{https://github.com/DmitryUlyanov/deep-image-prior}} (the ``JPEG artifacts removal'' variant) by integrating our models there. First, we compare stability of optimization with different PGN setups. We sample $N = 100$ images from the ImageNet validation set (unseen by both VGG and PGN networks during training) and use PGNs to train Deep Image Prior networks to generate these images ($T = 10000$ iterations, Adam, $lr=0.01$). At each iteration, we measure the true VGG-19 perceptual loss, and we terminate training prematurely either if it becomes twice as large as its initial value (we qualify this as divergence), or if it does not improve for $2000$ iterations (stagnation).

We compared (1) a PGN that predicts the gradient directly, (2) a PGN that predicts the gradient via an unconstrained proxy target, and (3) multiple PGNs that apply a constraint to the proxy target (with different backbones). The first two PGNs used ResNet-8 as the backbone. The first variant diverged 10 times out of 100 and stagnated twice; the second one diverged 3 times but stagnated 12 times. PGNs that used a constrained proxy never diverged and stagnated 1 or 2 times out of 100; see supplementary material (\tab{stability}) for more detailed information.

We emphasize the fact that we performed optimization for up to $T = 10000$ steps using synthetic gradient only, and the process never diverged.
In fact, we have \textit{never} observed divergence in our experiments with PGN variants that employ a constraint on the proxy.
This is remarkable since our synthetic gradient is not guaranteed to be the gradient of any function (i.e.\ the corresponding vector field need not be conservative), and therefore convergence of optimization is not guaranteed even with small learning rates. This problem is discussed in more detail in Proposition 4 in \cite{NIPS2017_7015}, to which we refer the inquisitive reader.

\begin{figure*}
    \centering
    \redeflen{twolensplash}{0.142\linewidth}
    \renewcommand\makespy[1]{%
        \begin{tikzpicture}[spy using outlines={rectangle, magnification=3, height=3cm, width=4.2cm, every spy on node/.append style={line width=2mm}}]
                \node (nd1){\includegraphics{#1}};
                \RelativeSpy{nd1-spy1}{nd1}{(0.2,0.28)}{(0.255,1.17)}{red}
                \RelativeSpy{nd1-spy2}{nd1}{(0.55,0.75)}{(0.735,1.17)}{blue}
        \end{tikzpicture}%
    }
    \begin{subfigure}[b]{\twolensplash}
        \resizebox{1.02\textwidth}{!}{
            \makespy{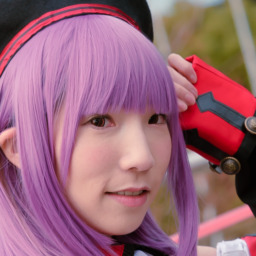}
        }
    \end{subfigure}%
    \begin{subfigure}[b]{\twolensplash}
        \resizebox{1.02\textwidth}{!}{
            \makespy{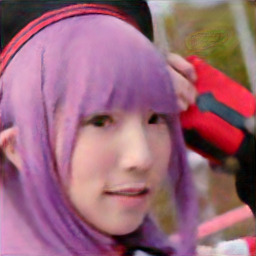}
        }
    \end{subfigure}%
    \begin{subfigure}[b]{\twolensplash}
        \resizebox{1.02\textwidth}{!}{
            \makespy{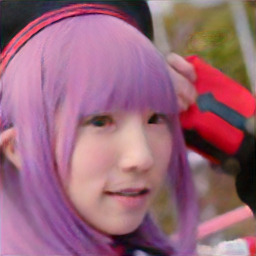}
        }
    \end{subfigure}%
    \begin{subfigure}[b]{\twolensplash}
        \resizebox{1.02\textwidth}{!}{
            \makespy{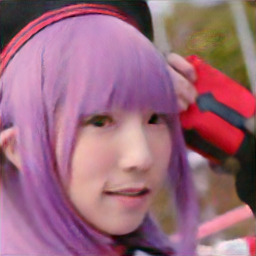}
        }
    \end{subfigure}%
    \begin{subfigure}[b]{\twolensplash}
        \resizebox{1.02\textwidth}{!}{
            \makespy{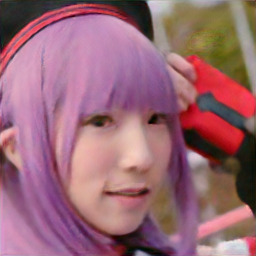}
        }
    \end{subfigure}%
    \begin{subfigure}[b]{\twolensplash}
        \resizebox{1.02\textwidth}{!}{
            \makespy{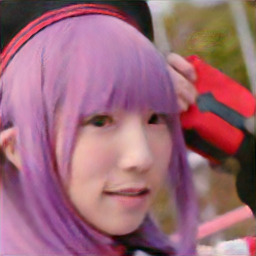}
        }
    \end{subfigure}%
    \begin{subfigure}[b]{\twolensplash}
        \resizebox{1.02\textwidth}{!}{
            \makespy{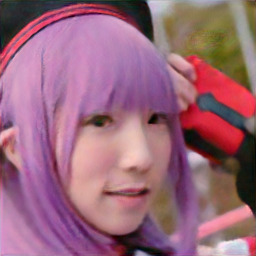}
        }
    \end{subfigure}%
    \\
    \renewcommand\makespy[1]{%
        \begin{tikzpicture}[spy using outlines={rectangle, magnification=3, height=3cm, width=4.2cm, every spy on node/.append style={line width=2mm}}]
                \node (nd1){\includegraphics{#1}};
                \RelativeSpy{nd1-spy1}{nd1}{(0.21,0.6)}{(0.255,-0.17)}{red}
                \RelativeSpy{nd1-spy2}{nd1}{(0.71,0.68)}{(0.735,-0.17)}{blue}
        \end{tikzpicture}%
    }
    \vspace{-0.5mm}
    \begin{subfigure}[b]{\twolensplash}
        \resizebox{1.02\textwidth}{!}{
            \makespy{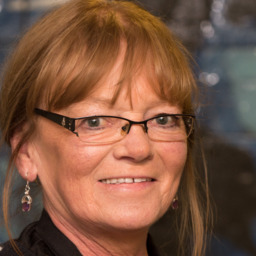}
        }
        \vspace*{-5mm}\caption{Gr.\ truth}
    \end{subfigure}%
    \begin{subfigure}[b]{\twolensplash}
        \resizebox{1.02\textwidth}{!}{
            \makespy{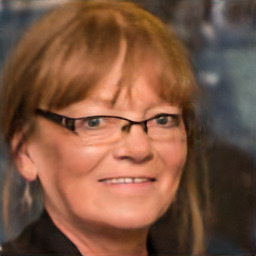}
        }
        \vspace*{-5mm}\caption{MSE}
    \end{subfigure}%
    \begin{subfigure}[b]{\twolensplash}
        \resizebox{1.02\textwidth}{!}{
            \makespy{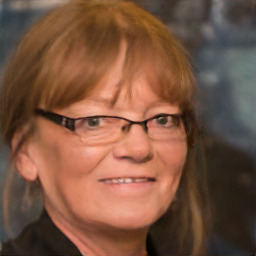}
        }
        \vspace*{-5mm}\caption{VGG-19}
    \end{subfigure}%
    \begin{subfigure}[b]{\twolensplash}
        \resizebox{1.02\textwidth}{!}{
            \makespy{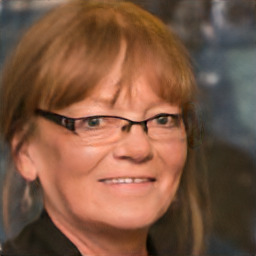}
        }
        \vspace*{-5mm}\caption{ResNet-4}
    \end{subfigure}%
    \begin{subfigure}[b]{\twolensplash}
        \resizebox{1.02\textwidth}{!}{
            \makespy{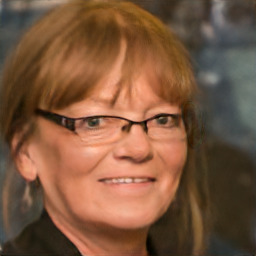}
        }
        \vspace*{-5mm}\caption{ResNet-6}
    \end{subfigure}%
    \begin{subfigure}[b]{\twolensplash}
        \resizebox{1.02\textwidth}{!}{
            \makespy{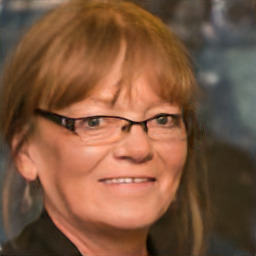}
        }
        \vspace*{-5mm}\caption{ResNet-8}
    \end{subfigure}%
    \begin{subfigure}[b]{\twolensplash}
        \resizebox{1.02\textwidth}{!}{
            \makespy{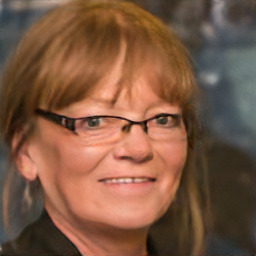}
        }
        \vspace*{-5mm}\caption{UNet}
    \end{subfigure}
    \caption{Examples of reconstructed images from the FFHQ validation set using Image2StyleGAN with different losses. Note how more sharp hair is with PGN than with MSE.}
    \label{fig:stylegan-comparison}
\end{figure*}

\begin{figure*}[t]
    \begin{subfigure}{0.39\textwidth}
        \includegraphics[width=\linewidth]{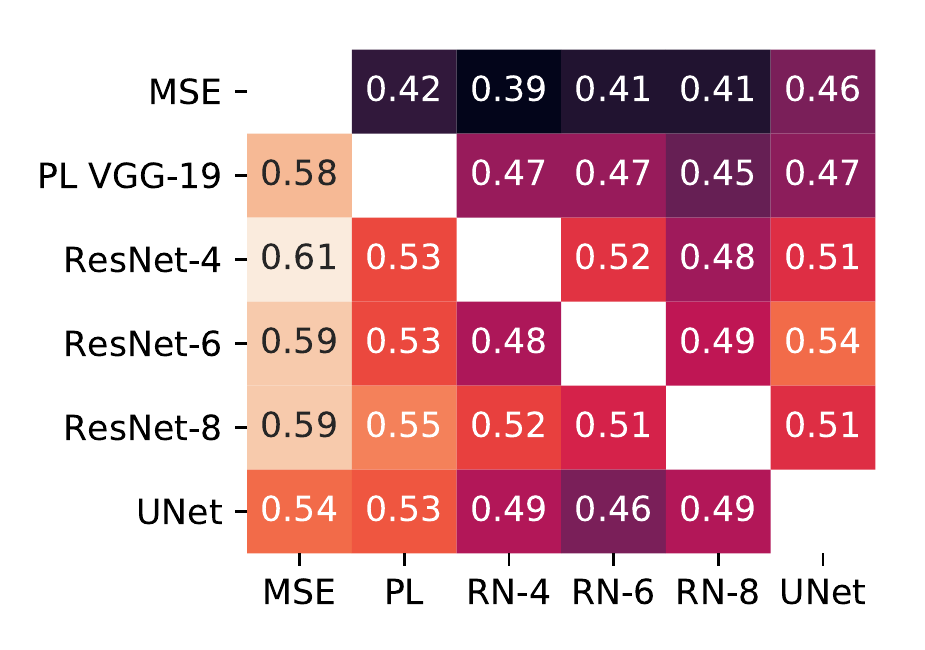}
        \vspace*{-1.8em}\caption{Image2StyleGAN}
        \label{fig:stylegan-toloka}
    \end{subfigure}%
    \begin{subfigure}{0.305\textwidth}
        \includegraphics[width=\linewidth]{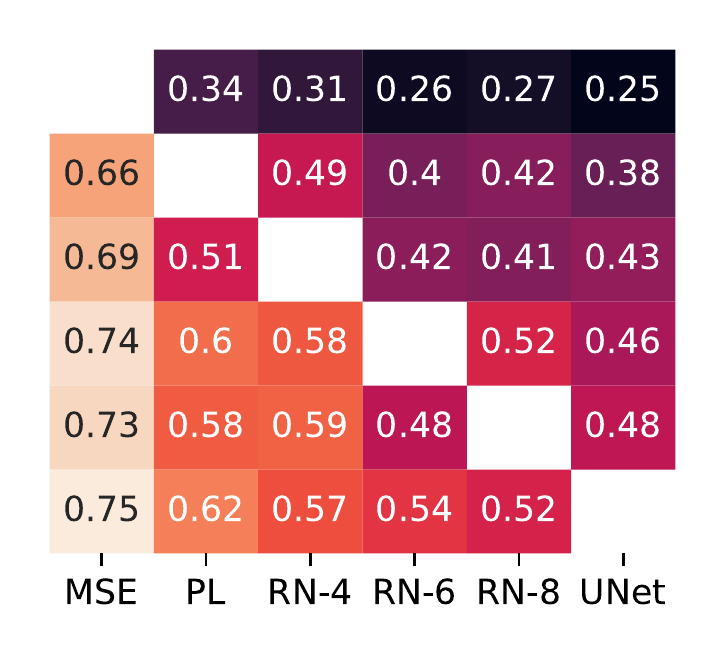}
        \vspace*{-1.8em}\caption{Neural Talking Heads, train}
        \label{fig:avatars-toloka-train}
    \end{subfigure}%
    \begin{subfigure}{0.305\textwidth}
        \includegraphics[width=\linewidth]{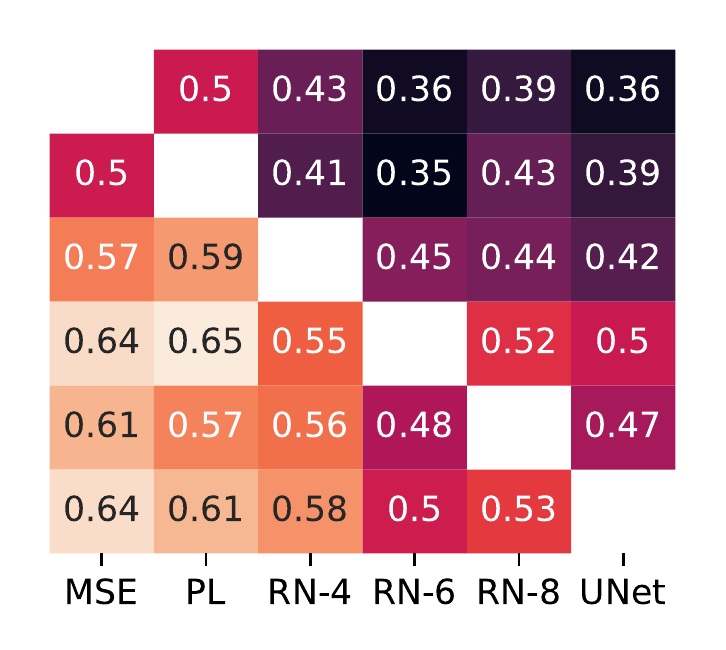}
        \vspace*{-1.8em}\caption{Neural Talking Heads, holdout}
        \label{fig:avatars-toloka-test}
    \end{subfigure}
    \caption{Fraction of row-over-column wins in side-by-side user study for Image2StyleGAN (a) and talking head models (b, c). For talking heads, models were fine-tuned on 8 ``train'' images and then evaluated on 8 more ``holdout'' images. Assessors were shown pairs of images and asked to identify the one with more fine detail. Each pair was labeled by 3 independent assessors.
    }
    \label{fig:toloka}
\end{figure*}

We proceed to comparing the quality of models that make use of a constrained proxy target. Here, we sample $N = 400$ images and train Deep Image Prior networks for $T = 3000$ iterations using either MSE loss, perceptual loss based on VGG-19, or our PGN models, leaving the rest of parameters the same. We measure the quality of the final images using MSE, perceptual loss, and LPIPS~\cite{zhang2018unreasonable}.
The results are shown in \tab{dip-stylegan}, with illustrations in \fig{dip-comparison}. From the metrics it is clear that our PGN models optimize perceptual loss better than MSE does, thus confirming that synthetic gradient can be used as a substitute for the real perceptual loss gradient, resulting in time and memory savings without significant drop in image quality.

We however note that MSE ends up being substantially higher when optimizing with synthetic gradient only, which manifests in color shift artifacts. In our experiments, these artifacts are especially pronounced with ResNet-6 and ResNet-8 models. This (and specifics of the next experiment) motivates us to use synthetic gradient alongside the MSE loss in the next section.



\paragraph{StyleGAN inference.} We now consider a more practical task of StyleGAN inference, which has multiple applications concerned with semantic image editing~\cite{Bau19,Abdal19}. We take an off-the-shelf implementation\footnote{\url{https://github.com/rosinality/style-based-gan-pytorch}} of StyleGAN~\cite{Karras19} pretrained on the FFHQ~\cite{Karras19} dataset and reimplement the Image2StyleGAN~\cite{Abdal19} algorithm, which essentially finds an inverse mapping from an image to the (extended) StyleGAN latent space. Our goal is to verify that PGN works well in novel domains and yields better reconstruction quality than inference with MSE loss only.


\begin{figure*}[t]
    \centering
    \redeflen{twolensplash}{0.142\linewidth}
    \renewcommand\makespy[1]{%
        \begin{tikzpicture}[spy using outlines={rectangle, magnification=3, height=3cm, width=4.2cm, every spy on node/.append style={line width=2mm}}]
                \node (nd1){\includegraphics{#1}};
                \RelativeSpy{nd1-spy1}{nd1}{(0.15,0.75)}{(0.255,1.17)}{red}
                \RelativeSpy{nd1-spy2}{nd1}{(0.8,0.3)}{(0.735,1.17)}{blue}
        \end{tikzpicture}%
    }
    \begin{subfigure}[b]{\twolensplash}
        \resizebox{1.02\textwidth}{!}{
            \makespy{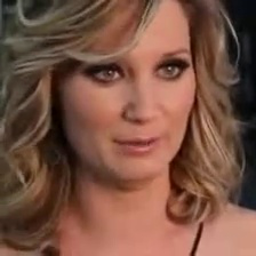}
        }
    \end{subfigure}%
    \begin{subfigure}[b]{\twolensplash}
        \resizebox{1.02\textwidth}{!}{
            \makespy{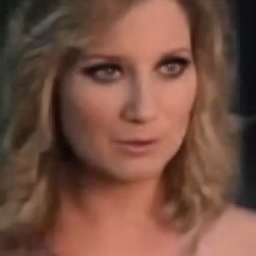}
        }
    \end{subfigure}%
    \begin{subfigure}[b]{\twolensplash}
        \resizebox{1.02\textwidth}{!}{
            \makespy{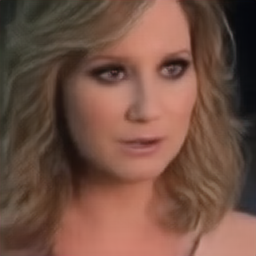}
        }
    \end{subfigure}%
    \begin{subfigure}[b]{\twolensplash}
        \resizebox{1.02\textwidth}{!}{
            \makespy{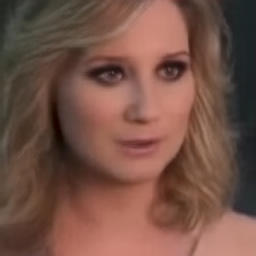}
        }
    \end{subfigure}%
    \begin{subfigure}[b]{\twolensplash}
        \resizebox{1.02\textwidth}{!}{
            \makespy{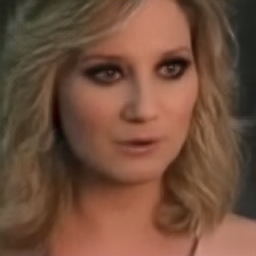}
        }
    \end{subfigure}%
    \begin{subfigure}[b]{\twolensplash}
        \resizebox{1.02\textwidth}{!}{
            \makespy{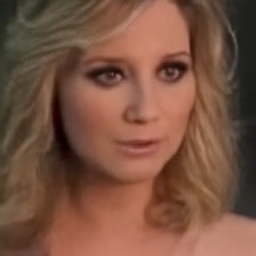}
        }
    \end{subfigure}%
    \begin{subfigure}[b]{\twolensplash}
        \resizebox{1.02\textwidth}{!}{
            \makespy{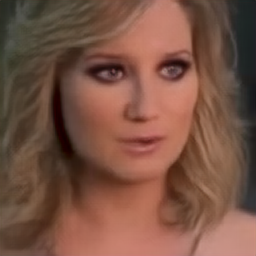}
        }
    \end{subfigure}%
    \\
    \renewcommand\makespy[1]{%
        \begin{tikzpicture}[spy using outlines={rectangle, magnification=3, height=3cm, width=4.2cm, every spy on node/.append style={line width=2mm}}]
                \node (nd1){\includegraphics{#1}};
                \RelativeSpy{nd1-spy1}{nd1}{(0.21,0.55)}{(0.255,-0.17)}{red}
                \RelativeSpy{nd1-spy2}{nd1}{(0.48,0.86)}{(0.735,-0.17)}{blue}
        \end{tikzpicture}%
    }
    \vspace{-0.5mm}
    \begin{subfigure}[b]{\twolensplash}
        \resizebox{1.02\textwidth}{!}{
            \makespy{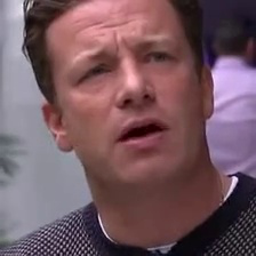}
        }
        \vspace*{-5mm}\caption{Gr.\ truth}
    \end{subfigure}%
    \begin{subfigure}[b]{\twolensplash}
        \resizebox{1.02\textwidth}{!}{
            \makespy{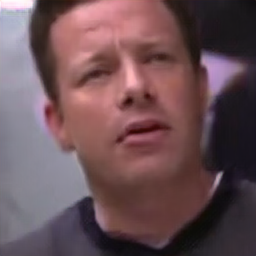}
        }
        \vspace*{-5mm}\caption{MSE}
    \end{subfigure}%
    \begin{subfigure}[b]{\twolensplash}
        \resizebox{1.02\textwidth}{!}{
            \makespy{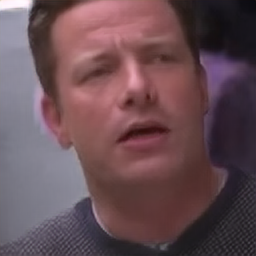}
        }
        \vspace*{-5mm}\caption{VGG-19}
    \end{subfigure}%
    \begin{subfigure}[b]{\twolensplash}
        \resizebox{1.02\textwidth}{!}{
            \makespy{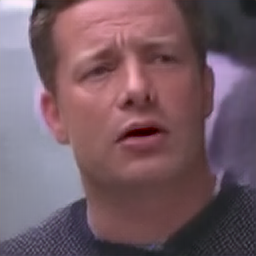}
        }
        \vspace*{-5mm}\caption{ResNet-4}
    \end{subfigure}%
    \begin{subfigure}[b]{\twolensplash}
        \resizebox{1.02\textwidth}{!}{
            \makespy{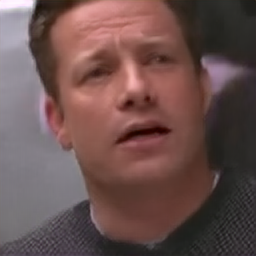}
        }
        \vspace*{-5mm}\caption{ResNet-6}
    \end{subfigure}%
    \begin{subfigure}[b]{\twolensplash}
        \resizebox{1.02\textwidth}{!}{
            \makespy{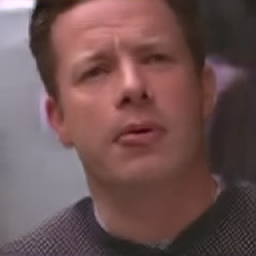}
        }
        \vspace*{-5mm}\caption{ResNet-8}
    \end{subfigure}%
    \begin{subfigure}[b]{\twolensplash}
        \resizebox{1.02\textwidth}{!}{
            \makespy{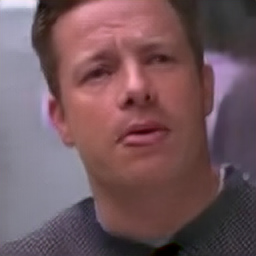}
        }
        \vspace*{-5mm}\caption{UNet}
    \end{subfigure}
    \caption{Examples of reconstructed ``holdout'' images from neural talking head models fine-tuned with different losses. Note that details with PGNs are better than with MSE.}
    \label{fig:avatars-comparison}
\end{figure*}
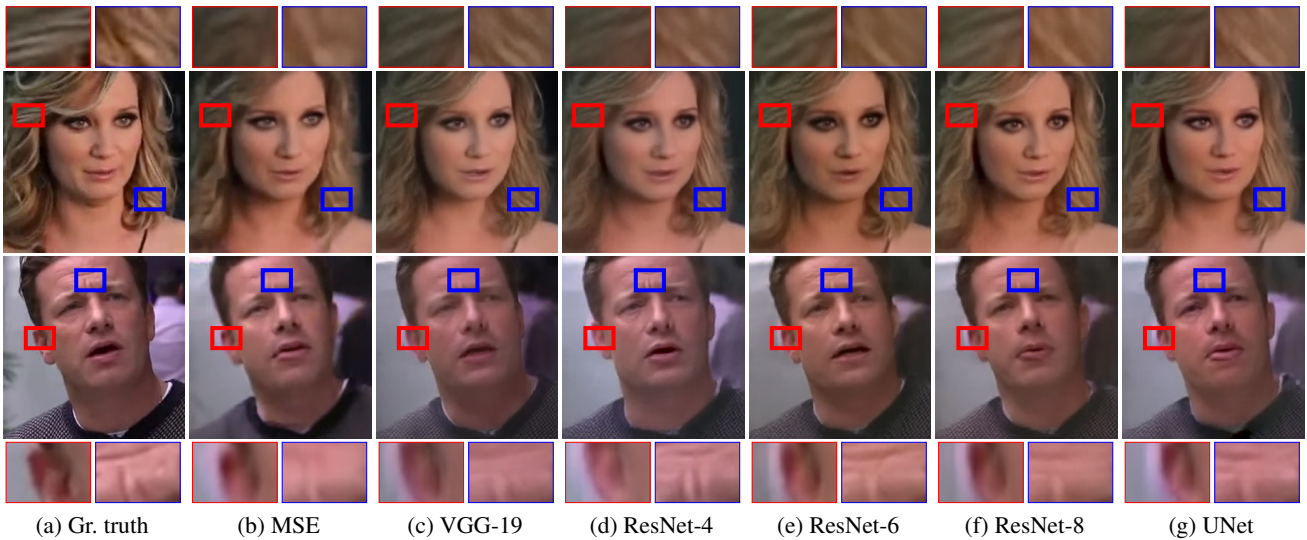

{\setlength{\tabcolsep}{0.5em}
\begin{table*}[t]
    \centering
    \begin{tabular}{|l|lll|lll|}
        \hline
                    &                                   & \multicolumn{1}{c}{Train}         &                            
                    &                                   & \multicolumn{1}{c}{Holdout}       &                                   \\
        \cline{2-7}
                    & MSE $_{\times 10^3} \downarrow$   & PL $_\textrm{VGG-19} \downarrow$  & LPIPS $\downarrow$
                    & MSE $_{\times 10^3} \downarrow$   & PL $_\textrm{VGG-19} \downarrow$  & LPIPS $\downarrow$                \\ \hline
        MSE         & \textbf{3.87} \tiny{$\pm$ 2.69}   & 3.66 \tiny{$\pm$ 0.53}            & 0.28 \tiny{$\pm$ 0.05}
                    & \textbf{16.9} \tiny{$\pm$ 14.9}   & 4.53 \tiny{$\pm$ 0.69}            & 0.36 \tiny{$\pm$ 0.07}            \\
        PL$_\textrm{VGG-19}$ & 10.9 \tiny{$\pm$ 8.3}             & \textbf{2.77} \tiny{$\pm$ 0.47}   & \textbf{0.19} \tiny{$\pm$ 0.04}
                    & 20.5 \tiny{$\pm$ 13.4}            & \textbf{4.26} \tiny{$\pm$ 0.68}   & \textbf{0.33} \tiny{$\pm$ 0.08}   \\
        \hline
        ResNet-4    & \textbf{5.56} \tiny{$\pm$ 2.98}   & \textbf{2.86} \tiny{$\pm$ 0.49}   & \textbf{0.20} \tiny{$\pm$ 0.04}   
                    & 17.8 \tiny{$\pm$ 13.7}            & \textbf{4.34} \tiny{$\pm$ 0.70}   & \textbf{0.34} \tiny{$\pm$ 0.08}   \\
        ResNet-6    & 5.80 \tiny{$\pm$ 3.20}            & 2.89 \tiny{$\pm$ 0.47}            & 0.20 \tiny{$\pm$ 0.04}            
                    & 17.8 \tiny{$\pm$ 13.4}            & 4.38 \tiny{$\pm$ 0.69}            & 0.34 \tiny{$\pm$ 0.08}            \\
        ResNet-8    & 5.66 \tiny{$\pm$ 3.08}            & 2.88 \tiny{$\pm$ 0.48}            & 0.20 \tiny{$\pm$ 0.04}          
                    & 17.9 \tiny{$\pm$ 13.8}            & 4.37 \tiny{$\pm$ 0.70}            & \textbf{0.34} \tiny{$\pm$ 0.08}   \\
        UNet        & 6.2 \tiny{$\pm$ 3.1}              & 2.97 \tiny{$\pm$ 0.46}            & 0.21 \tiny{$\pm$ 0.04}          
                    & \textbf{17.7} \tiny{$\pm$ 12.9}   & 4.39 \tiny{$\pm$ 0.70}            & 0.34 \tiny{$\pm$ 0.08}            \\ \hline
    \end{tabular}
    \caption{Quantitative evaluation of fine-tuning quality in Neural Talking Head Models.}
    \label{tab:avatars}
\end{table*}
}

To facilitate experimentation, we first prepare style vectors for $N = 100$ images by optimizing them for $T_0 = 5000$ iterations using MSE only. Then we use these vectors as starting points for further optimization for $T = 1000$ more iterations with either MSE, VGG-based perceptual loss, or PGN.

A critical part of Image2StyleGAN is the combination of two losses: MSE and VGG-based perceptual loss. We implemented this in our experiments, but, since in Deep Image Prior experiments we observed that PGN does not reproduce colors precisely when used in isolation, we increase the coefficient for MSE loss from 1 to 20 when optimizing using PGN. 

Again, we see that not only optimization with synthetic gradients does not diverge, but also it bridges a large part of the quality gap between optimization via MSE and via real gradient of the perceptual loss (see \tab{dip-stylegan} and \fig{stylegan-comparison}). 

Inference with PGN results in higher perceptual loss than inference with the VGG-19 network. A natural question is then \textit{does such difference matter}? To answer it, we conducted user preference study using crowdsourcing. We provided crowdsourcers with pairs of images corresponding to different reconstructions of the same ground truth image and asked them which of the two images contained more fine detail. Each pair of images was labeled by 3 different assessors. According to human judgements (\fig{stylegan-toloka}), PGN-based inference occasionally even surpasses the real perceptual loss, and in general performs very similarly to VGG-based inference according to the user preferences. In contrast, users do notice that inference based on MSE-loss lack the detailization provided by PGN-based and VGG-based inference.

\vspace{-2mm}

\paragraph{Neural Talking Head Models} is another highly practical case. It considers the need to create realistic user avatars on-device. In such a scenario, avoiding the need to store and execute the full VGG network might be a clear advantage. 

\begin{figure*}[t]
    \centering
    \redeflen{twolensplash}{0.166\linewidth}
    \renewcommand\makespy[1]{%
        \begin{tikzpicture}[spy using outlines={rectangle, magnification=3, height=3cm, width=4.2cm, every spy on node/.append style={line width=2mm}}]
                \node (nd1){\includegraphics{#1}};
                \RelativeSpy{nd1-spy1}{nd1}{(0.1,0.35)}{(0.255,1.17)}{red}
                \RelativeSpy{nd1-spy2}{nd1}{(0.67,0.57)}{(0.735,1.17)}{blue}
        \end{tikzpicture}%
    }
    \begin{subfigure}[b]{\twolensplash}
        \resizebox{1.02\textwidth}{!}{
            \makespy{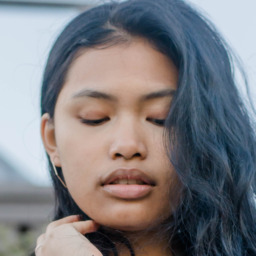}
        }
        \vspace*{-5mm}\caption{Ground truth}
    \end{subfigure}%
    \begin{subfigure}[b]{\twolensplash}
        \resizebox{1.02\textwidth}{!}{
            \makespy{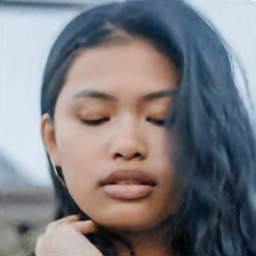}
        }
        \vspace*{-5mm}\caption{Before finetune}
    \end{subfigure}%
    \begin{subfigure}[b]{\twolensplash}
        \resizebox{1.02\textwidth}{!}{
            \makespy{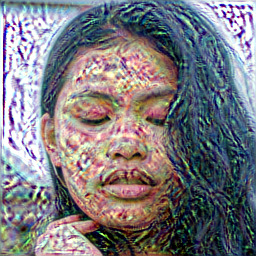}
        }
        \vspace*{-5mm}\caption{ResNet-4}
    \end{subfigure}%
    \begin{subfigure}[b]{\twolensplash}
        \resizebox{1.02\textwidth}{!}{
            \makespy{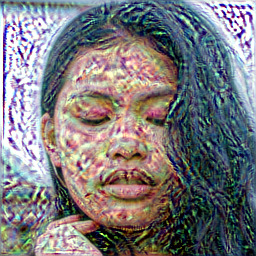}
        }
        \vspace*{-5mm}\caption{ResNet-6}
    \end{subfigure}%
    \begin{subfigure}[b]{\twolensplash}
        \resizebox{1.02\textwidth}{!}{
            \makespy{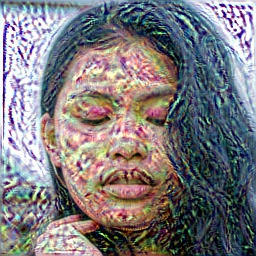}
        }
        \vspace*{-5mm}\caption{ResNet-8}
    \end{subfigure}%
    \begin{subfigure}[b]{\twolensplash}
        \resizebox{1.02\textwidth}{!}{
            \makespy{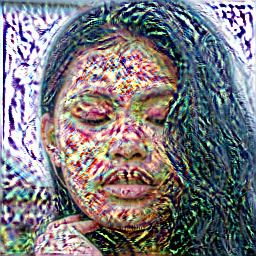}
        }
        \vspace*{-5mm}\caption{UNet}
    \end{subfigure}
    \caption{Examples of proxy targets for the first iteration of StyleGAN inversion. Note how the model highlights edges, including the blurry triangle in the bottom left corner and the fine detail in the woman's hair, and how much noise there is in the background.}
    \label{fig:proxy-comparison}
\end{figure*}
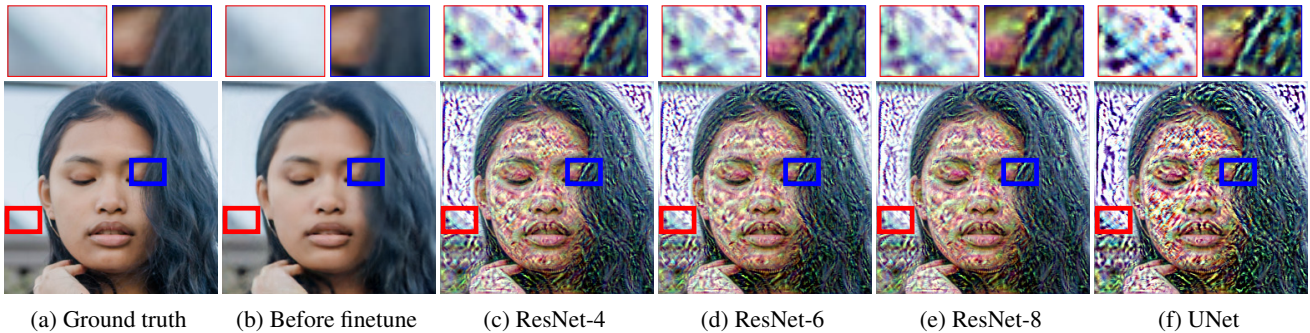

We consider an implementation of~\cite{Zakharov19} and use different losses to fine-tune the generators to $N = 50$ new identities as described in their paper. Each generator is initialized through the AdaIN-based identity encoding pretrained on the VoxCeleb2 dataset~\cite{Chung18b} and then fine-tuned using eight images of a person. This is done by minimizing perceptual loss between the generated images and the real ones (the original implementation also uses the discriminator for fine-tuning, but we disable it here to measure the impact from the perceptual loss only). Fine-tuning is performed for $T = 50$ iterations using Adam with learning rate $5 \cdot 10^{-5}$.

We demonstrate the results both on the same eight images (``train'') and on eight more images (``holdout'', same persons but unseen poses) in \tab{avatars}, with samples of reconstruction of holdout images in \fig{avatars-comparison}. We also conduct a qualitative crowdsourced evaluation and show the results in \fig{toloka} (b, c). Similarly to the experiment with StyleGAN inversion, we see that optimization via synthetic gradients is a viable alternative to optimization of the real perceptual loss that leads to comparable results in terms of user preferences and better results than the MSE loss.
\section{Discussion}
\label{sect:discussion}

We demonstrated that using a proxy target $\P \left( \hy, y; \theta \right)$ that is different from the ground truth image $y$ for mean-squared error (MSE) optimization may result in a similar effect on the resulting images as training with the original perceptual loss. We also proposed a way to approximate such target that is less time- and memory-intensive than computing the gradient of perceptual loss.

Our training pipeline relies on a set of autoencoders that we continuously train using synthetic gradients and occasionally reset. It is natural to ask whether this setup provides sufficient generalization, or whether PGNs overfit to the gradients specific to these autoencoders. In \tab{generalization} we provide evidence that our setup is fairly general. We measured cosine similarity between synthetic and real gradients during PGN training and on each of the three tasks we considered. These metrics indicate that while there is a performance gap between training and inference, it is not large, especially on shorter optimization trajectories.

\begin{table}[t]
    \centering
    \begin{tabular}{|l|llll|}
        \hline
                 & \multicolumn{4}{c|}{cosine similarity $\uparrow$}
                 \\ 
        \cline{2-5}
                 & Train                & DIP                  & Im2SG                & Heads
                 \\ 
        \hline
        ResNet-4 
                 & 0.76\tiny{$\pm$0.04} & 0.57\tiny{$\pm$0.05} & 0.72\tiny{$\pm$0.01} & 0.73\tiny{$\pm$0.01}
                 \\
        \hline
        ResNet-6 
                 & 0.78\tiny{$\pm$0.03} & 0.57\tiny{$\pm$0.06} & 0.75\tiny{$\pm$0.01} & 0.75\tiny{$\pm$0.01}
                 \\
        \hline
        ResNet-8 
                 & \textbf{0.80}\tiny{$\pm$0.03} & \textbf{0.61}\tiny{$\pm$0.04} & \textbf{0.77}\tiny{$\pm$0.01} & \textbf{0.77}\tiny{$\pm$0.01}
                 \\
        \hline
        UNet     
                 & 0.68\tiny{$\pm$0.04} & 0.42\tiny{$\pm$0.06} & 0.58\tiny{$\pm$0.02} & 0.65\tiny{$\pm$0.01}
                 \\
        \hline
    \end{tabular}
    \caption{Cosine similarity between real and synthetic gradients during the last 5120 batches of training (``Train''), after $T=3000$ iterations of Deep Image Prior optimization (``DIP''), after $T=1000$ iterations of Image2StyleGAN optimization (``Im2SG''), and after $T=50$ iterations of Neural Talking Heads (``Heads'') finetuning. DIP, Im2SG, and Heads metrics are computed on the same images as in experiments in \sect{experiments}. Note that the fewer optimization steps are performed, the better the metrics are, and this effect is stronger than distributional shift (DIP is trained on ImageNet validation set, while Im2SG uses FFHQ and Heads use VoxCeleb2). In fact, we have consistently observed that PGNs perform well at the start of trajectories, and only start to struggle after several thousands optimization steps.
    }
    \label{tab:generalization}
\end{table}

In \fig{proxy-comparison} we demonstrate examples of proxy targets for the problem of StyleGAN inversion. Note that while the output is extremely noisy, the model strongly exaggerates even blurry edges. We believe that a potential avenue for future research might be investigation of losses or synthetic gradient models that preserve this property of edge detection while keeping background regions clean, which might pave way for an improved and more efficient perceptual loss.




\FloatBarrier


{\small
\bibliographystyle{ieee_fullname}
\bibliography{refs}
}

\pagebreak

\section{Overview of supplementary materials}

\begin{figure*}[t]
    \centering
    \begin{subfigure}{\linewidth}
        \begin{center}
            \includegraphics[width=0.77\linewidth]{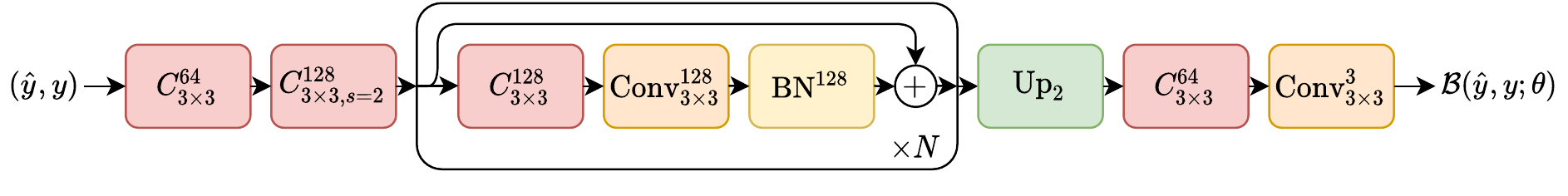}

   \end{center}     \caption{ResNet-like~\cite{he2015deep} models with $N$ blocks ($N \in \{4,6,8\}$ in the text). $\operatorname{BN}^{out}$ denotes a batch normalization layer with $out$ channels. $C^{out}_{k \times k}$ stands for a sequence of $\operatorname{Conv}^{out}_{k \times k}$, $\operatorname{BN}^{out}$, and a $\operatorname{LeakyReLU}$ with slope $0.2$. $s=2$ refers to strides.
        The architecture is heavily inspired by an implementation
        of~\cite{CycleGAN2017} and~\cite{isola2017image}
        available on Github.\footnote{\url{https://github.com/junyanz/pytorch-CycleGAN-and-pix2pix/blob/29bbc96/models/networks.py\#L315-L433}}
        }
    \end{subfigure}
    \begin{subfigure}{\linewidth}
        \begin{center}
            \includegraphics[width=0.77\linewidth]{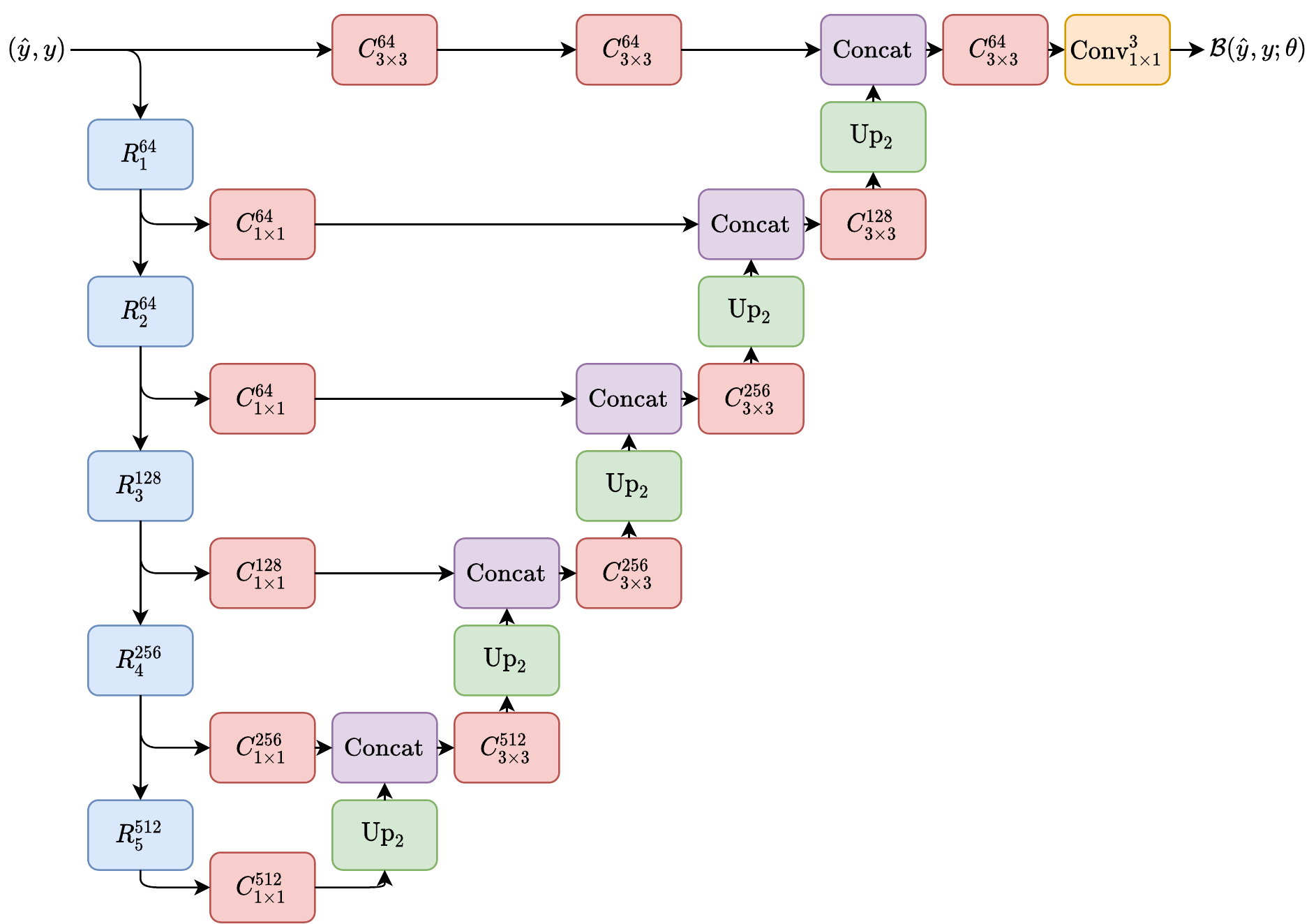}

   \end{center}     \caption{Our UNet-like~\cite{Ronneberger15} architecture. Blocks $R^{64}_1$ through $R^{512}_5$ are derived from the blocks \texttt{conv1}, \texttt{conv2\_x}--\texttt{conv5\_x} in ResNet-18~\cite{he2015deep}, with the following changes: (1) we use Double BlazeBlocks~\cite{BlazeFace} instead of plain residual blocks; (2) we change the number of input channels in $R^{64}_1$ from 3 to 6. $C^{out}_{k \times k}$ denotes a sequence of a separable~\cite{xCeption} version of $\operatorname{Conv}^{out}_{k \times k}$ followed by $\operatorname{ReLU}$. "$\operatorname{Concat}$" layers perform concatenation along the channel axis.}
    \end{subfigure}
    \begin{subfigure}{\linewidth}
        \begin{center}
            \includegraphics[width=0.77\linewidth]{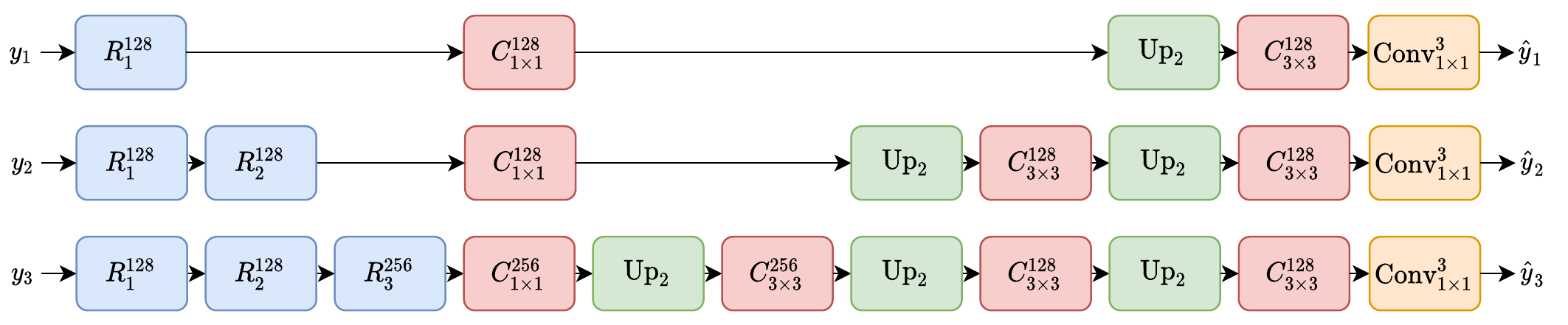}

   \end{center}     \caption{Autoencoders. On each GPU, we spawned 3 independent autoencoders with different depths, shown here on different rows. The batch $y$ (which is different for different GPUs) is split into sub-batches $y_1$, $y_2$, $y_3$, which pass through corresponding autoencoders to form $\hy_1$, $\hy_2$, $\hy_3$, and are subsequently concatenated again in the batch dimension. $R^{128}_1$, $R^{128}_2$ $R^{256}_3$ refer to the same Double BlazeBlocks as we used in the UNet architecture, but with more channels (specified in the superscripts); $C^{out}_{k \times k}$ are the same as in UNet. The whole architecture is derived from the UNet, but does not have skip connections and is more shallow.}
    \end{subfigure}

    \caption{Neural network architectures we used for all experiments in the paper. For all architectures, $\operatorname{Conv}^{out}_{k \times k}$ denotes a convolutional layer with $out$ output channels, kernel size $k \times k$, and padding $\frac {k - 1} 2$. "$\operatorname{Up}_2$" performs bilinear upsampling by a factor of $2$. Refer to subfigure captions for notation specific to each architecture.}
    \label{fig:architectures}
\end{figure*}

We show the architectures of all neural networks we experimented with in \fig{architectures} and provide full results of the Deep Image Prior stability experiment in \tab{stability}. In \sect{vgg-11}, we compare the performance of our PGN models against a VGG-11 baseline. We present more examples of performance of our models on Deep Image Prior in \fig{dip}, StyleGAN inference in \fig{stylegan}, and neural talking head models in \fig{avatartrain} (``train'') and \fig{avatartest} (``holdout''), including a comparison to the VGG-11 baseline. In \fig{proxy}, we show more examples of learned proxy targets. Finally, in \fig{styleganmse} we compare the results of StyleGAN inference using PGN with different MSE loss coefficients.

\section{Comparison with VGG-11 baseline}\label{sect:vgg-11}

In addition to the MSE and VGG19-based perceptual loss baselines we considered in the text, here we also include a comparison against a VGG11-based perceptual loss (constructed similarly to VGG-19, also up to the layer \texttt{conv5\_1}). This baseline is of interest since its forward-backward pass can be considered as a distillation of the VGG-19 forward-backward pass. Here, we include an extended version of the size/MACs table (\tab{performance}), time/memory comparison (\fig{performance}), as well as performance comparison on Deep Image Prior, StyleGAN (both in \tab{dip-stylegan}), and neural talking head models (\tab{avatars}).

We remark that our smallest ResNet-4 model (which is also the best model overall according to our experiments) computes gradients faster than VGG-11, while using less memory. While ResNet-4 is the only model that is faster than VGG-11, other ResNet models also use less memory than VGG-11. Notably, UNet is both slower than VGG-11 and uses more memory, but still has significantly fewer parameters.

While VGG-11 is a strong baseline for VGG-19 distillation, the VGG11-based perceptual loss is different compared to the VGG19-based perceptual loss, and yields different results. In this work, we focused on distilling and stabilizing the gradient of VGG19-based perceptual loss. We believe that our approach should work for distillation of other losses (including the VGG11-based perceptual loss), but we leave this for future work.

{\setlength{\tabcolsep}{0.5em}
\begin{table*}
    \centering
    \scalebox{0.97}{
    \begin{tabular}{|l|lllll|lll|lll|} 
        \hline
                        & \multicolumn{5}{c|}{Diverged}                                    & \multicolumn{3}{c|}{Stagnated} & \multicolumn{3}{c|}{Successful}  \\
        \cline{2-12}
                        & \#                 & Last it.\hspace{-3mm}   & Last PL\hspace{-3mm} & Best it.\hspace{-3mm}   & Best PL & \#   & Best it.\hspace{-3mm}   & Best PL    & \#   & Best it.\hspace{-3mm}   & Best PL     \\ 
        \hline
        Direct grad.    & 10\hspace{-2mm}    & 3897       & 40.75   & 3251       & 2.69    & 2    & 5524       & 2.49       & 88   & 9763       & \textbf{1.54}        \\
        Unconstr. proxy & 3                  & 2440       & 39.47   & 1492       & 3.80    & 12\hspace{-2mm}   & 3567       & 4.53       & 85   & 9741       & 3.14        \\ 
        \hline
        ResNet-4        & \multirow{4}{*}{0} & \multicolumn{4}{c|}{\multirow{4}{*}{n/a}}   & 2    & 1477       & 3.82       & 98   & 9795       & 1.68        \\
        ResNet-6        &                    & \multicolumn{4}{c|}{}                       & 2    & 5490       & 2.89       & 98   & 9752       & 2.05        \\
        ResNet-8        &                    & \multicolumn{4}{c|}{}                       & \textbf{1}    & 7588       & 1.36       & \textbf{99}   & 9849       & 1.91        \\
        UNet            &                    & \multicolumn{4}{c|}{}                       & \textbf{1}    & 7579       & 1.29       & \textbf{99}   & 9821       & 1.83        \\
        \hline
    \end{tabular}
    }
    \vspace{2mm}\caption{Full results of the Deep Image Prior stability experiment. We sample $N=100$ images from the ImageNet~\cite{ILSVRC15} validation set and use PGNs to train Deep Image Prior~\cite{ulyanov2017deep} networks for $T=10000$ iterations to generate these images. At each iteration, we measure the true VGG-19 perceptual loss, and we terminate training prematurely either if it becomes twice as large as its initial value (we qualify this as divergence) or if it does not improve for $2000$ iterations (stagnation). We record the iteration with the lowest perceptual loss (``Best it.'') and its value (``Best PL''). For runs that ended with divergence, we also record the last iteration (``Last it.'') and the final value of the perceptual loss (``Last PL''). We compare (1) a PGN that predicts the gradient directly, (2) a PGN that predicts the gradient via an unconstrained proxy target, and (3) multiple PGNs that apply a constraint to the proxy target (with different backbones). The first two PGNs used ResNet-8 as the backbone. All metrics in the table are averages.}
    \label{tab:stability}
\end{table*}
}

\begin{table*}
    \centering
    \vspace{-4mm}
    \begin{tabular}{|l|ll|}
        \hline
        Model    & Params & MACs  \\ \hline
        VGG-19   & 12.94M         & 108.84B* \\
        VGG-11   & 6.86M          & 42.21B*  \\ \hline
        ResNet-4 & \textbf{1.34M} & 19.79B   \\
        ResNet-6 & 1.93M          & 27.23B   \\
        ResNet-8 & 2.52M          & 34.66B   \\
        UNet     & 1.77M          & \textbf{2.10B}   \\ \hline
    \end{tabular}
    \vspace{2mm}
    \caption[]{Size comparison. For VGG-19 and VGG-11, we approximate the number of multiply-accumulate operations (MACs)
    for gradient computation with the number of MACs for a forward pass multiplied by 3 (2 forward and 1 backward).
    }
    \label{tab:performance}
    \vspace{-1.5em}
\end{table*}

\begin{figure*}
    \centering
    \includegraphics[width=\textwidth]{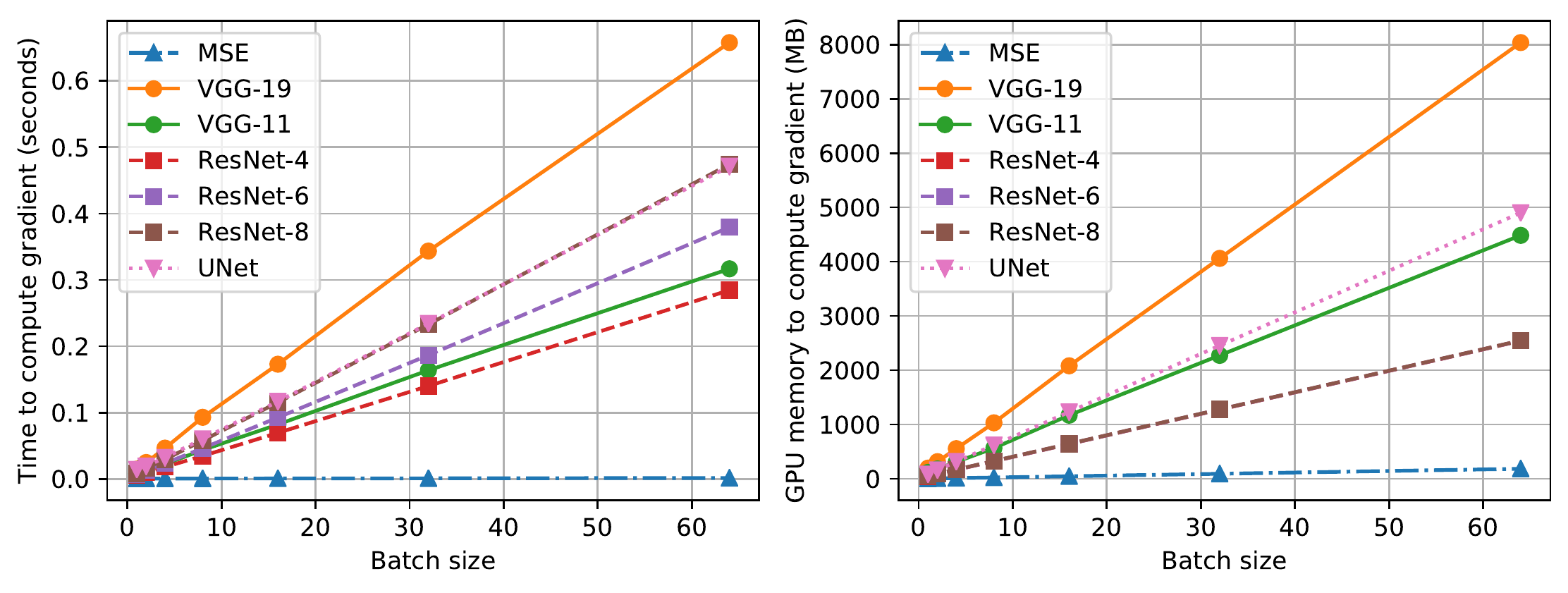}
    \vspace{-7mm}\caption{Time and memory comparison on an Nvidia GeForce 1080 Ti, including the VGG-11 baseline. Note that (1) despite having a drastically lower number of multiply-add operations, our UNet architecture runs in approximately the same time as the ResNet architecture with 8 blocks, and (2) all ResNet-based models use approximately the same amount of memory, since they differ only in the number of residual blocks.}
    \label{fig:performance}
    \vspace{-0.5em}
\end{figure*}

{\setlength{\tabcolsep}{0.5em}
\begin{table*}
    \centering
    \begin{tabular}{|l|lll|lll|}
        \hline
                    & \multicolumn{3}{c|}{Deep Image Prior} & \multicolumn{3}{c|}{Image2StyleGAN} \\
        \cline{2-7}
                    & MSE $_{\times 10^3} \downarrow$   & PL$_\textrm{VGG-19} \downarrow$ & LPIPS $\downarrow$
                    & MSE $_{\times 10^3} \downarrow$   & PL$_\textrm{VGG-19} \downarrow$ & LPIPS  $\downarrow$           \\ \hline
        MSE         & \textbf{1.43} \tiny{$\pm$ 1.06}   & 3.31 \tiny{$\pm$ 0.69}            & 0.25 \tiny{$\pm$ 0.07}          
                    & \textbf{0.83} \tiny{$\pm$ 0.48}   & 3.57 \tiny{$\pm$ 0.55}            & 0.26 \tiny{$\pm$ 0.04}            \\
        PL$_\textrm{VGG-19}$ & 2.57 \tiny{$\pm$ 2.04}            & \textbf{1.77} \tiny{$\pm$ 0.53}   & \textbf{0.11} \tiny{$\pm$ 0.04}
                    & 1.54 \tiny{$\pm$ 0.8}             & \textbf{2.72} \tiny{$\pm$ 0.47}   & \textbf{0.18} \tiny{$\pm$ 0.03}   \\
        PL$_\textrm{VGG-11}$ & 1.65 \tiny{$\pm$ 1.35}   & 1.99 \tiny{$\pm$ 0.56}            & 0.12 \tiny{$\pm$ 0.04}          
                    & 1.08 \tiny{$\pm$ 0.59}            & 2.95 \tiny{$\pm$ 0.51}            & 0.19 \tiny{$\pm$ 0.04}          \\
        \hline
        ResNet-4    & 9.13 \tiny{$\pm$ 21.2}            & \textbf{2.46} \tiny{$\pm$ 0.64}   & \textbf{0.16} \tiny{$\pm$ 0.06} 
                    & \textbf{1.41} \tiny{$\pm$ 0.71}   & 3.07 \tiny{$\pm$ 0.55}            & 0.22 \tiny{$\pm$ 0.04}            \\
        ResNet-6    & 60.3 \tiny{$\pm$ 92.4}            & 2.84 \tiny{$\pm$ 0.72}            & 0.21 \tiny{$\pm$ 0.09}          
                    & 1.46 \tiny{$\pm$ 0.73}            & 3.08 \tiny{$\pm$ 0.56}            & \textbf{0.22} \tiny{$\pm$ 0.05}   \\
        ResNet-8    & 39.6 \tiny{$\pm$ 84.1}            & 2.63 \tiny{$\pm$ 0.66}            & 0.18 \tiny{$\pm$ 0.07}          
                    & 1.46 \tiny{$\pm$ 0.71}            & \textbf{3.06} \tiny{$\pm$ 0.55}   & \textbf{0.22} \tiny{$\pm$ 0.04}   \\
        UNet        & \textbf{4.81} \tiny{$\pm$ 3.9}    & 2.48 \tiny{$\pm$ 0.58}            & 0.17 \tiny{$\pm$ 0.05}          
                    & 1.50 \tiny{$\pm$ 0.64}            & 3.19 \tiny{$\pm$ 0.54}            & 0.24 \tiny{$\pm$ 0.04}            \\ \hline
    \end{tabular}
    \vspace{2mm}
    \caption{Quantitative evaluation of reconstruction in Deep Image Prior and Image2StyleGAN, including the VGG-11 baseline.}
    \label{tab:dip-stylegan}
    \vspace{-1.5em}
\end{table*}
}

{\setlength{\tabcolsep}{0.5em}
\begin{table*}
    \centering
    \begin{tabular}{|l|lll|lll|}
        \hline
                    &                                   & \multicolumn{1}{c}{Train}         &                            
                    &                                   & \multicolumn{1}{c}{Holdout}       &                                   \\
        \cline{2-7}
                    & MSE $_{\times 10^3} \downarrow$   & PL $_\textrm{VGG-19} \downarrow$  & LPIPS $\downarrow$
                    & MSE $_{\times 10^3} \downarrow$   & PL $_\textrm{VGG-19} \downarrow$  & LPIPS $\downarrow$                \\ \hline
        MSE         & \textbf{3.87} \tiny{$\pm$ 2.69}   & 3.66 \tiny{$\pm$ 0.53}            & 0.28 \tiny{$\pm$ 0.05}
                    & \textbf{16.9} \tiny{$\pm$ 14.9}   & 4.53 \tiny{$\pm$ 0.69}            & 0.36 \tiny{$\pm$ 0.07}            \\
        PL$_\textrm{VGG-19}$ & 10.9 \tiny{$\pm$ 8.3}             & \textbf{2.77} \tiny{$\pm$ 0.47}   & 0.19 \tiny{$\pm$ 0.04}
                    & 20.5 \tiny{$\pm$ 13.4}            & \textbf{4.26} \tiny{$\pm$ 0.68}   & 0.33 \tiny{$\pm$ 0.08}   \\
        PL$_\textrm{VGG-11}$ & 8.25 \tiny{$\pm$ 7.07}   & 2.83 \tiny{$\pm$ 0.49}            & \textbf{0.18} \tiny{$\pm$ 0.04}
                    & 19.12 \tiny{$\pm$ 13.65}        & 4.36 \tiny{$\pm$ 0.71}              & \textbf{0.32} \tiny{$\pm$ 0.08} \\
        \hline
        ResNet-4    & \textbf{5.56} \tiny{$\pm$ 2.98}   & \textbf{2.86} \tiny{$\pm$ 0.49}   & \textbf{0.20} \tiny{$\pm$ 0.04}   
                    & 17.8 \tiny{$\pm$ 13.7}            & \textbf{4.34} \tiny{$\pm$ 0.70}   & \textbf{0.34} \tiny{$\pm$ 0.08}   \\
        ResNet-6    & 5.80 \tiny{$\pm$ 3.20}            & 2.89 \tiny{$\pm$ 0.47}            & 0.20 \tiny{$\pm$ 0.04}            
                    & 17.8 \tiny{$\pm$ 13.4}            & 4.38 \tiny{$\pm$ 0.69}            & 0.34 \tiny{$\pm$ 0.08}            \\
        ResNet-8    & 5.66 \tiny{$\pm$ 3.08}            & 2.88 \tiny{$\pm$ 0.48}            & 0.20 \tiny{$\pm$ 0.04}          
                    & 17.9 \tiny{$\pm$ 13.8}            & 4.37 \tiny{$\pm$ 0.70}            & \textbf{0.34} \tiny{$\pm$ 0.08}   \\
        UNet        & 6.2 \tiny{$\pm$ 3.1}              & 2.97 \tiny{$\pm$ 0.46}            & 0.21 \tiny{$\pm$ 0.04}          
                    & \textbf{17.7} \tiny{$\pm$ 12.9}   & 4.39 \tiny{$\pm$ 0.70}            & 0.34 \tiny{$\pm$ 0.08}            \\ \hline
    \end{tabular}
    \vspace{2mm}\caption{Quantitative evaluation of fine-tuning quality in Neural Talking Head Models, including the VGG-11 baseline.}
    \label{tab:avatars}
    \vspace{-2em}
\end{table*}
}

\newcommand{\dipfigure}[1]{%
\begin{center}
    \includegraphics[width=0.72\linewidth]{figures/model_comparison/supmat_2/01_dip/#1.jpg}
    \begin{tabular}{C{0.066\linewidth}C{0.066\linewidth}C{0.066\linewidth}C{0.066\linewidth}C{0.066\linewidth}C{0.066\linewidth}C{0.066\linewidth}C{0.066\linewidth}}
        \footnotesize{Gr. truth} & \footnotesize{MSE} & \footnotesize{VGG-19} & \footnotesize{VGG-11} & \footnotesize{ResNet-4} & \footnotesize{ResNet-6} & \footnotesize{ResNet-8} & \footnotesize{UNet} \\
    \end{tabular}

\end{center}%
}

\newcommand{\dipcaption}{%
\caption{Examples of image reconstruction using Deep Image Prior~\cite{ulyanov2017deep} networks with different losses. Here, we show what images learned purely with PGNs (without supplementing them with MSE loss) look like. Note how without MSE some PGN models occasionally end up with large blobs of incorrect color; as we demonstrate in StyleGAN experiments, this effect disappears when we use PGNs together with MSE. Zoom in to see finer details.}%
}

\begin{figure*}[p] \dipfigure{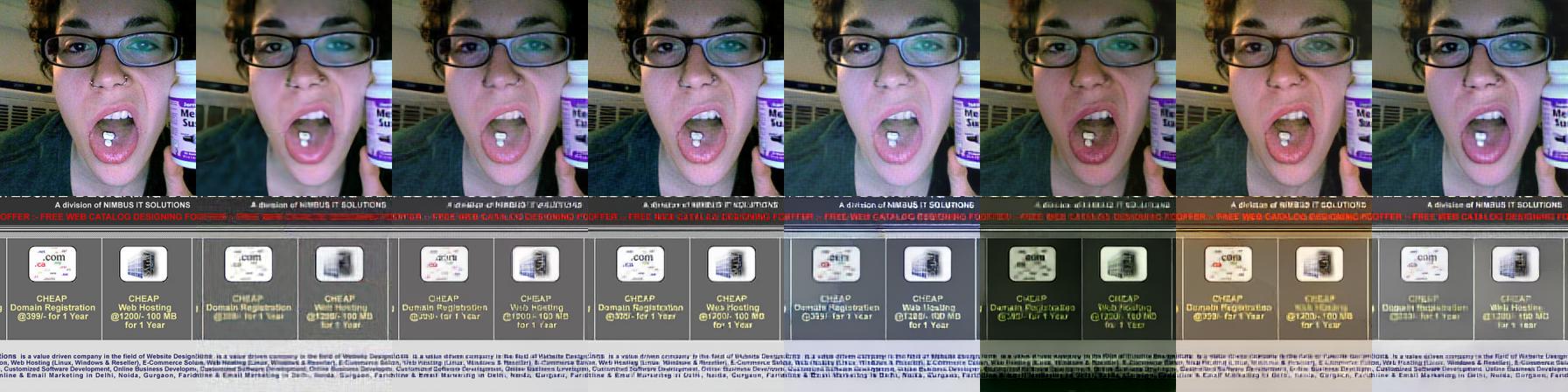} \dipfigure{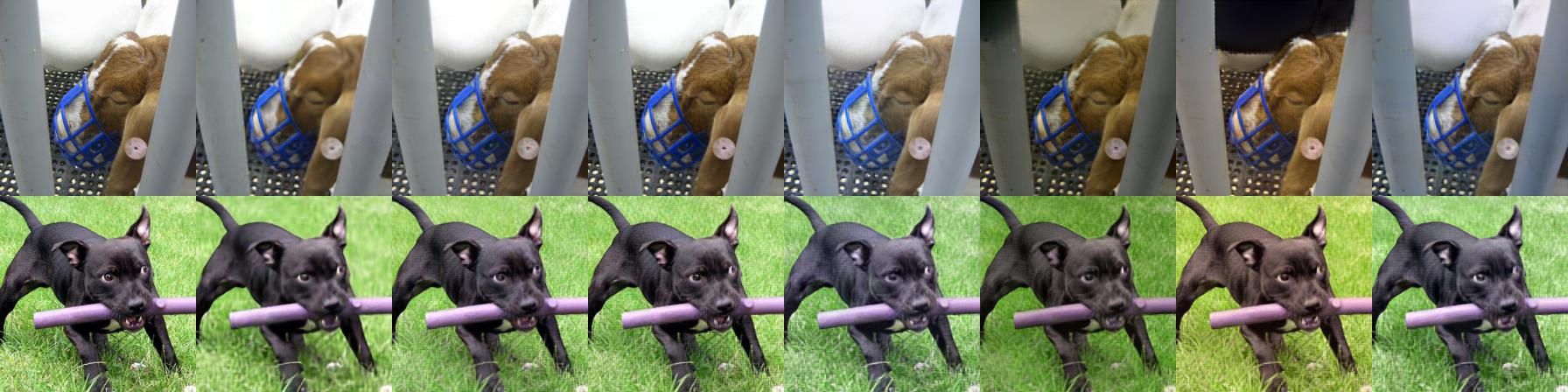} \dipfigure{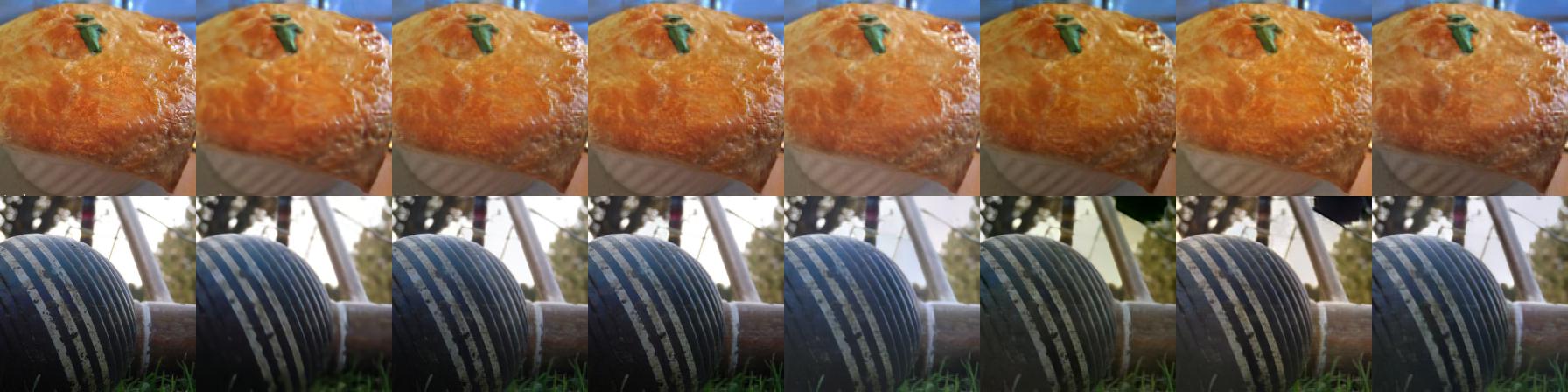} \dipfigure{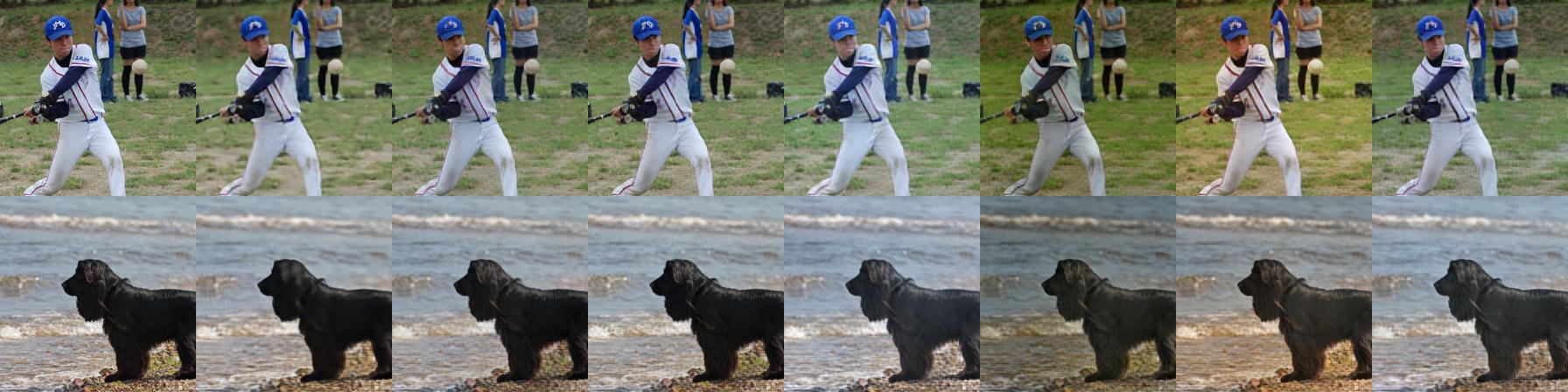} \dipfigure{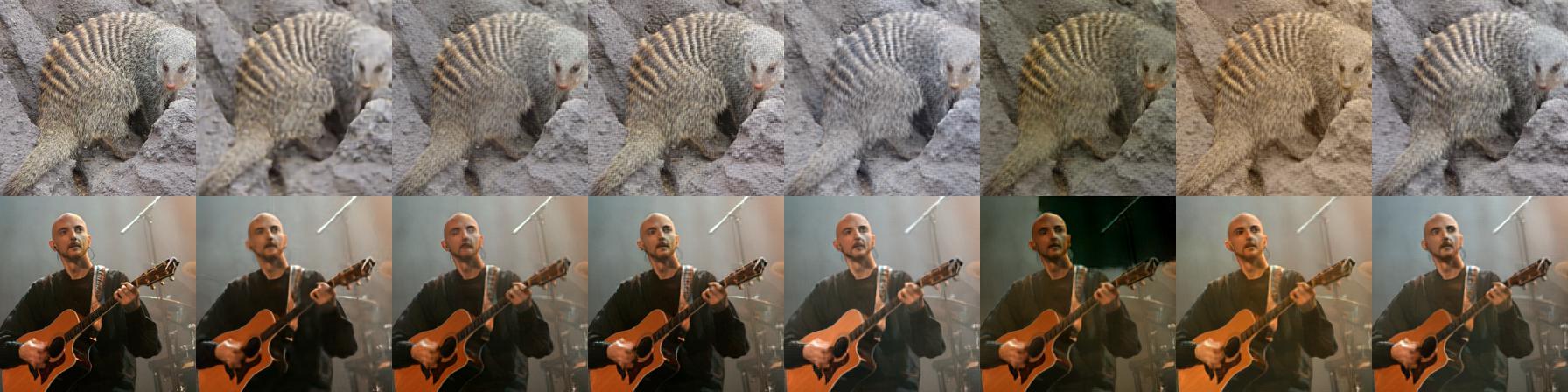} \dipcaption \label{fig:dip} \end{figure*}
\begin{figure*}[p]\ContinuedFloat \dipfigure{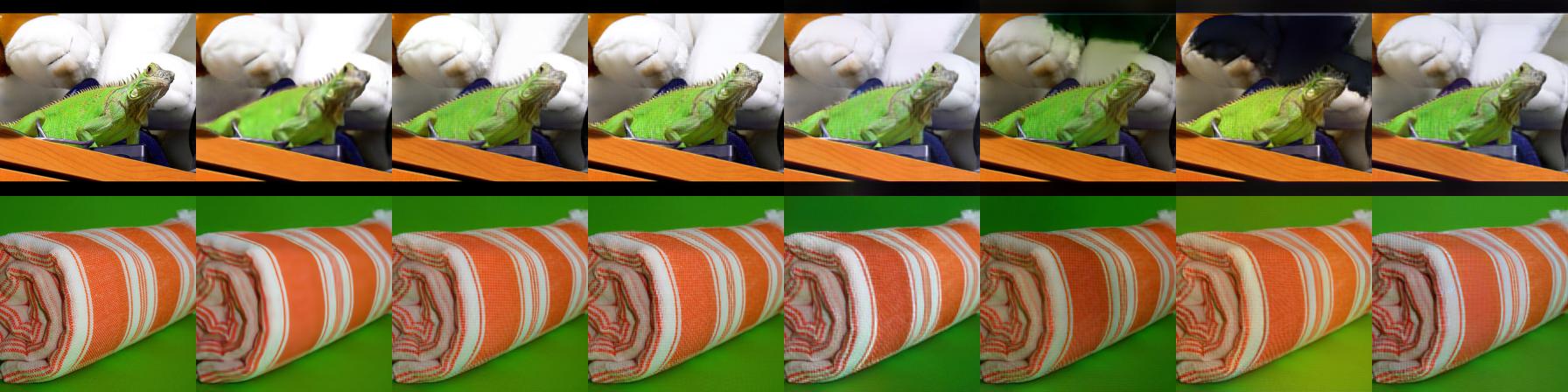} \dipfigure{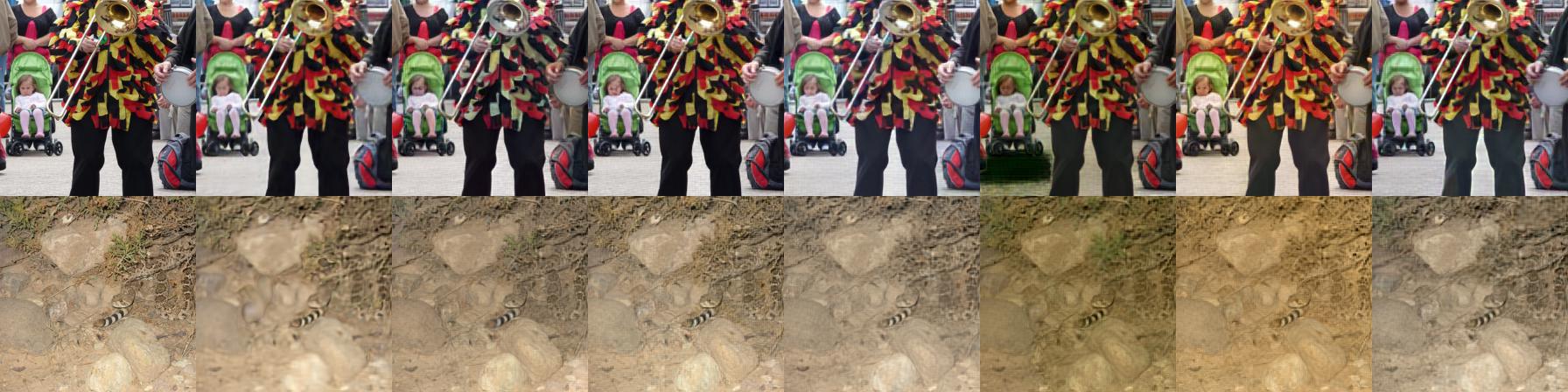} \dipfigure{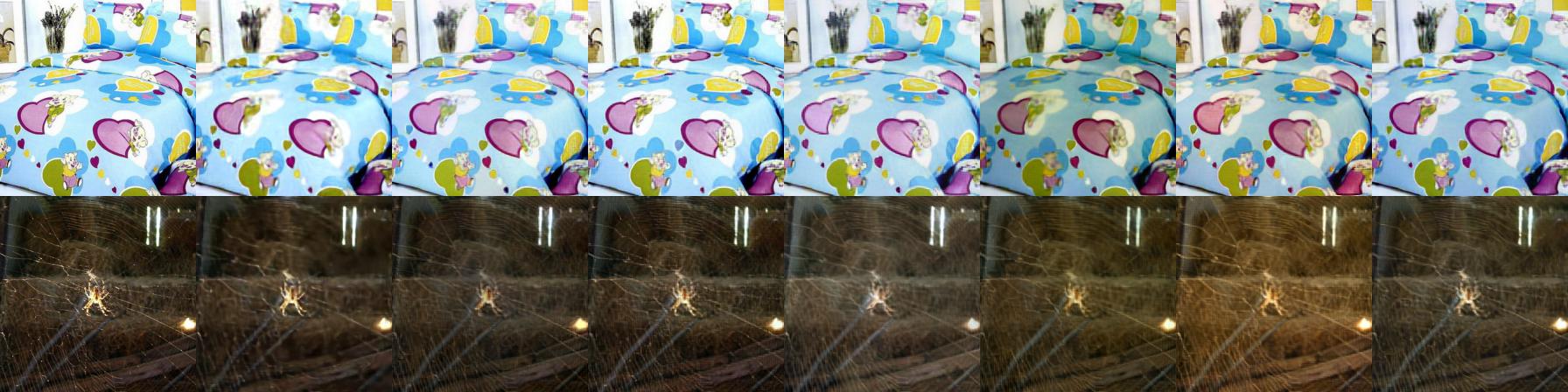} \dipfigure{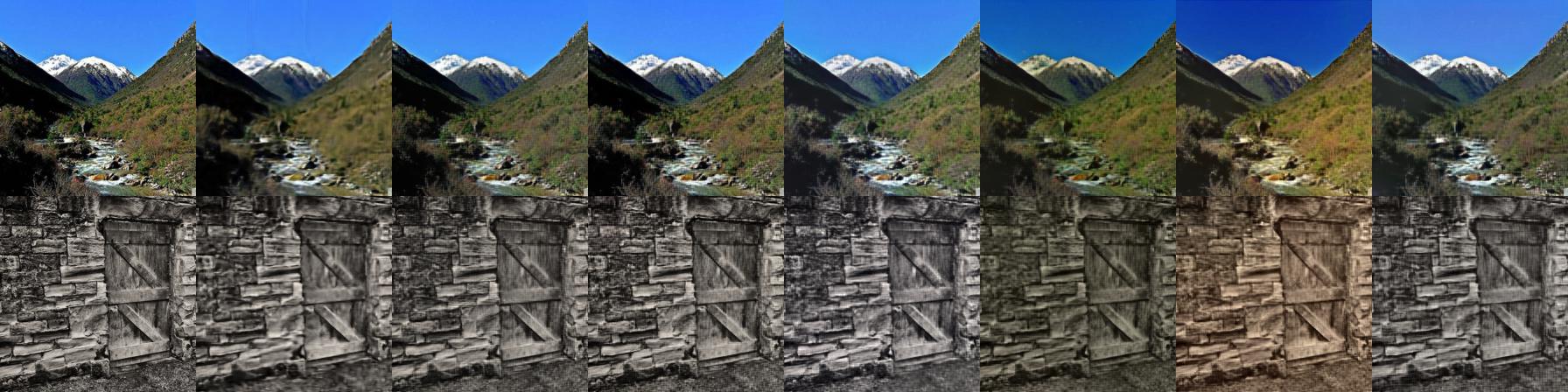} \dipfigure{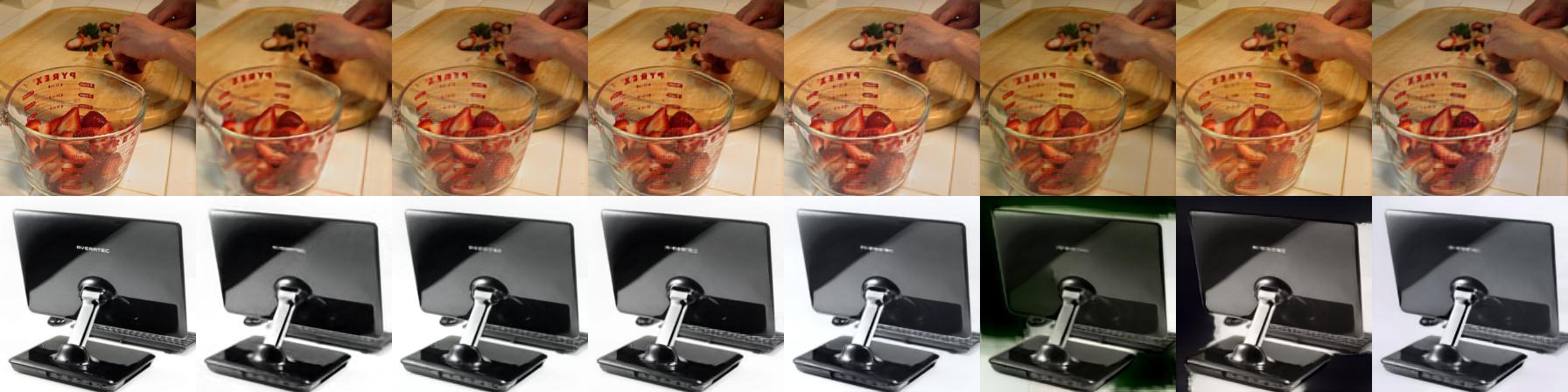} \dipcaption \end{figure*}
\begin{figure*}[p]\ContinuedFloat \dipfigure{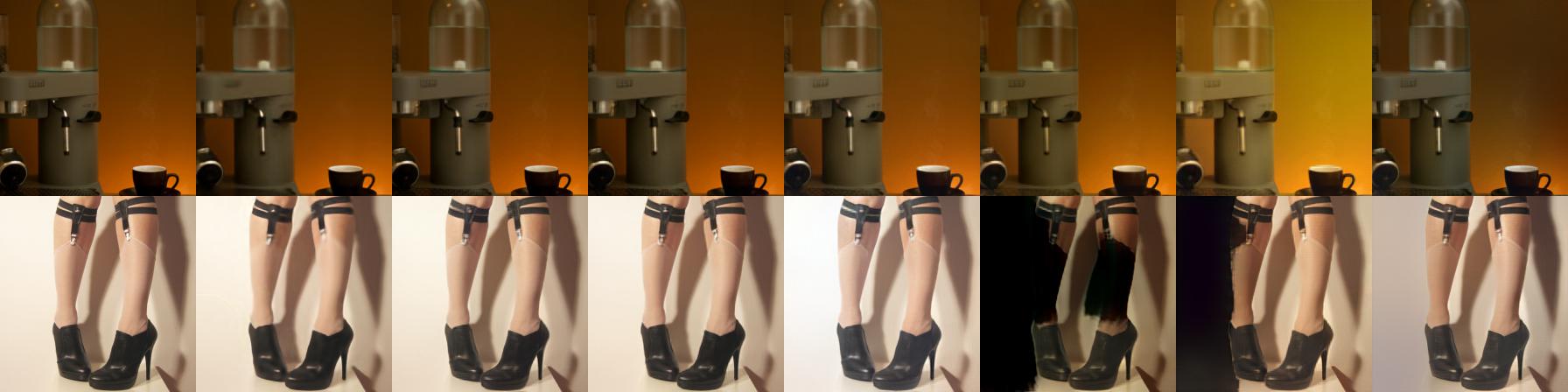} \dipfigure{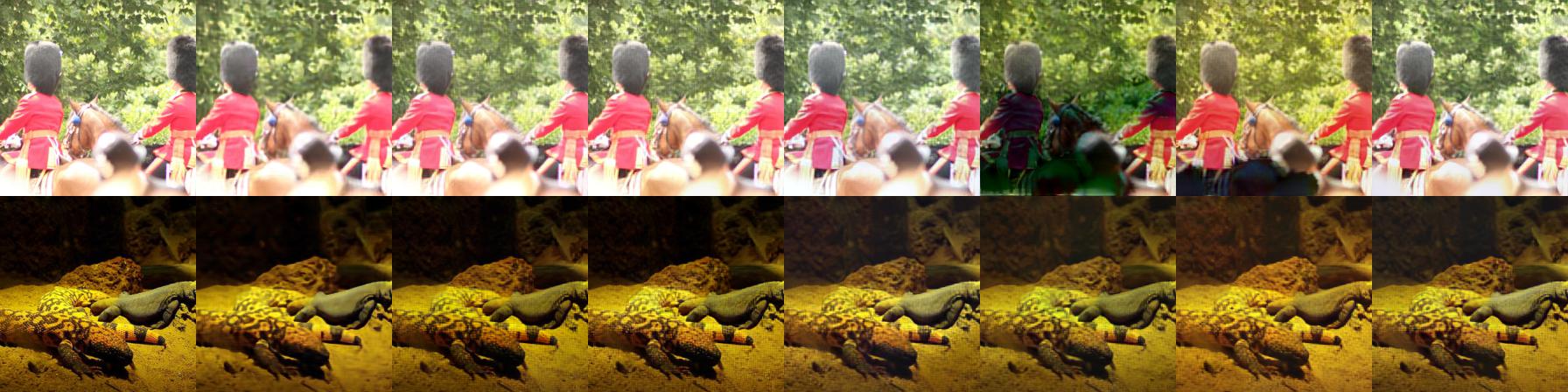} \dipfigure{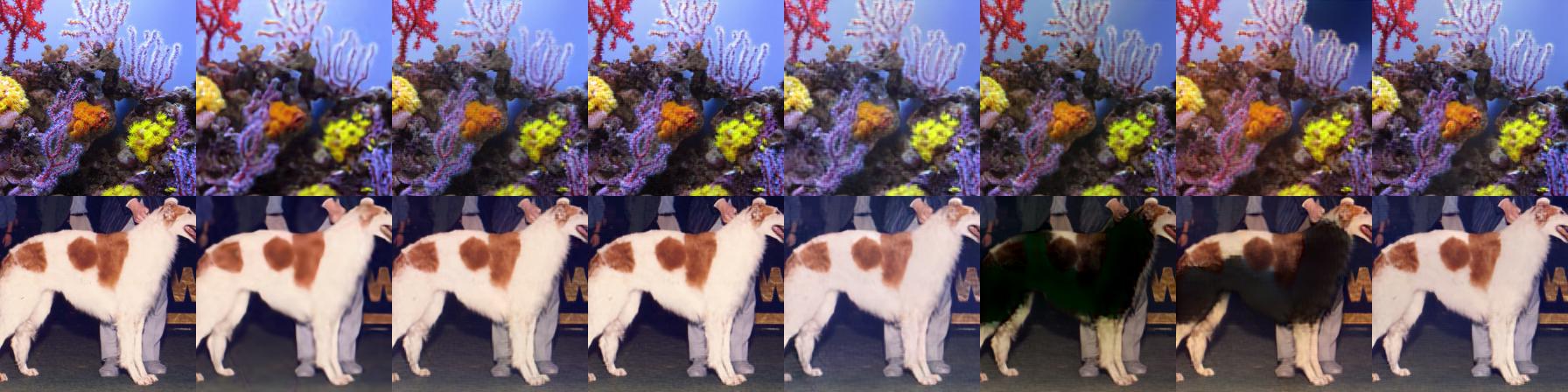} \dipfigure{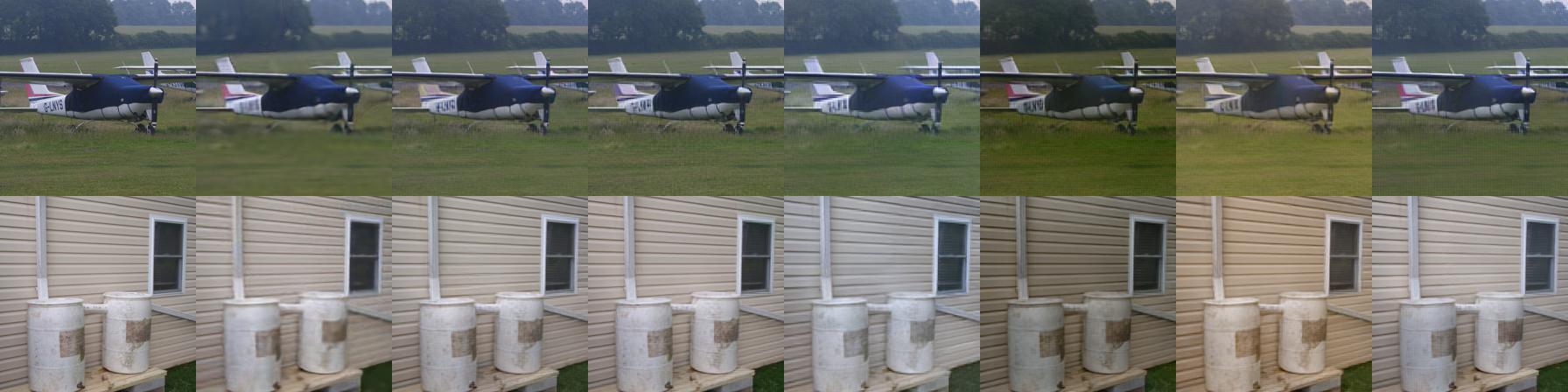} \dipfigure{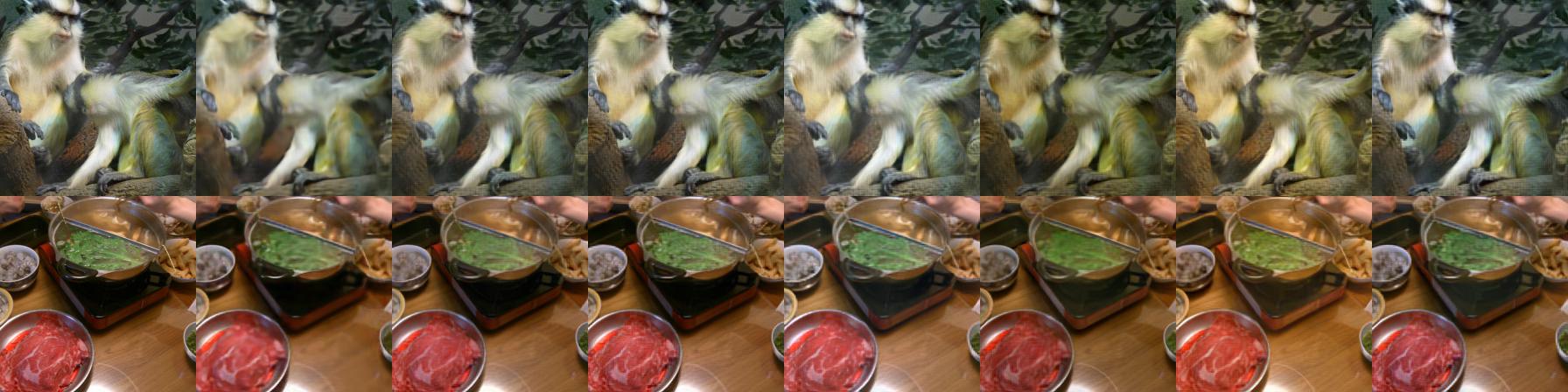} \dipcaption \end{figure*}

\newcommand{\styleganfigure}[1]{%
\begin{center}
    \includegraphics[width=0.72\linewidth]{figures/model_comparison/supmat_2/02_stylegan/#1.jpg}
    \begin{tabular}{C{0.066\linewidth}C{0.066\linewidth}C{0.066\linewidth}C{0.066\linewidth}C{0.066\linewidth}C{0.066\linewidth}C{0.066\linewidth}C{0.066\linewidth}}
        \footnotesize{Gr. truth} & \footnotesize{MSE} & \footnotesize{VGG-19} & \footnotesize{VGG-11} & \footnotesize{ResNet-4} & \footnotesize{ResNet-6} & \footnotesize{ResNet-8} & \footnotesize{UNet} \\
\end{tabular}
\end{center}%
}

\newcommand{\stylegancaption}{%
\caption{Examples of image reconstruction using Image2StyleGAN~\cite{Abdal19} with different losses. Following the original paper, we supplement optimization of perceptual loss with MSE loss. For VGG-based perceptual loss, we use MSE loss with coefficient $\lambda_{\operatorname{MSE}} = 1$. For PGN models, we use $\lambda_{\operatorname{MSE}} = 20$. See \fig{styleganmse} for examples of inference with other coefficients. Zoom in to see finer details.}%
}

\begin{figure*}[p] \styleganfigure{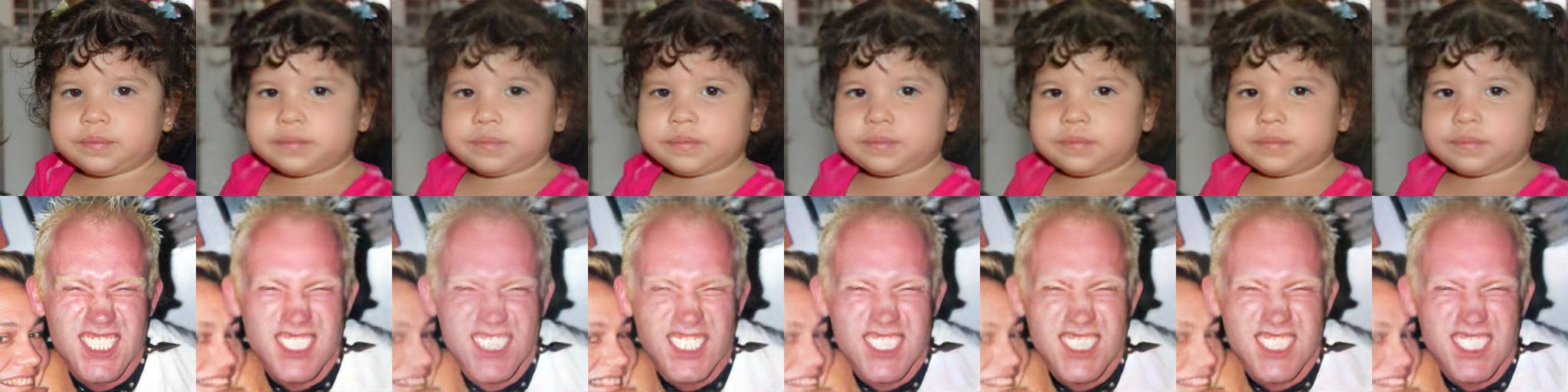} \styleganfigure{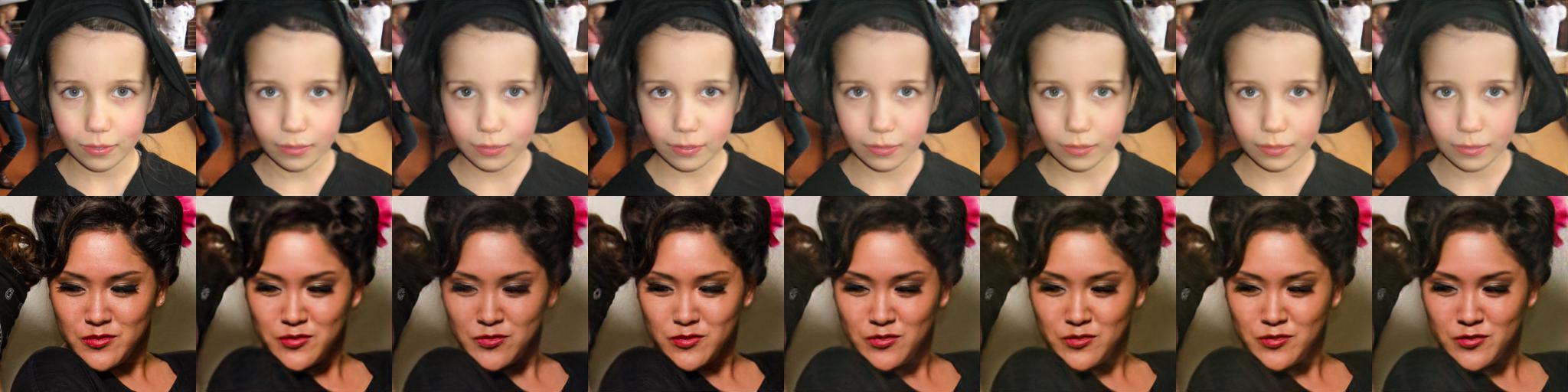} \styleganfigure{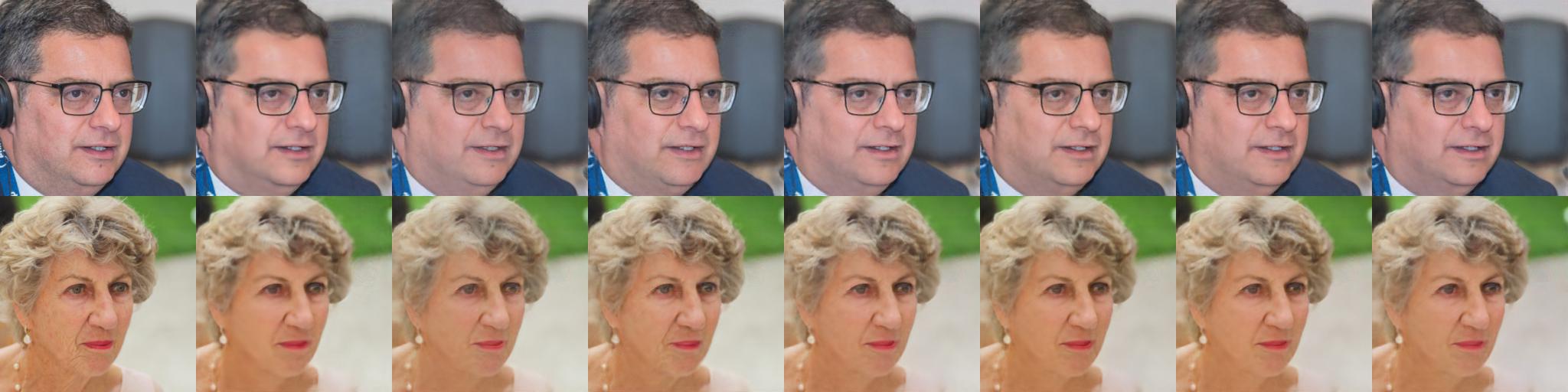} \styleganfigure{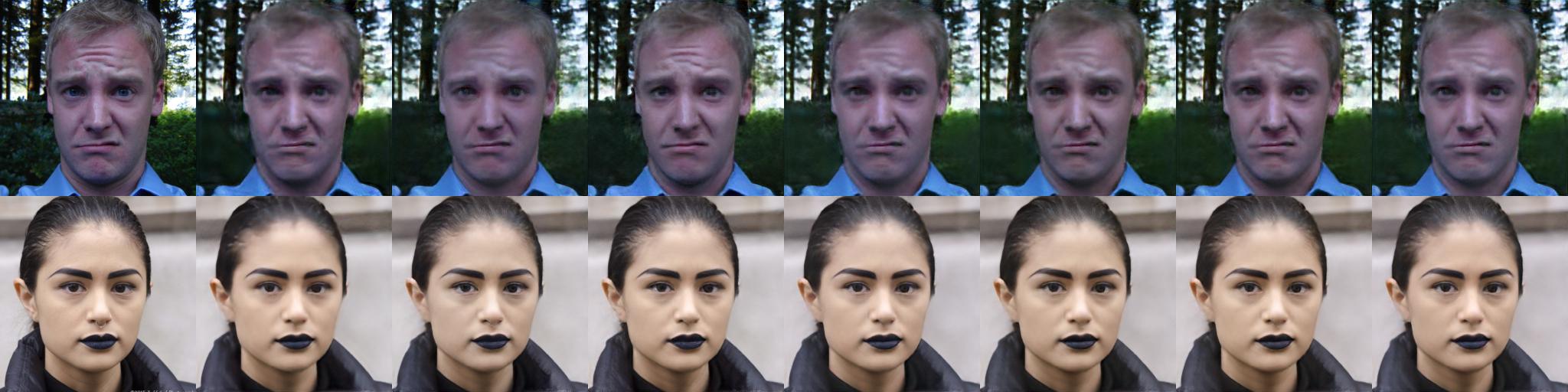} \styleganfigure{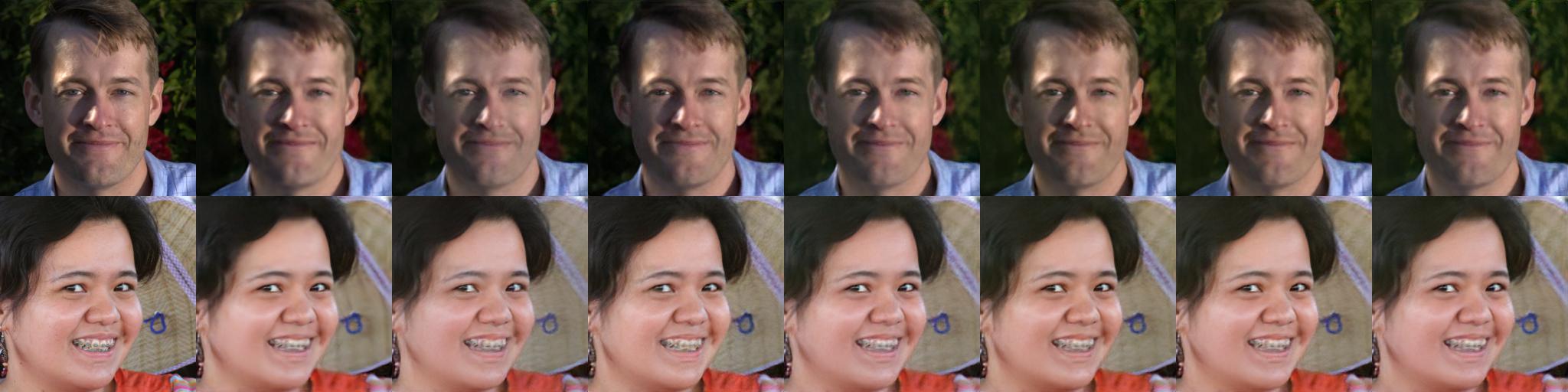} \stylegancaption \label{fig:stylegan} \end{figure*}
\begin{figure*}[p]\ContinuedFloat \styleganfigure{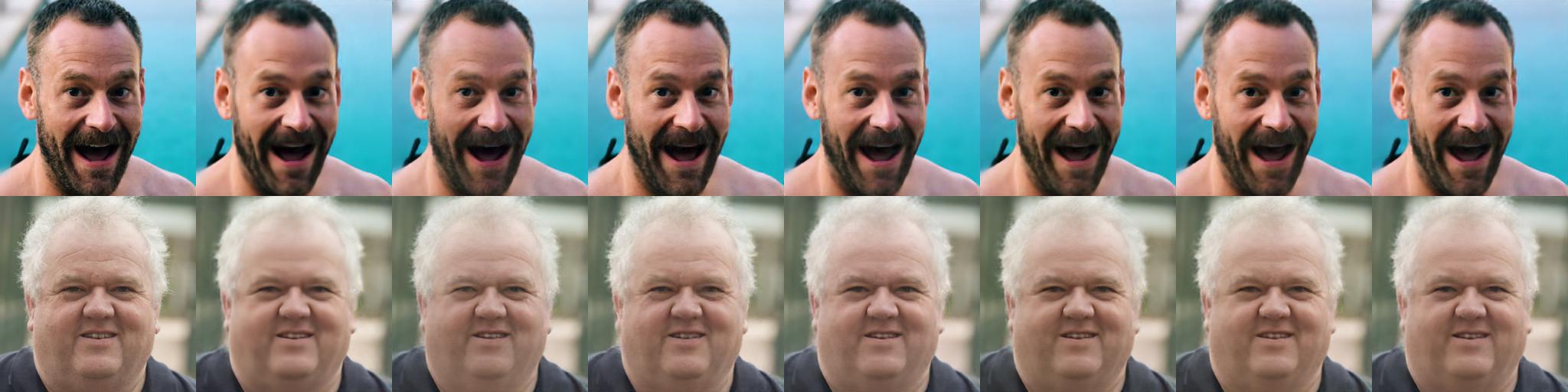} \styleganfigure{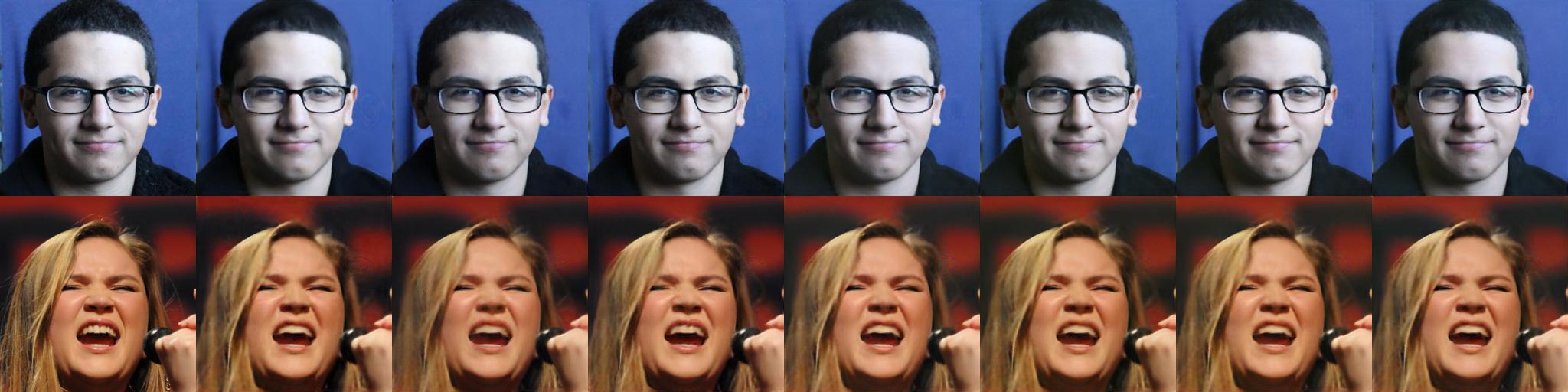} \styleganfigure{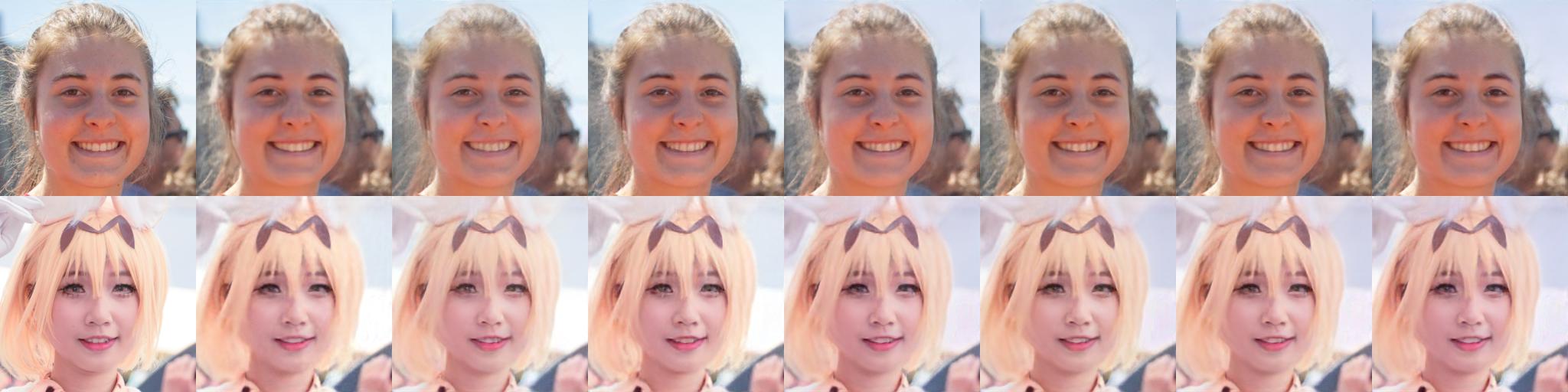} \styleganfigure{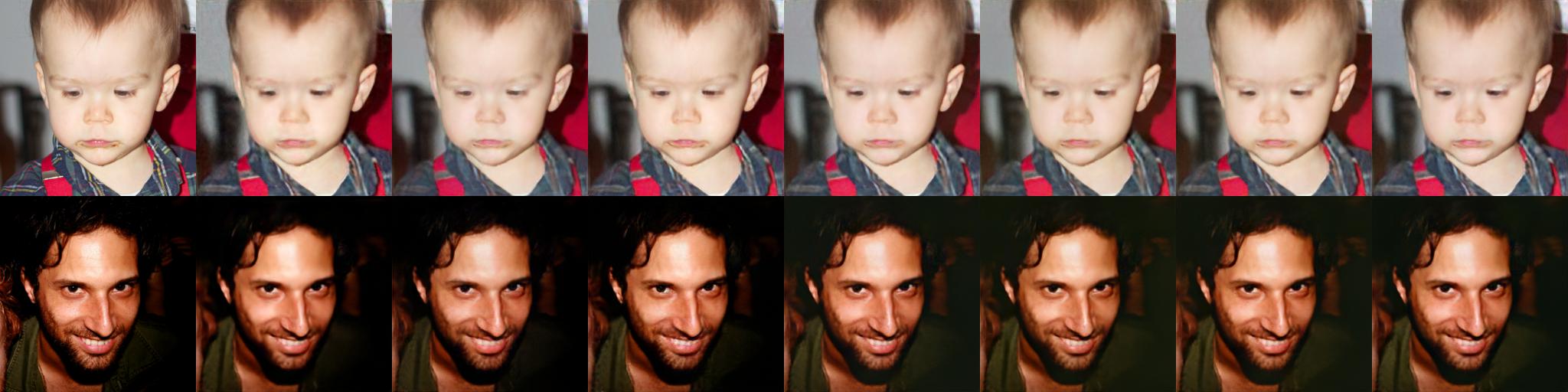} \styleganfigure{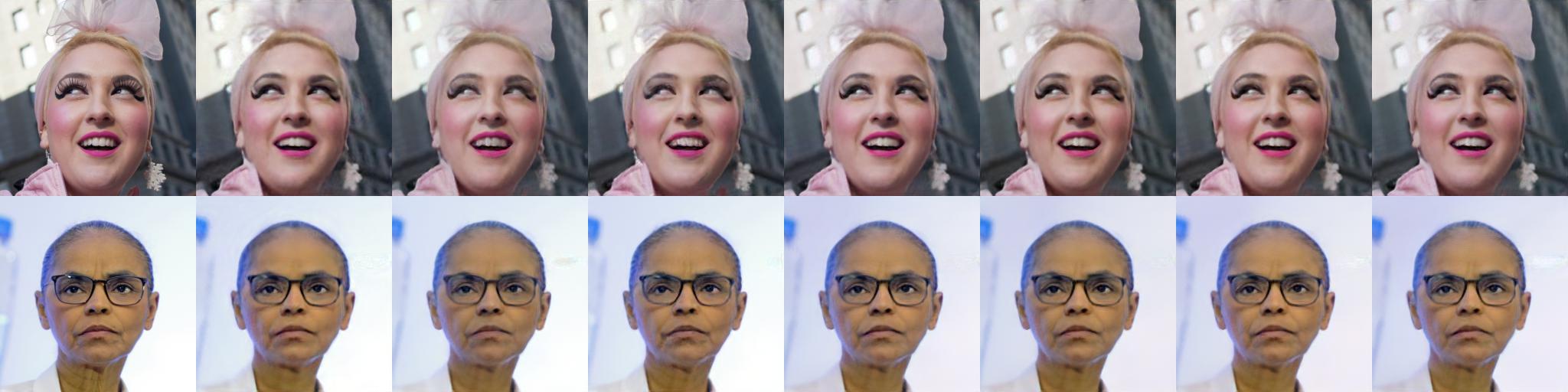} \stylegancaption \end{figure*}
\begin{figure*}[p]\ContinuedFloat \styleganfigure{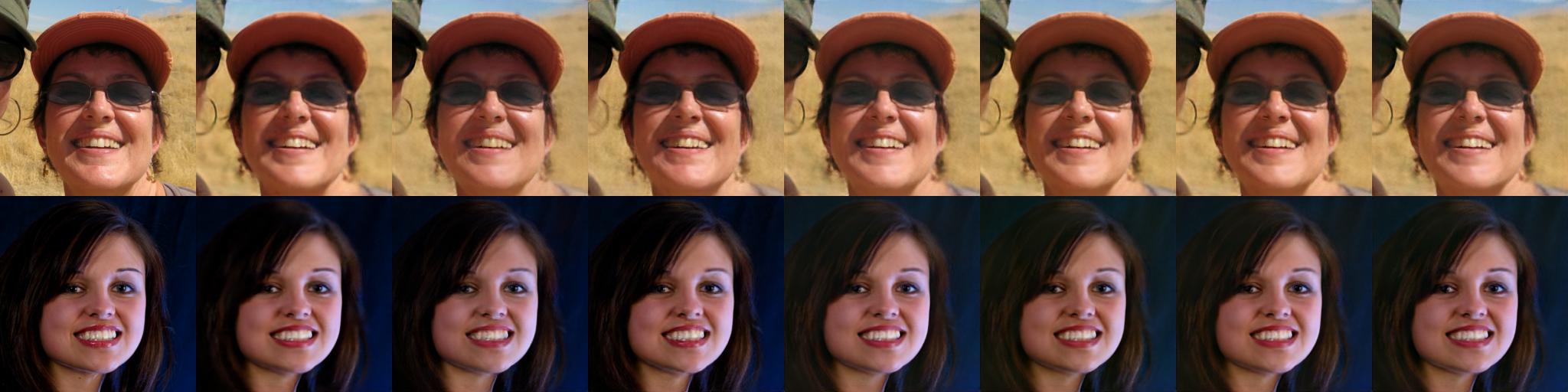} \styleganfigure{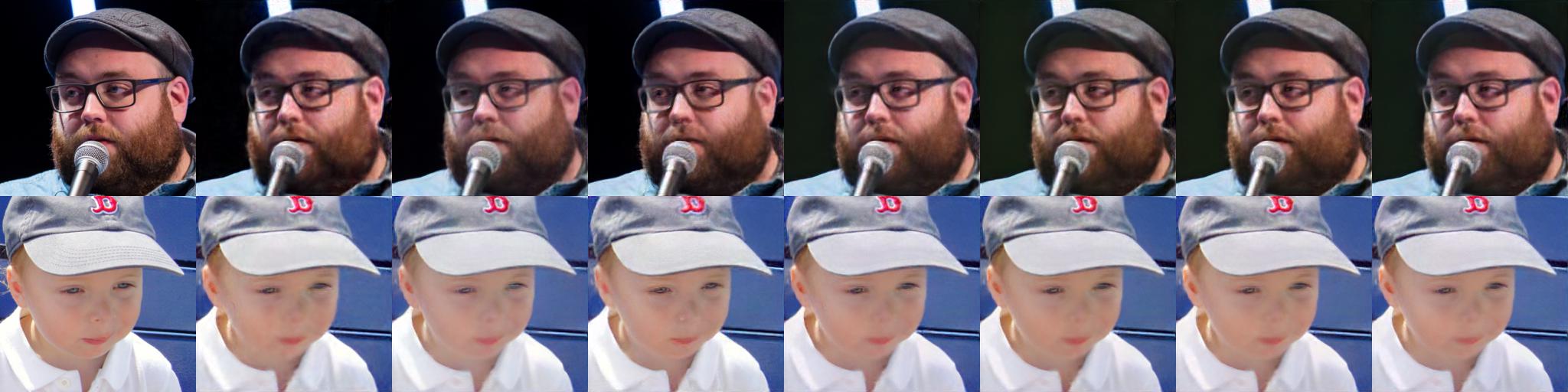} \styleganfigure{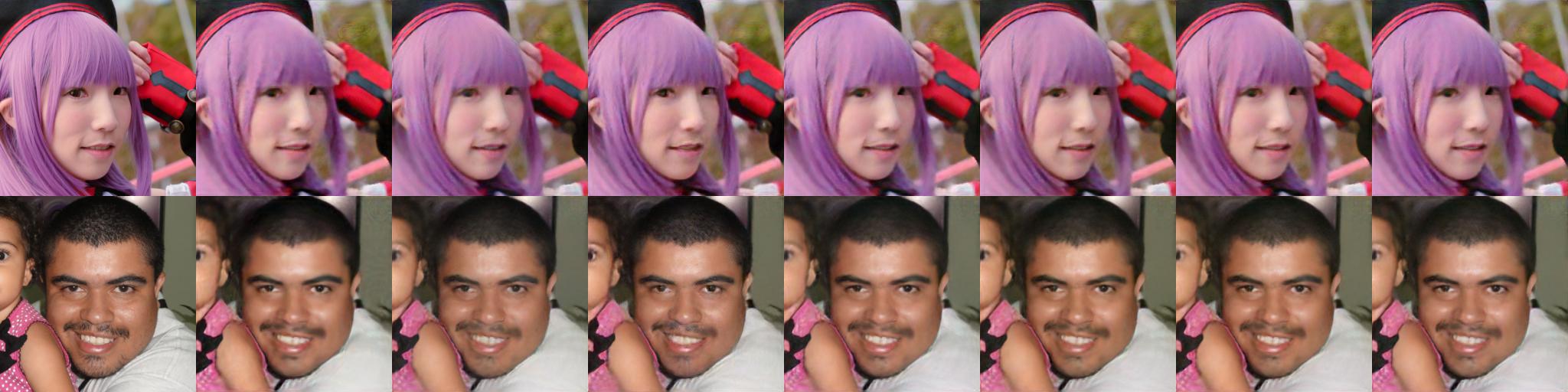} \styleganfigure{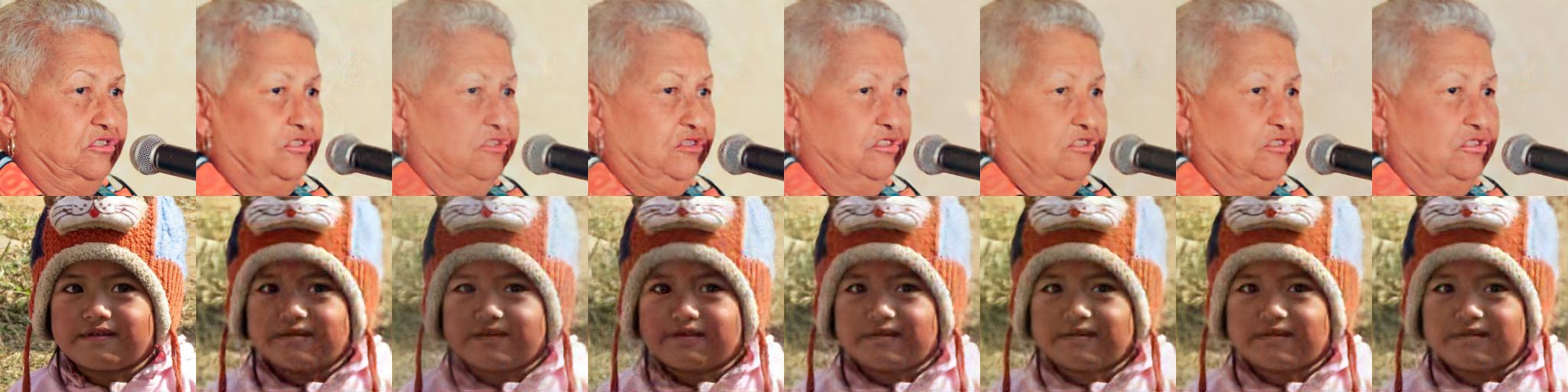} \styleganfigure{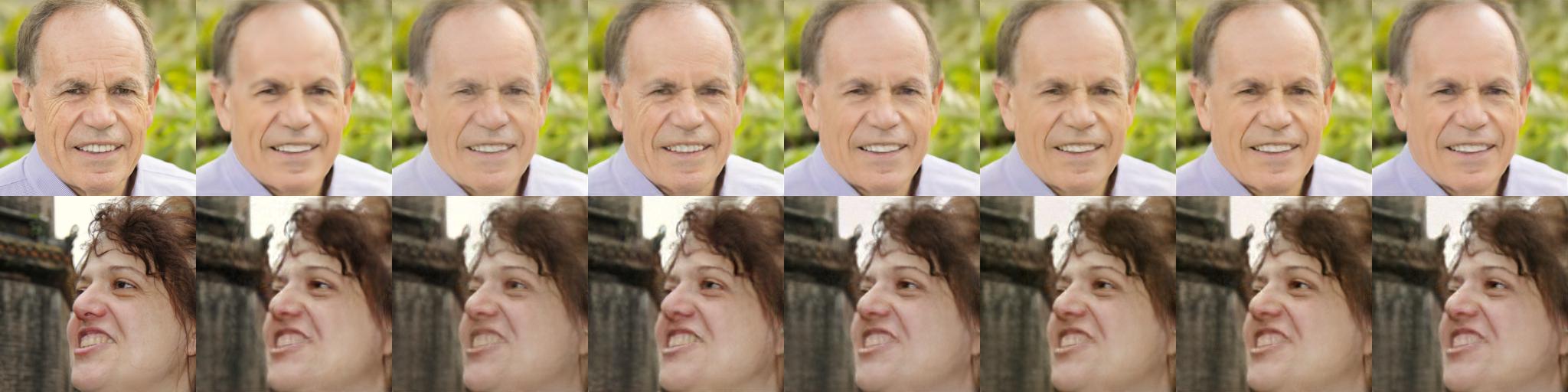} \stylegancaption \end{figure*}

\newcommand{\avatartrainfigure}[1]{%
\begin{center}
    \includegraphics[width=0.72\linewidth]{figures/model_comparison/supmat_2/03_avatars/01_train/#1.jpg}
    \begin{tabular}{C{0.066\linewidth}C{0.066\linewidth}C{0.066\linewidth}C{0.066\linewidth}C{0.066\linewidth}C{0.066\linewidth}C{0.066\linewidth}C{0.066\linewidth}}
        \footnotesize{Gr. truth} & \footnotesize{MSE} & \footnotesize{VGG-19} & \footnotesize{VGG-11} & \footnotesize{ResNet-4} & \footnotesize{ResNet-6} & \footnotesize{ResNet-8} & \footnotesize{UNet} \\
\end{tabular}
\end{center}%
}

\newcommand{\avatartraincaption}{%
\caption{Results of fine-tuning of neural talking head models~\cite{Zakharov19} with different losses. VGG-based perceptual loss is optimized without supplementing it with MSE loss. For PGN models, we use $\lambda_{\operatorname{MSE}} = 20$, similarly to the experiments with StyleGAN inference. Here, we show how the generator reproduces the same images as the ones it was fine-tuned on (``train''). In each row, we show the first of 8 ``train'' images for a particular person. Zoom in to see finer details.}%
}

\begin{figure*}[p] \avatartrainfigure{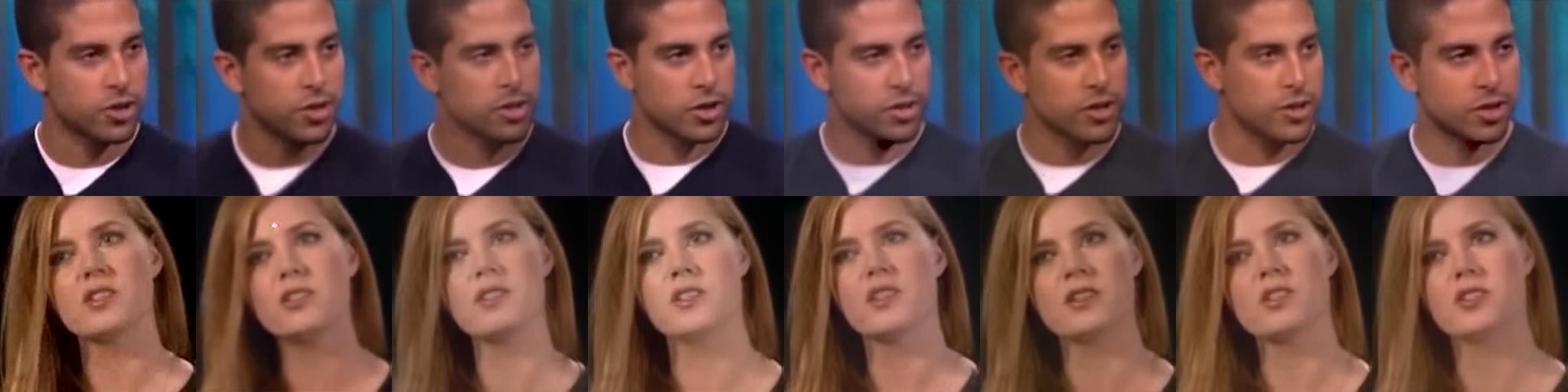} \avatartrainfigure{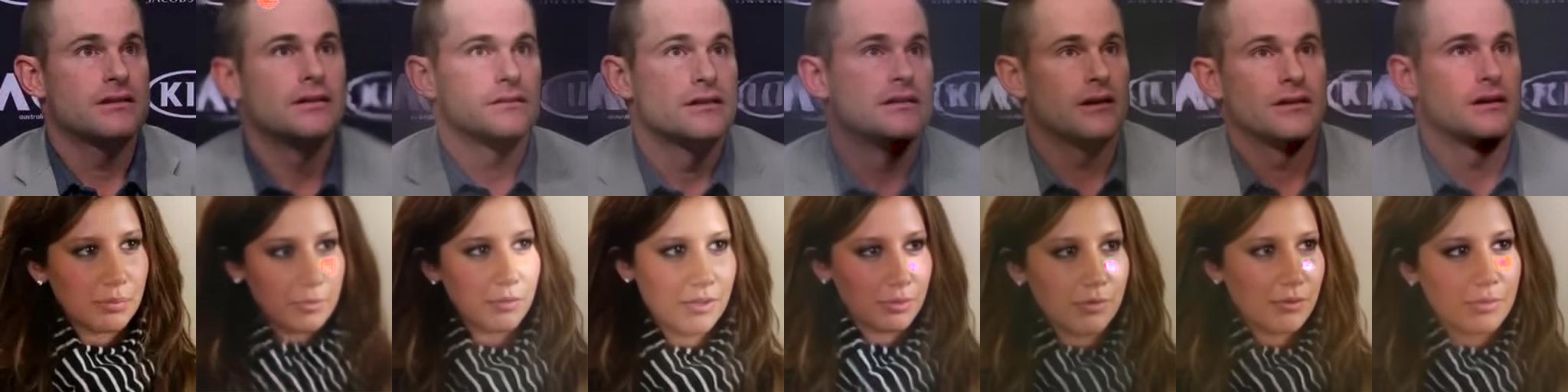} \avatartrainfigure{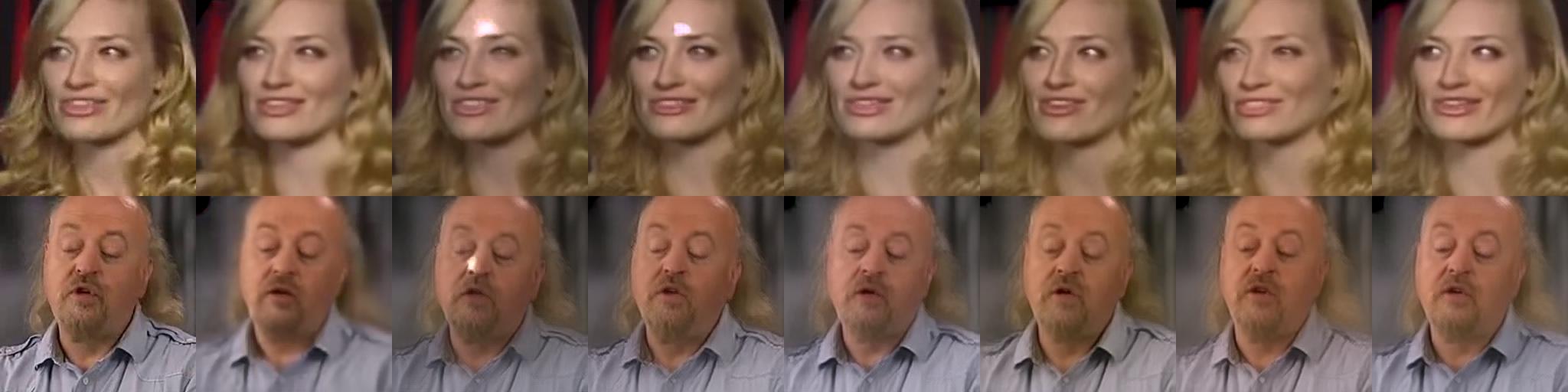} \avatartrainfigure{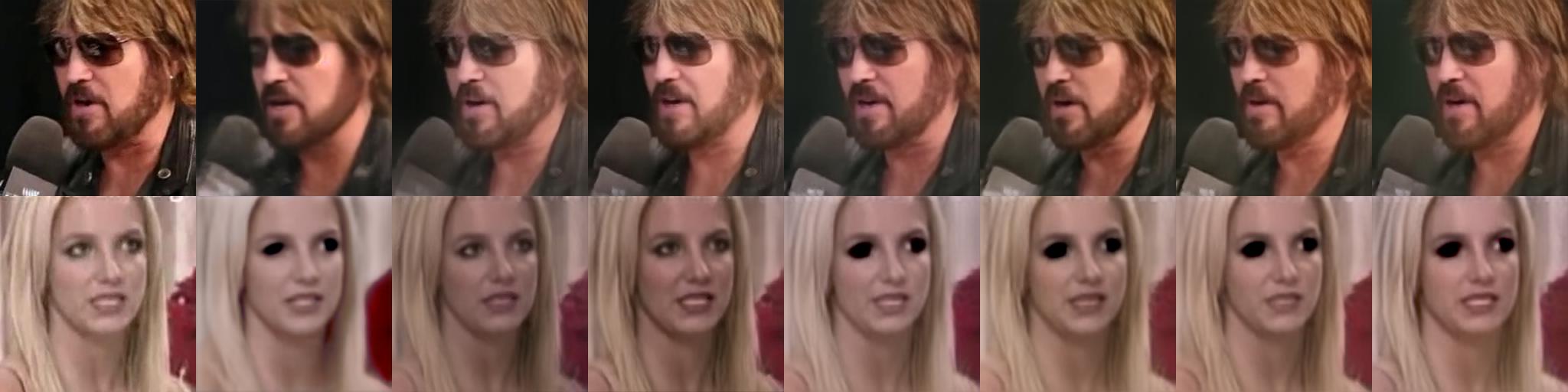} \avatartrainfigure{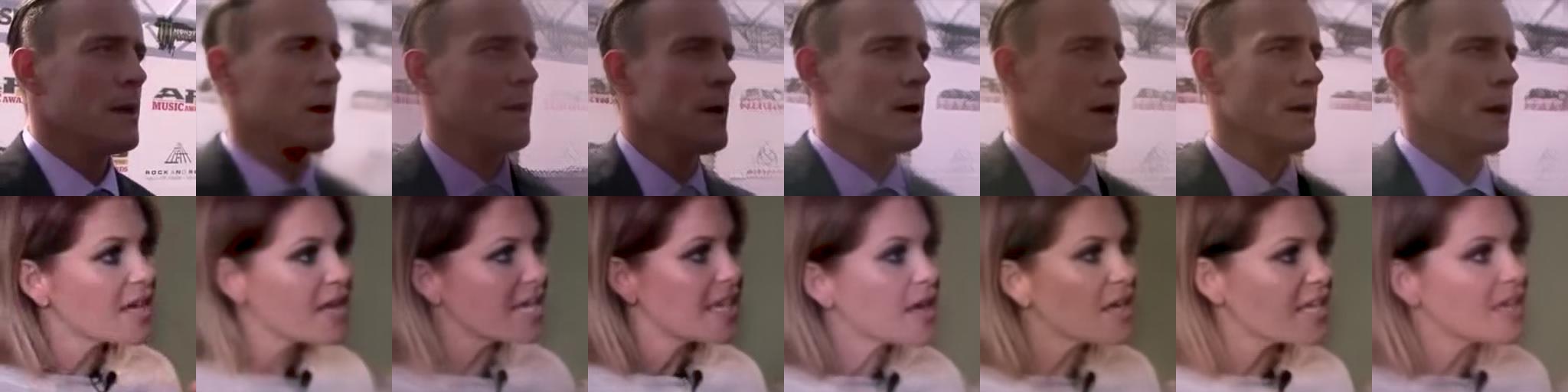} \avatartraincaption \label{fig:avatartrain} \end{figure*}

\newcommand{\avatartestfigure}[1]{%
\begin{center}
    \includegraphics[width=0.72\linewidth]{figures/model_comparison/supmat_2/03_avatars/02_test/#1.jpg}
    \begin{tabular}{C{0.066\linewidth}C{0.066\linewidth}C{0.066\linewidth}C{0.066\linewidth}C{0.066\linewidth}C{0.066\linewidth}C{0.066\linewidth}C{0.066\linewidth}}
        \footnotesize{Gr. truth} & \footnotesize{MSE} & \footnotesize{VGG-19} & \footnotesize{VGG-11} & \footnotesize{ResNet-4} & \footnotesize{ResNet-6} & \footnotesize{ResNet-8} & \footnotesize{UNet} \\
\end{tabular}
\end{center}%
}

\newcommand{\avatartestcaption}{%
\caption{Results of fine-tuning of neural talking head models~\cite{Zakharov19} with different losses. VGG-based perceptual loss is optimized without supplementing it with MSE loss. For PGN models, we use $\lambda_{\operatorname{MSE}} = 20$, similarly to the experiments with StyleGAN inference. Here, we show how the generator reproduces images it has not been fine-tuned on (``holdout''). In each row, we show the first of 8 ``holdout'' images for a particular person. Zoom in to see finer details.}%
}

\begin{figure*}[p] \avatartestfigure{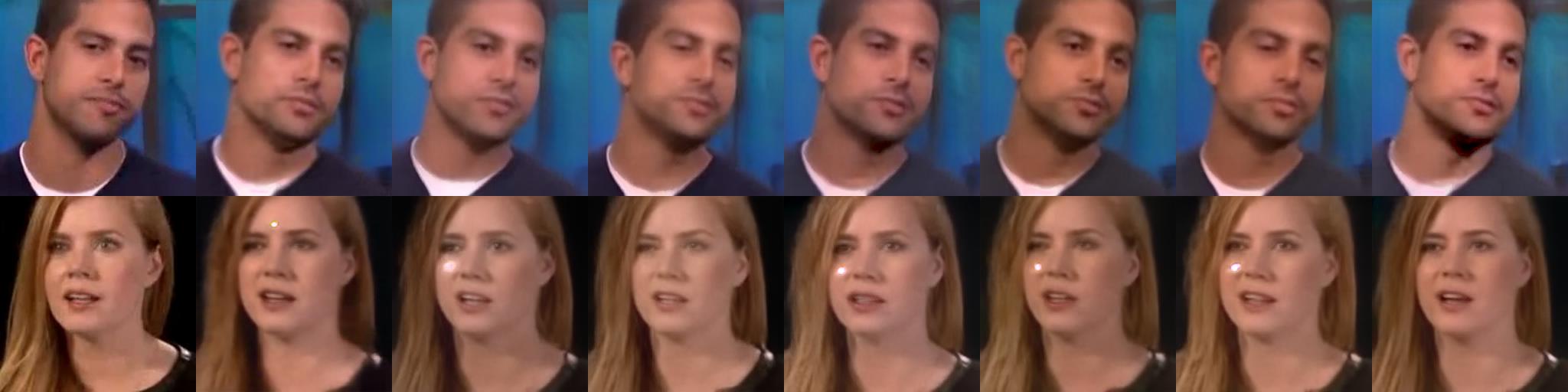} \avatartestfigure{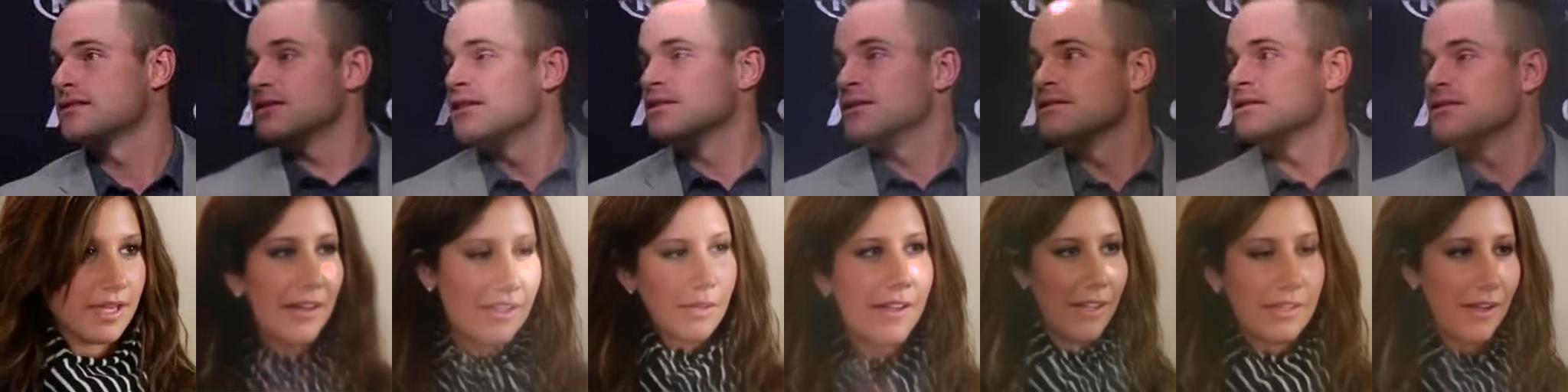} \avatartestfigure{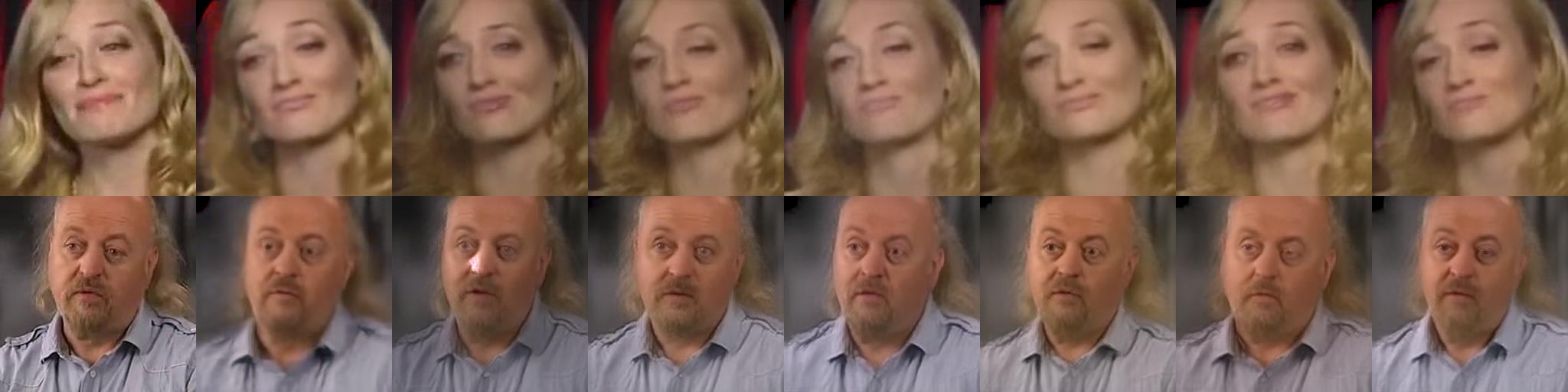} \avatartestfigure{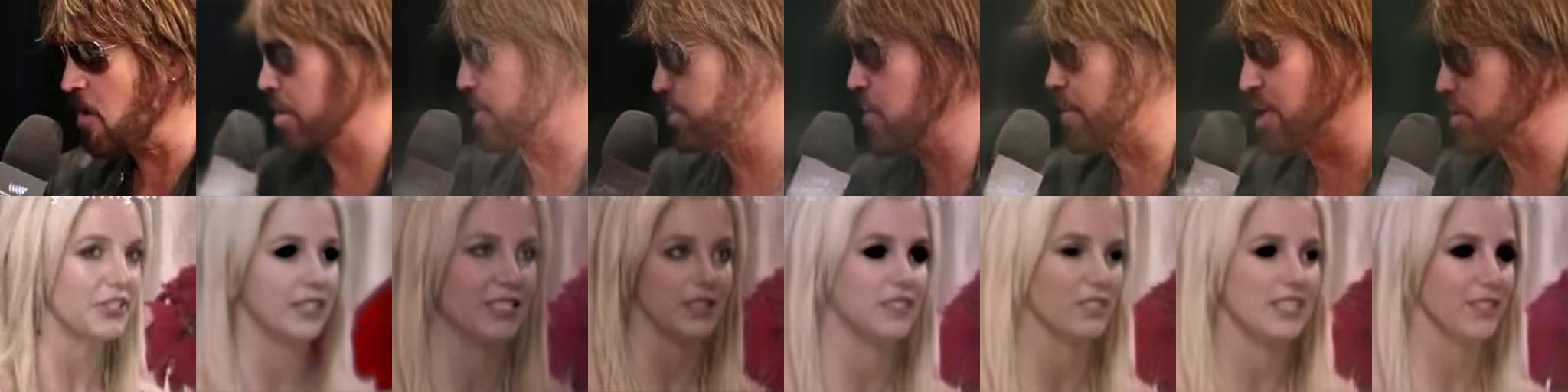} \avatartestfigure{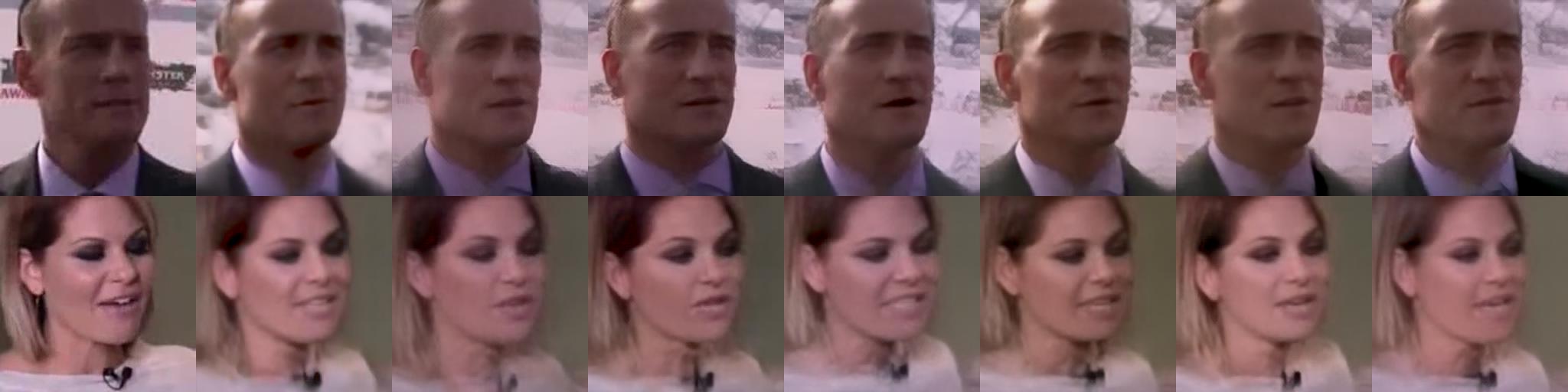} \avatartestcaption \label{fig:avatartest} \end{figure*}
\begin{figure*}[p]\ContinuedFloat \avatartestfigure{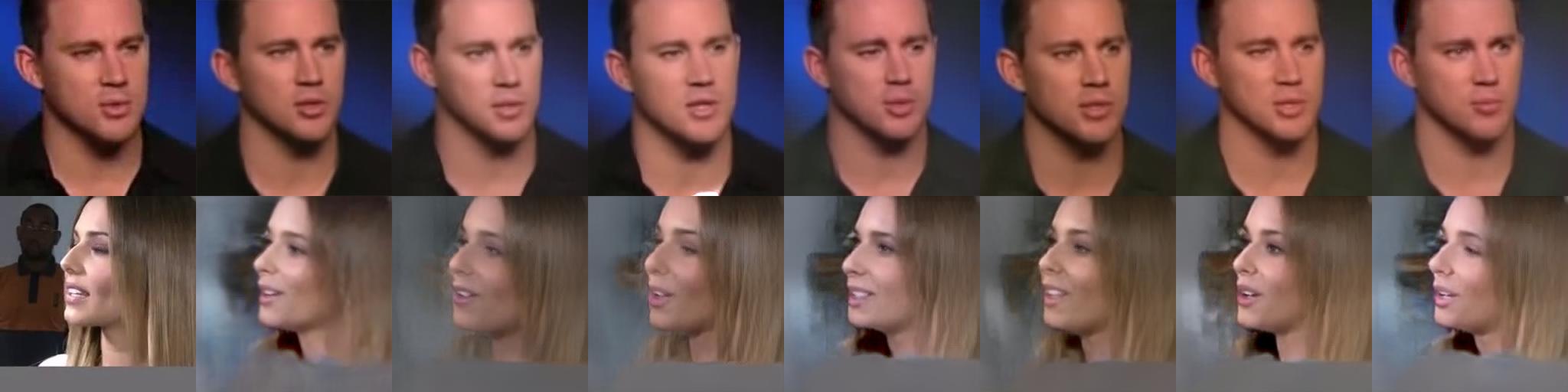} \avatartestfigure{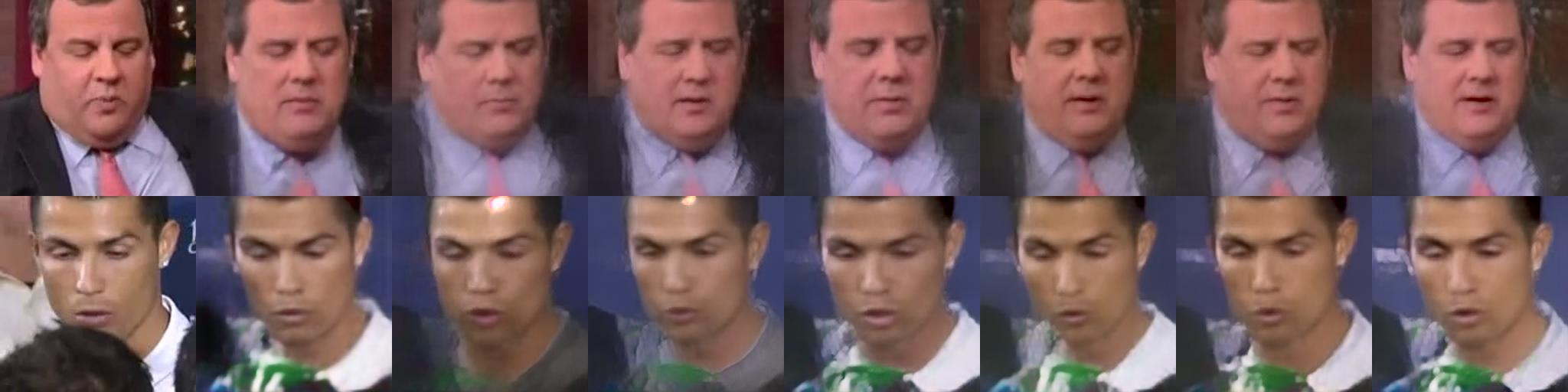} \avatartestfigure{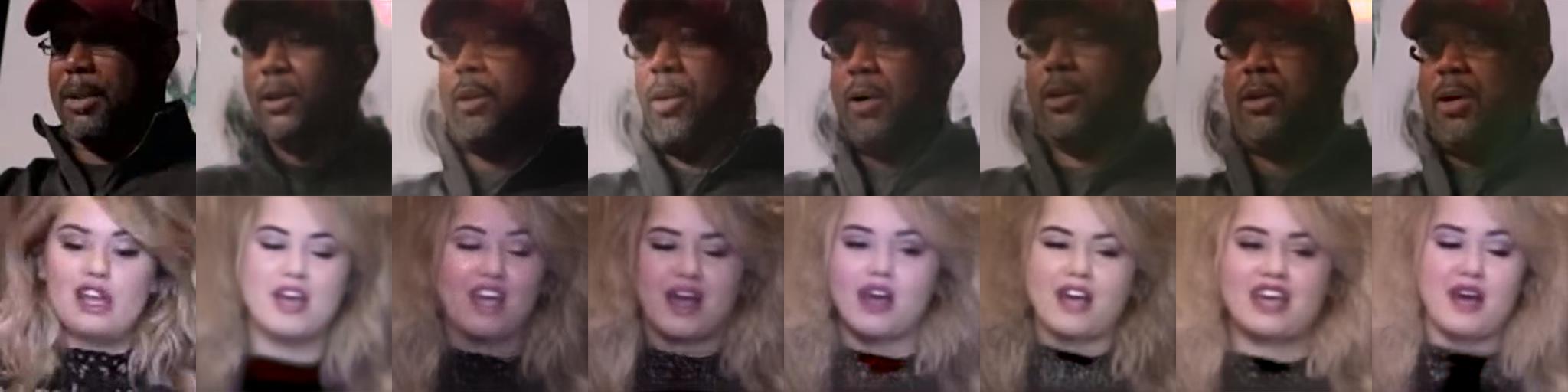} \avatartestfigure{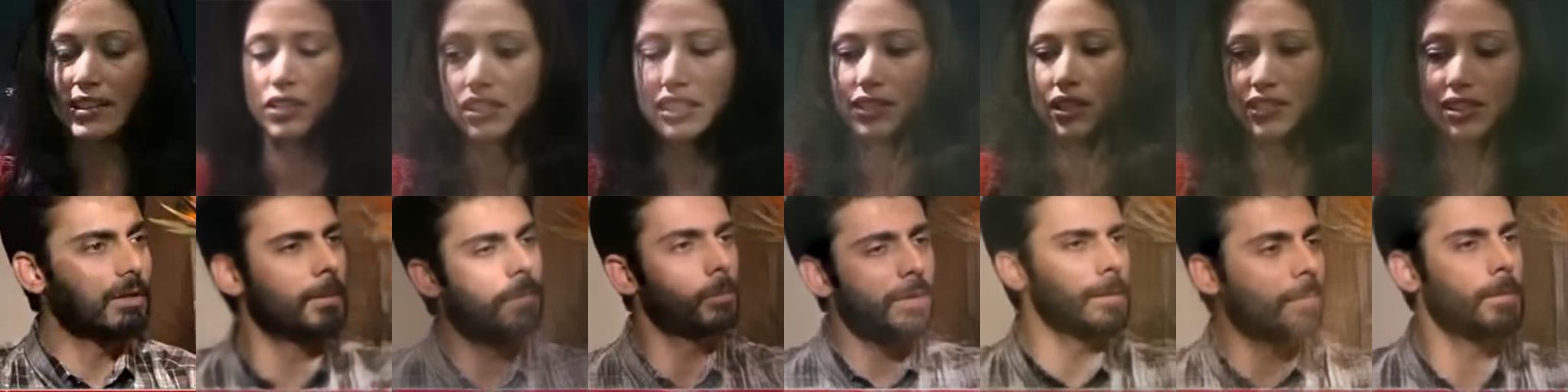} \avatartestfigure{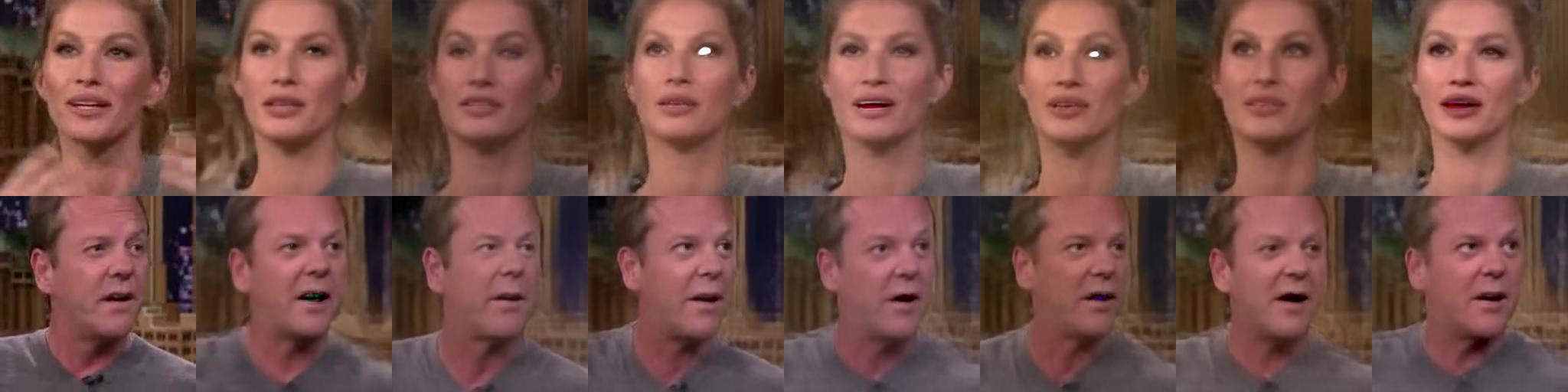} \avatartestcaption \end{figure*}

\newcommand{\proxyfigure}[1]{%
\includegraphics[height=0.15\textheight]{figures/model_comparison/supmat_2/04_proxy/#1.jpg}
\begin{tabular}{C{0.074\linewidth}C{0.074\linewidth}C{0.074\linewidth}C{0.074\linewidth}C{0.074\linewidth}C{0.074\linewidth}}
     \footnotesize{Gr. truth} & \footnotesize{MSE only} & \footnotesize{ResNet-4} & \footnotesize{ResNet-6} & \footnotesize{ResNet-8} & \footnotesize{UNet} \\
\end{tabular}%
}

\newcommand{\proxycaption}{%
\caption{Examples of proxy targets learned with different PGN backbones from Image2StyleGAN experiments. In these experiments, we pre-computed latent representations of images by optimizing only MSE loss for $T_0=5000$ steps. Here, the "MSE only" column refers to those images. Proxy targets are computed for $\hy = $ ``MSE only'', $y =$ ``gr. truth''. Note that while the output is very noisy, the proxies strongly highlight edges, suggesting that perceptual loss may be considered an elaborate edge detector. Also note that proxies learned by different models are very similar. Zoom in to see finer details.}%
}

\begin{figure*}[p] \centering \proxyfigure{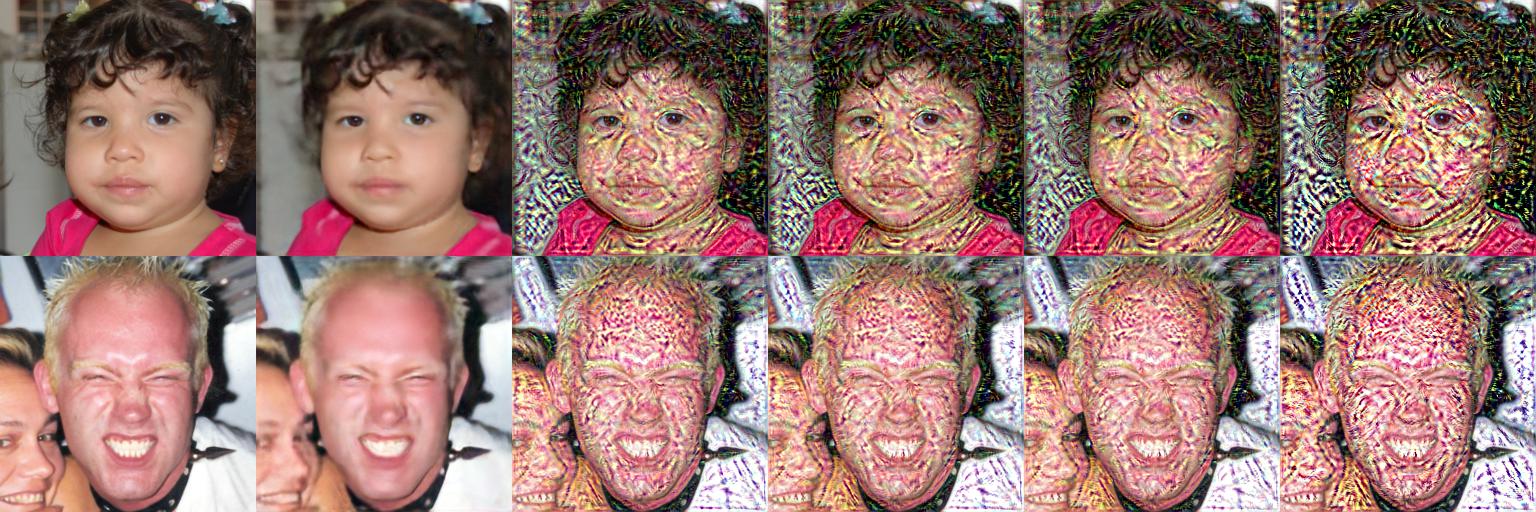} \proxyfigure{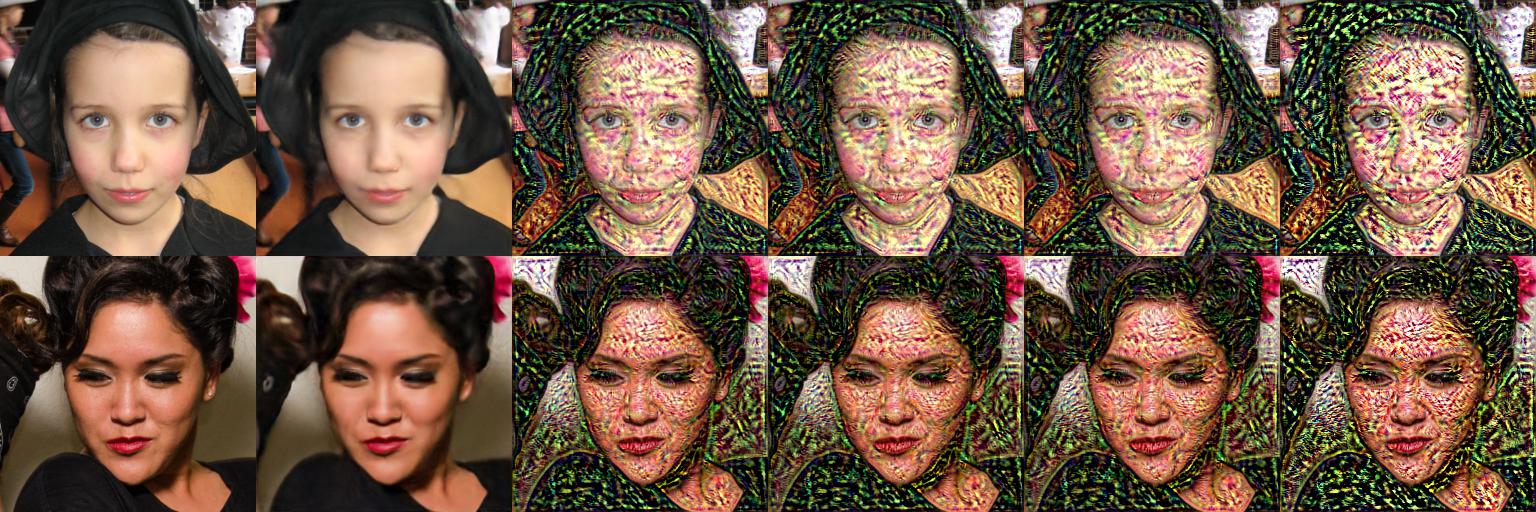} \proxyfigure{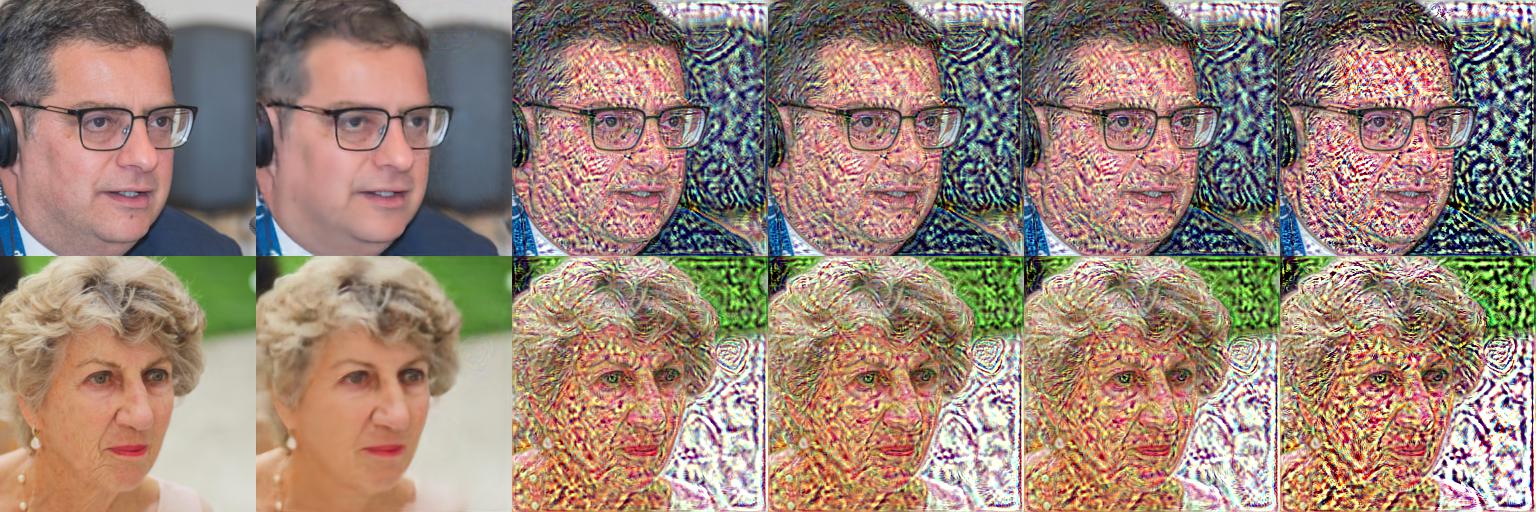} \proxyfigure{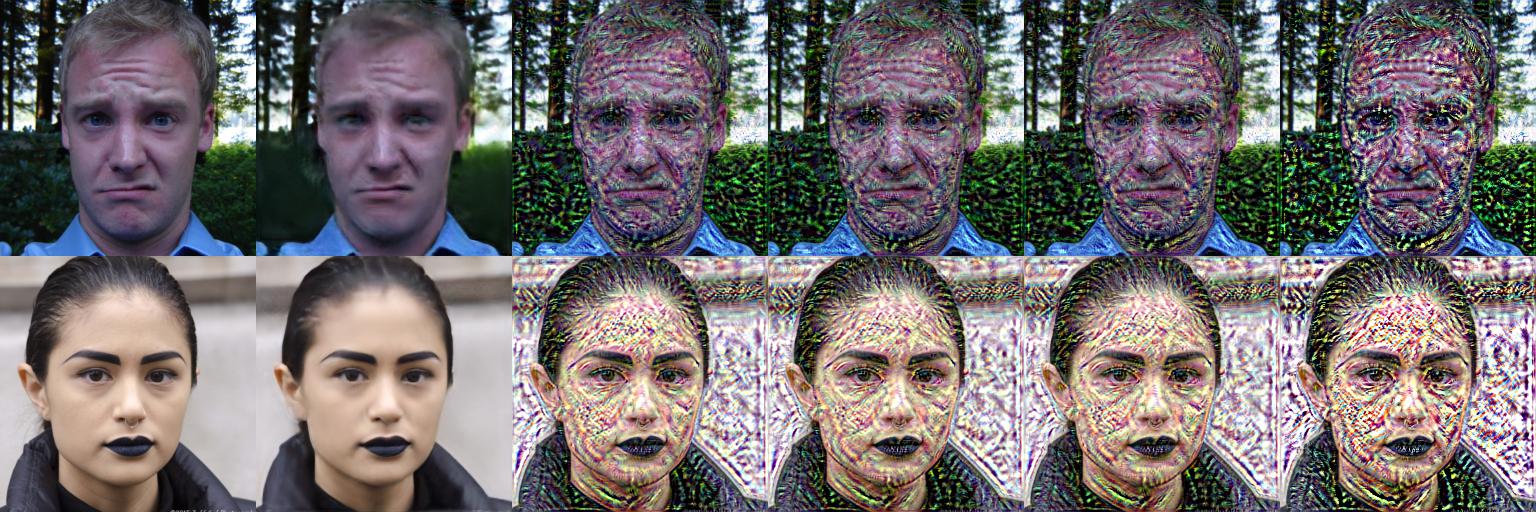} \proxyfigure{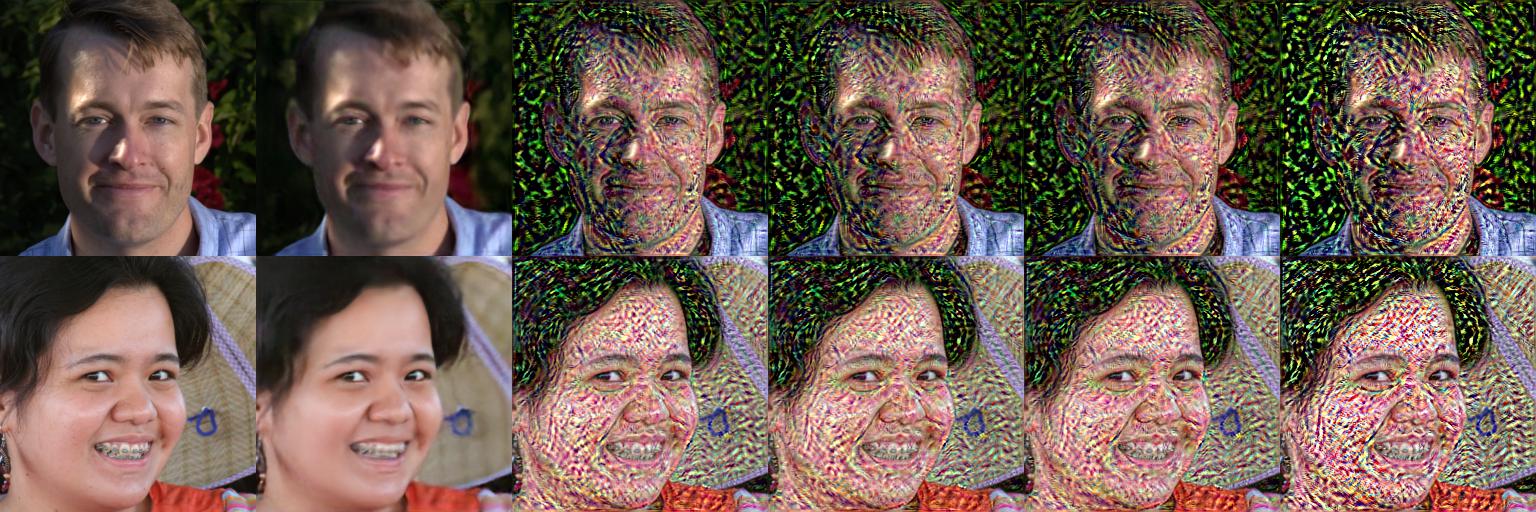} \proxycaption \label{fig:proxy} \end{figure*}
\begin{figure*}[p]\ContinuedFloat \centering \proxyfigure{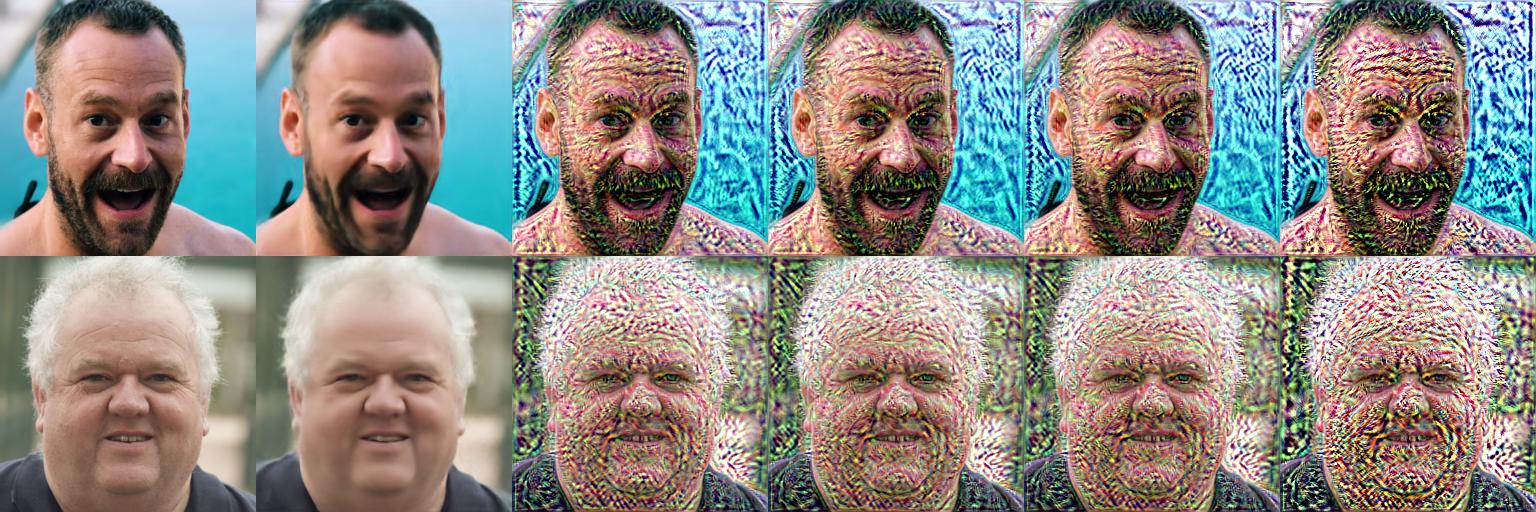} \proxyfigure{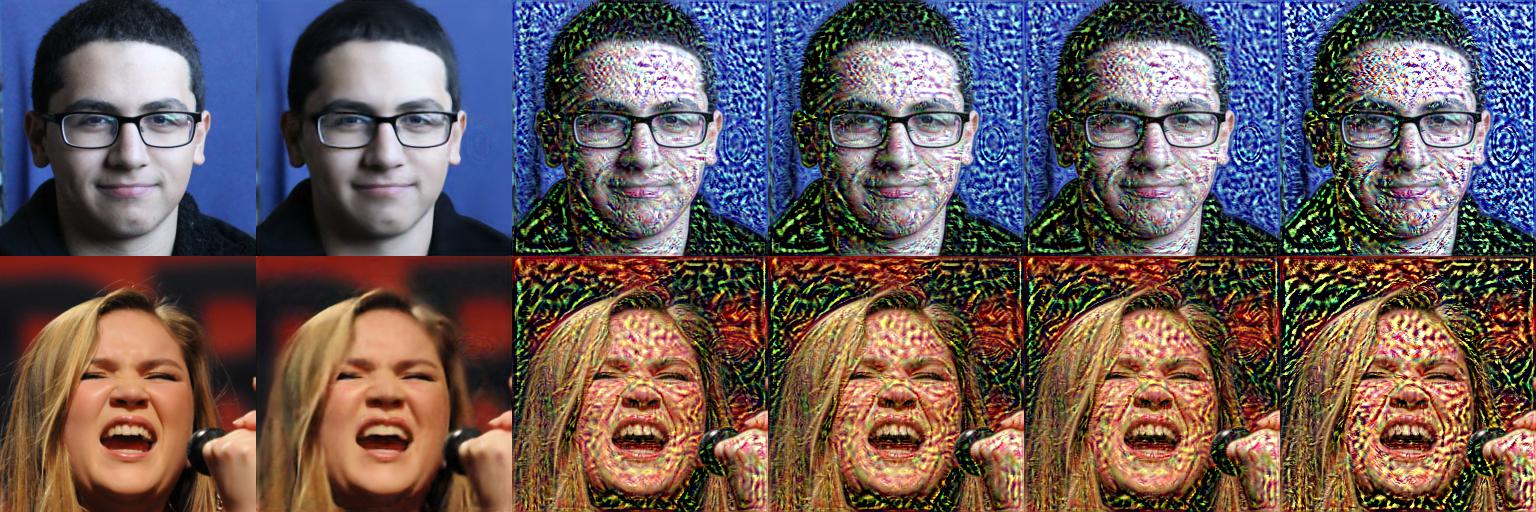} \proxyfigure{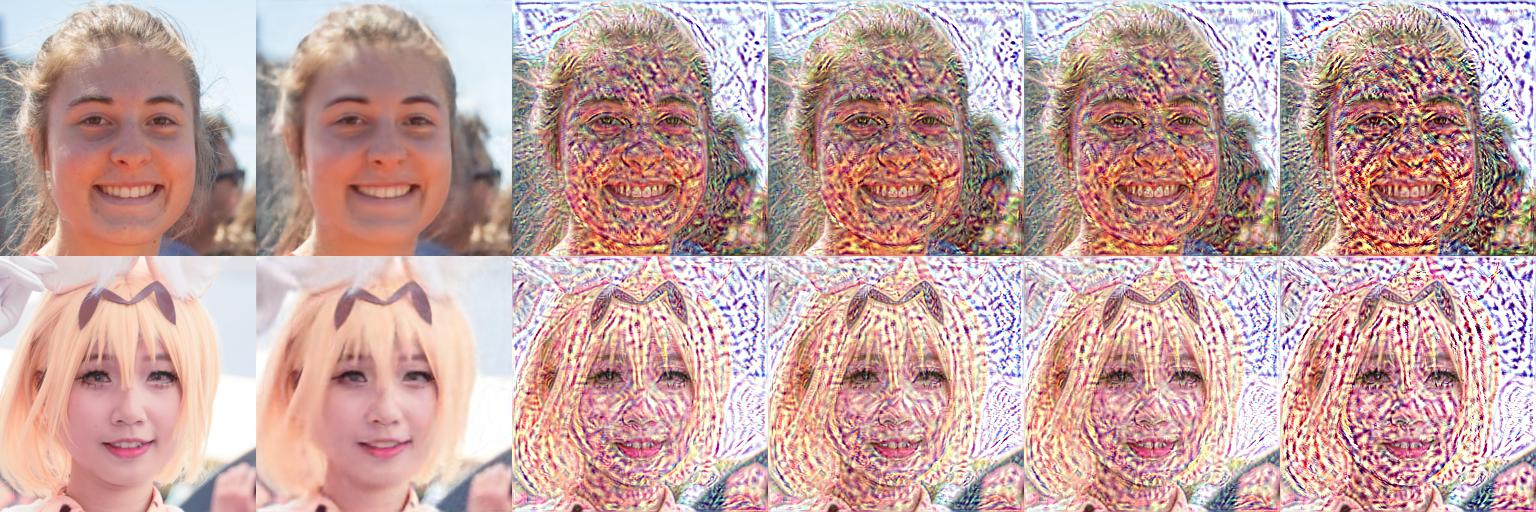} \proxyfigure{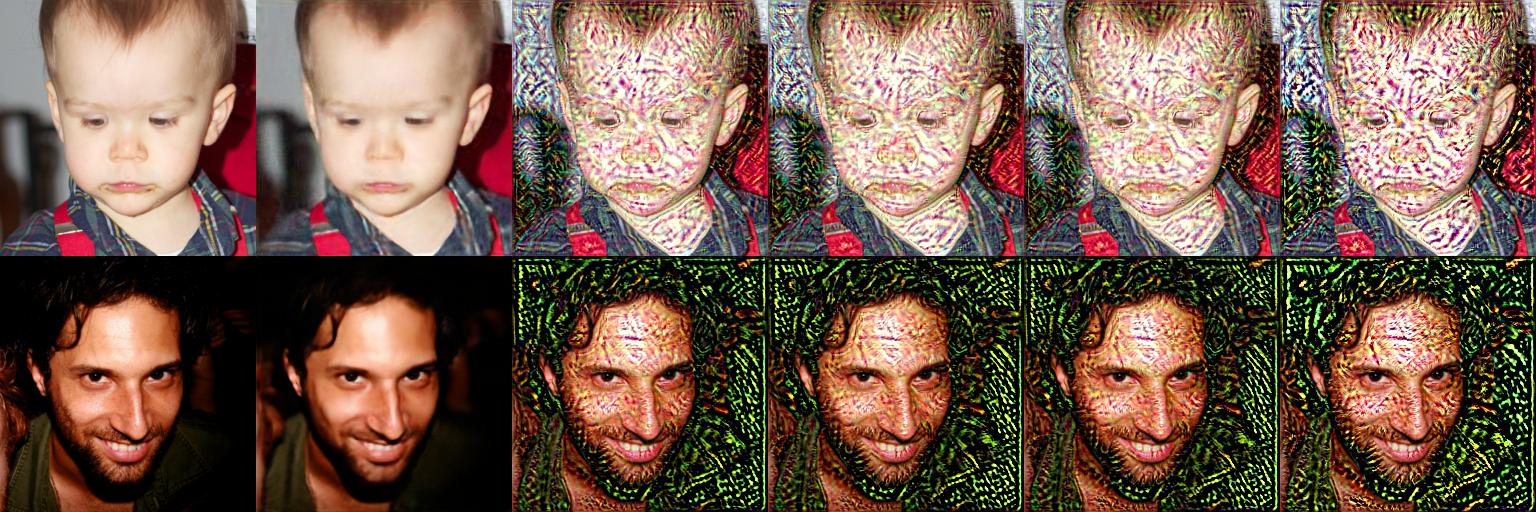} \proxyfigure{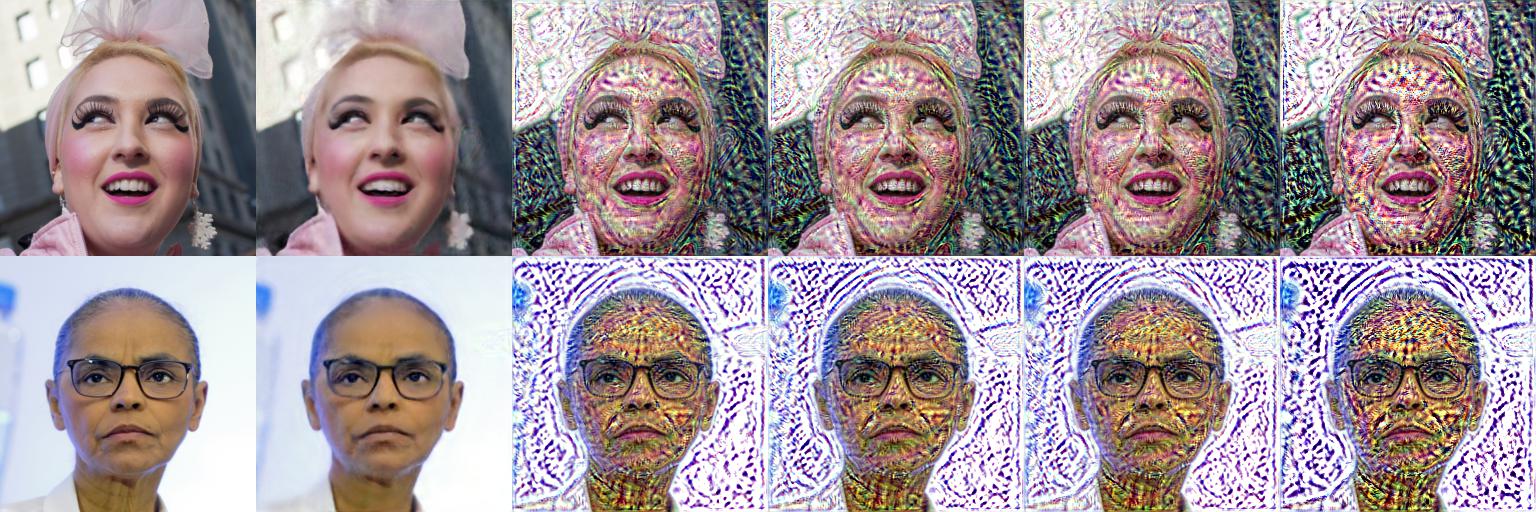} \proxycaption \end{figure*}
\begin{figure*}[p]\ContinuedFloat \centering \proxyfigure{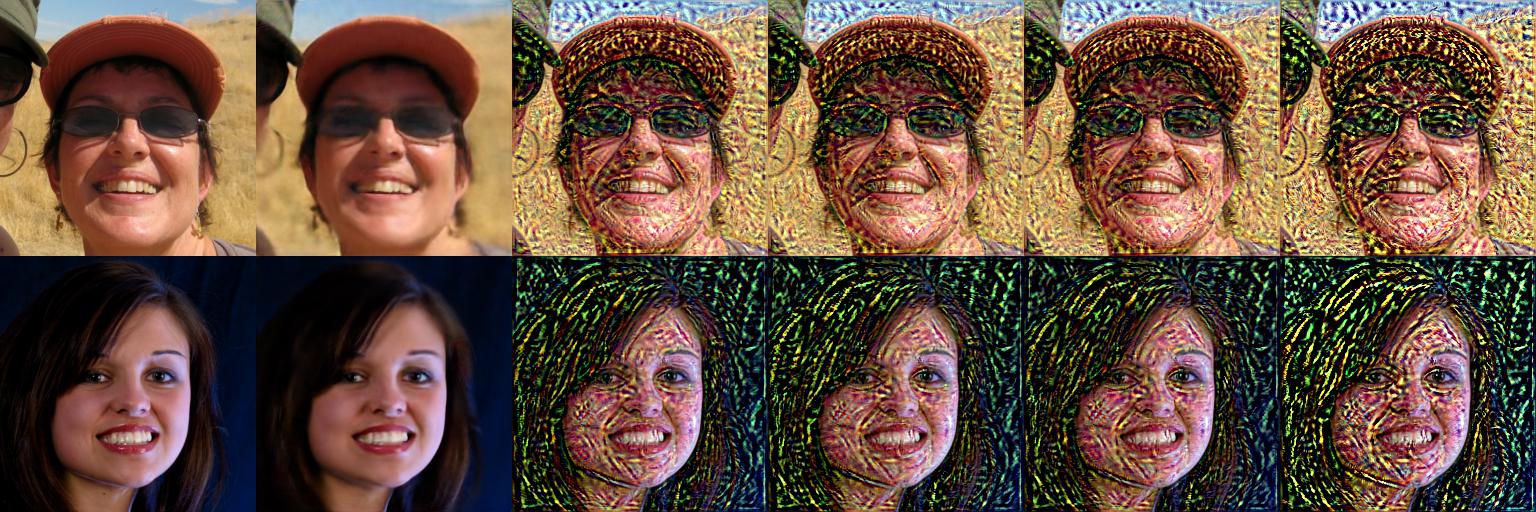} \proxyfigure{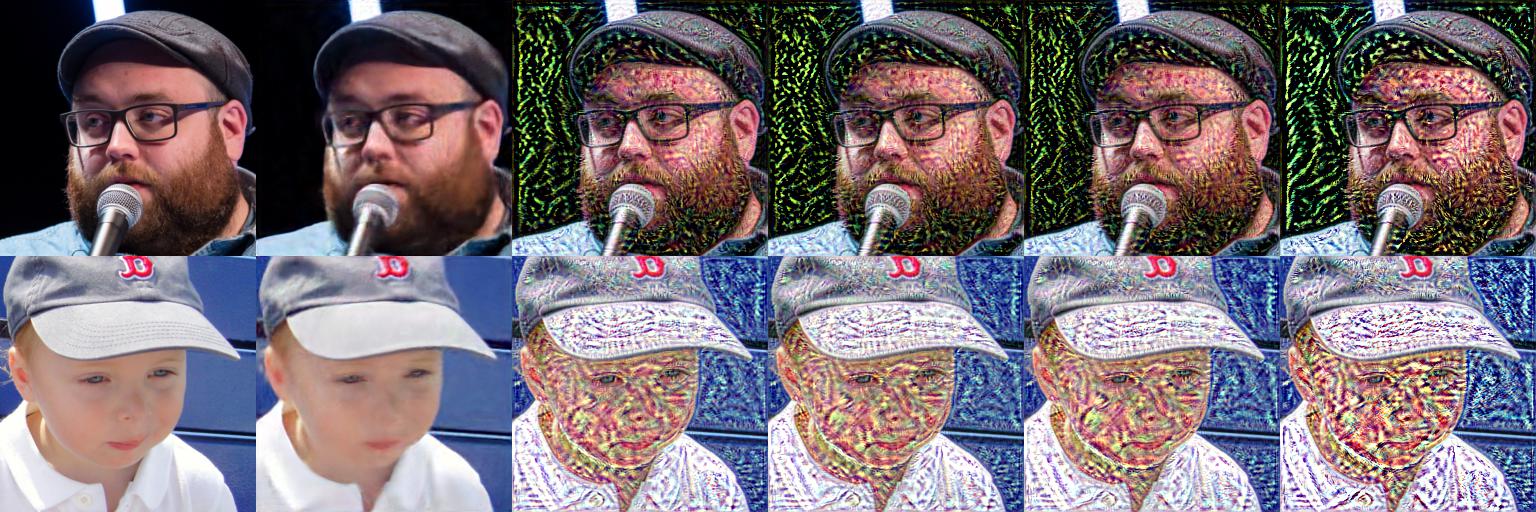} \proxyfigure{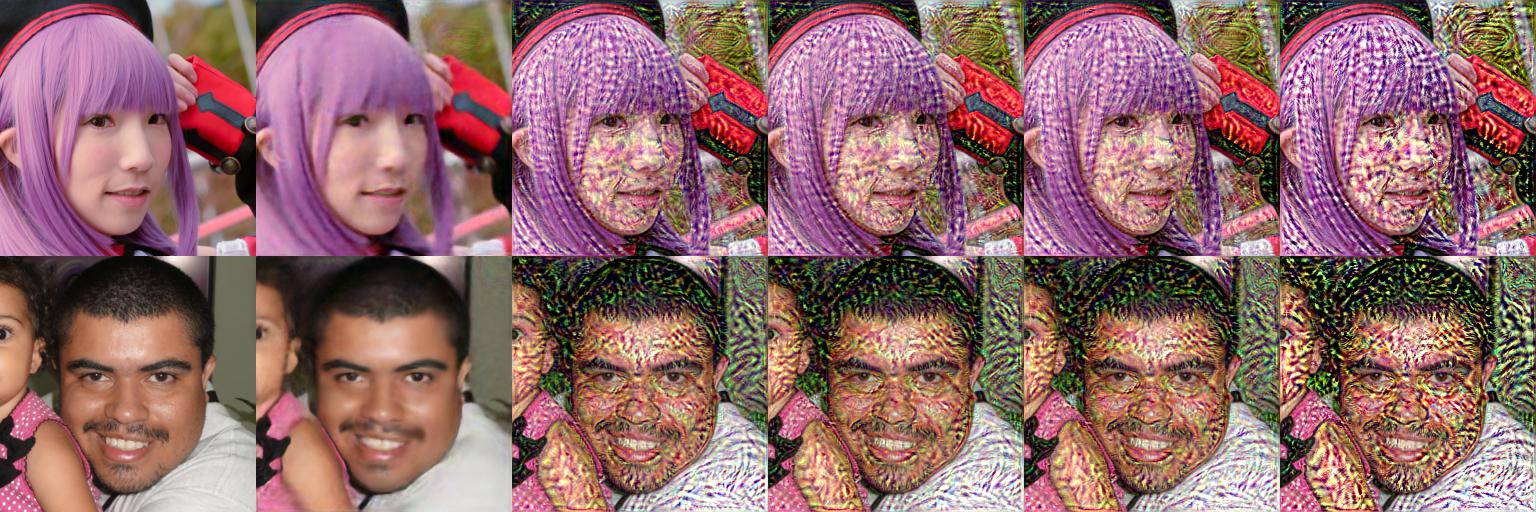} \proxyfigure{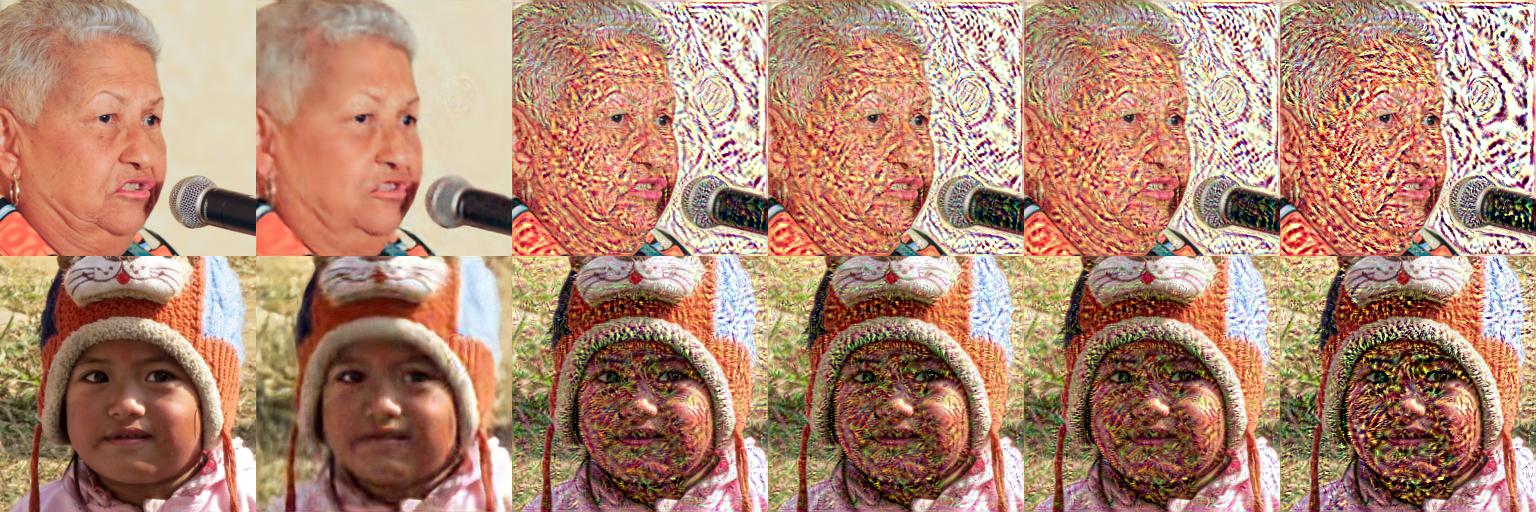} \proxyfigure{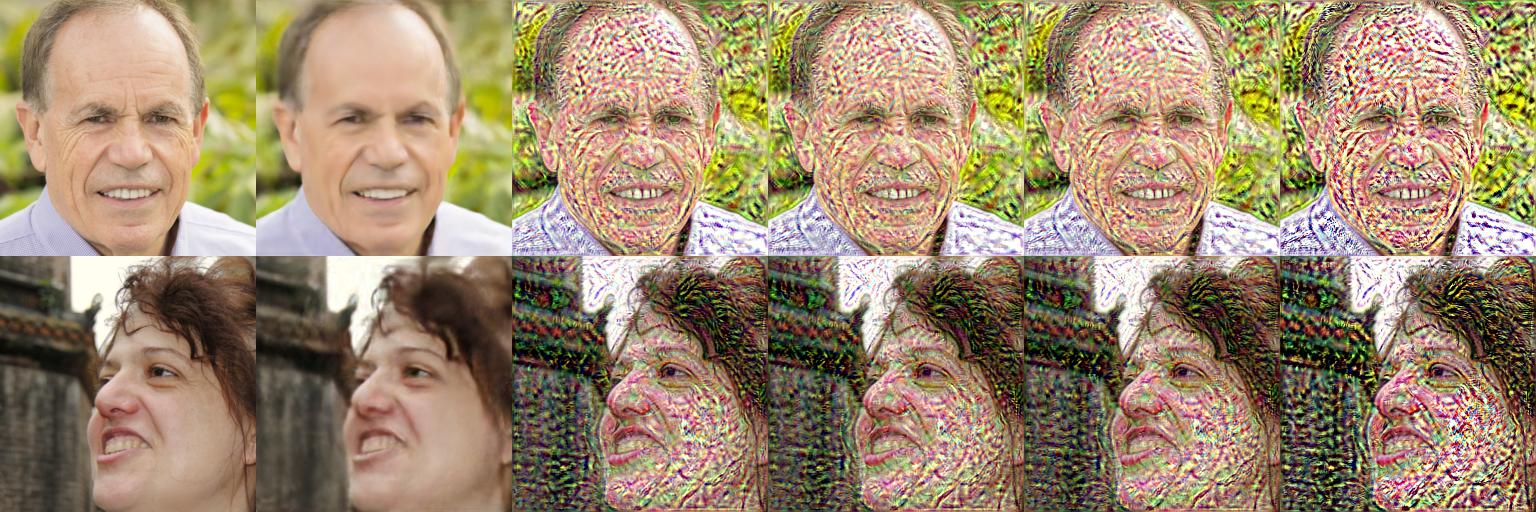} \proxycaption \end{figure*}

\newcommand{\styleganmsefigure}[1]{%
\includegraphics[height=0.15\textheight]{figures/model_comparison/supmat_2/05_stylegan_mse_comparison/#1.jpg}
{\setlength{\tabcolsep}{0mm}
\begin{tabular}{C{0.097\linewidth}C{0.097\linewidth}C{0.097\linewidth}C{0.097\linewidth}C{0.097\linewidth}C{0.097\linewidth}}
     \footnotesize{Gr. truth} & \footnotesize{$\lambda_{\operatorname{MSE}} = 0$} & \footnotesize{$\lambda_{\operatorname{MSE}} = 1$} & \footnotesize{$\lambda_{\operatorname{MSE}} = 5$} & \footnotesize{$\lambda_{\operatorname{MSE}} = 10$} & \footnotesize{$\lambda_{\operatorname{MSE}} = 20$} \\
\end{tabular}}%
}

\newcommand{\styleganmsecaption}{%
\caption{Examples of image reconstruction using Image2StyleGAN~\cite{Abdal19} by a PGN with a ResNet-4 backbone supplemented by MSE loss with different coefficients $\lambda_{\operatorname{MSE}}$. We used $\lambda_{\operatorname{MSE}} = 20$ to report all results in the paper. Zoom in to see finer details.}%
}

\begin{figure*}[p] \centering \styleganmsefigure{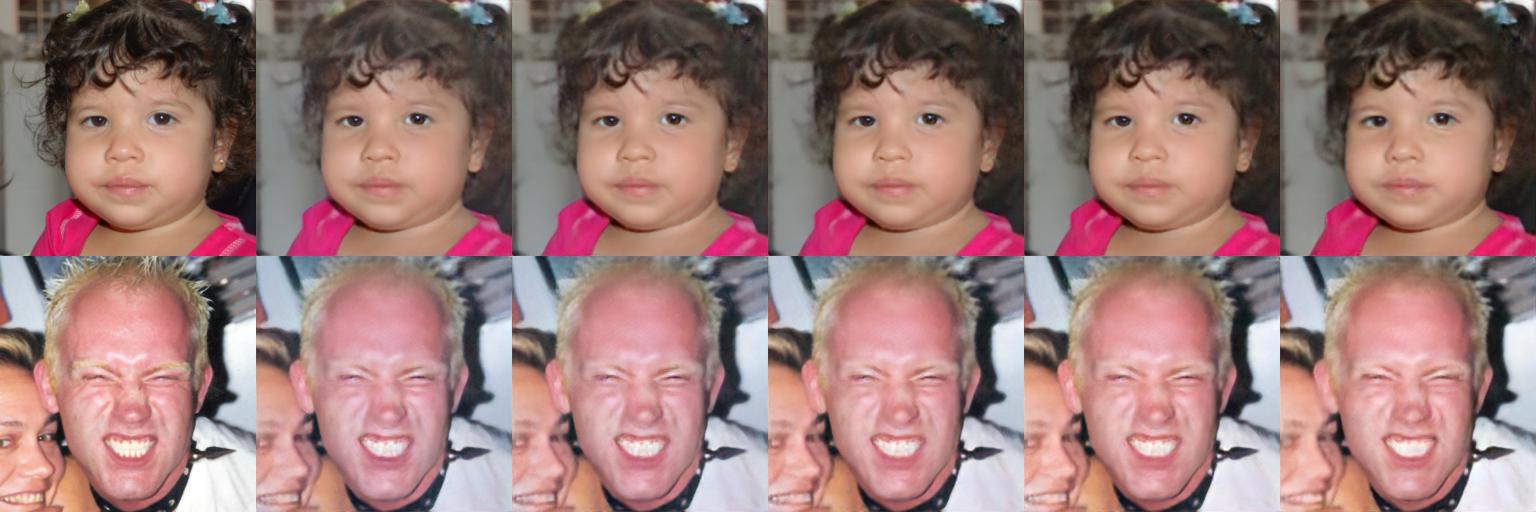} \styleganmsefigure{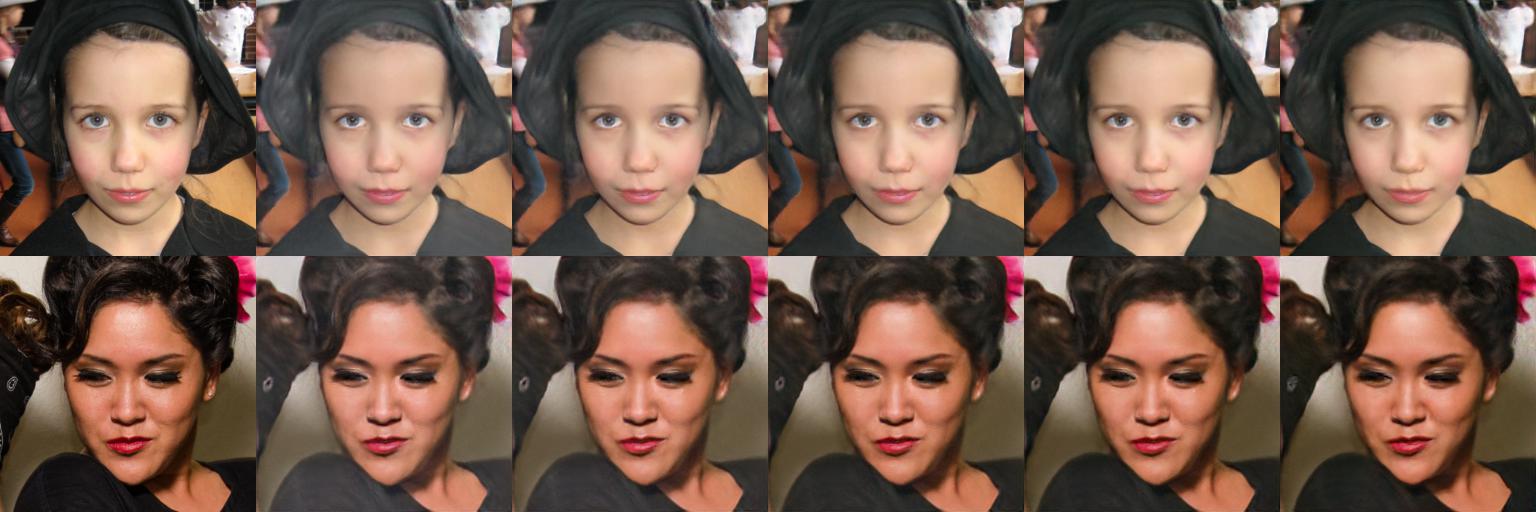} \styleganmsefigure{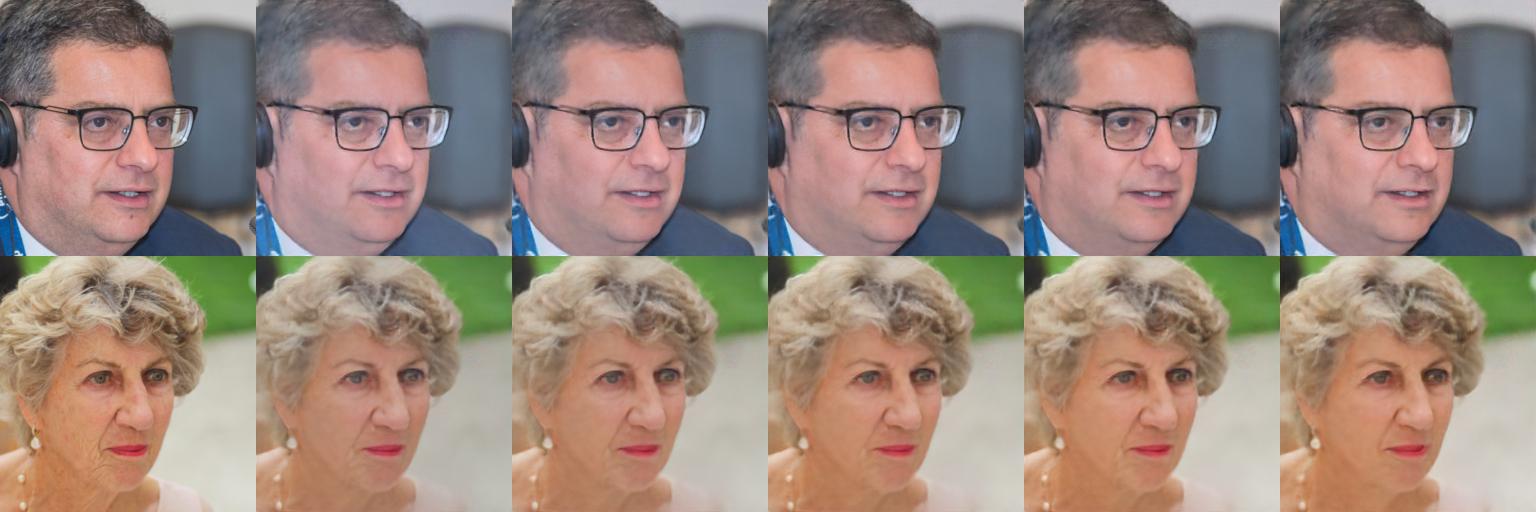} \styleganmsefigure{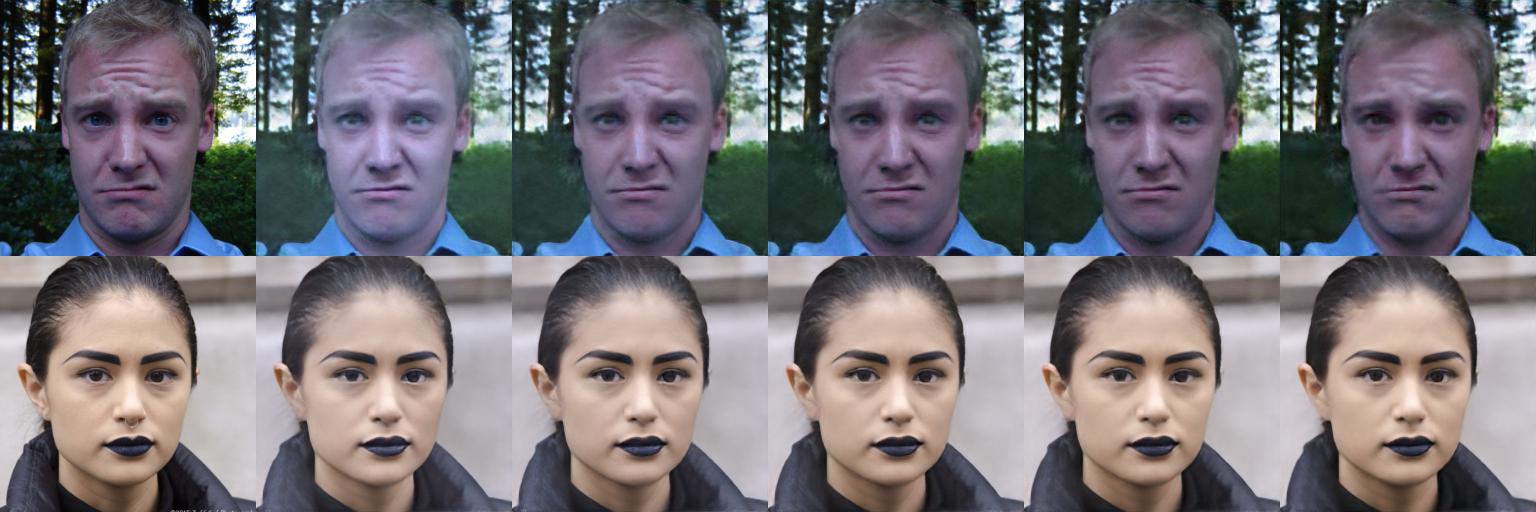} \styleganmsefigure{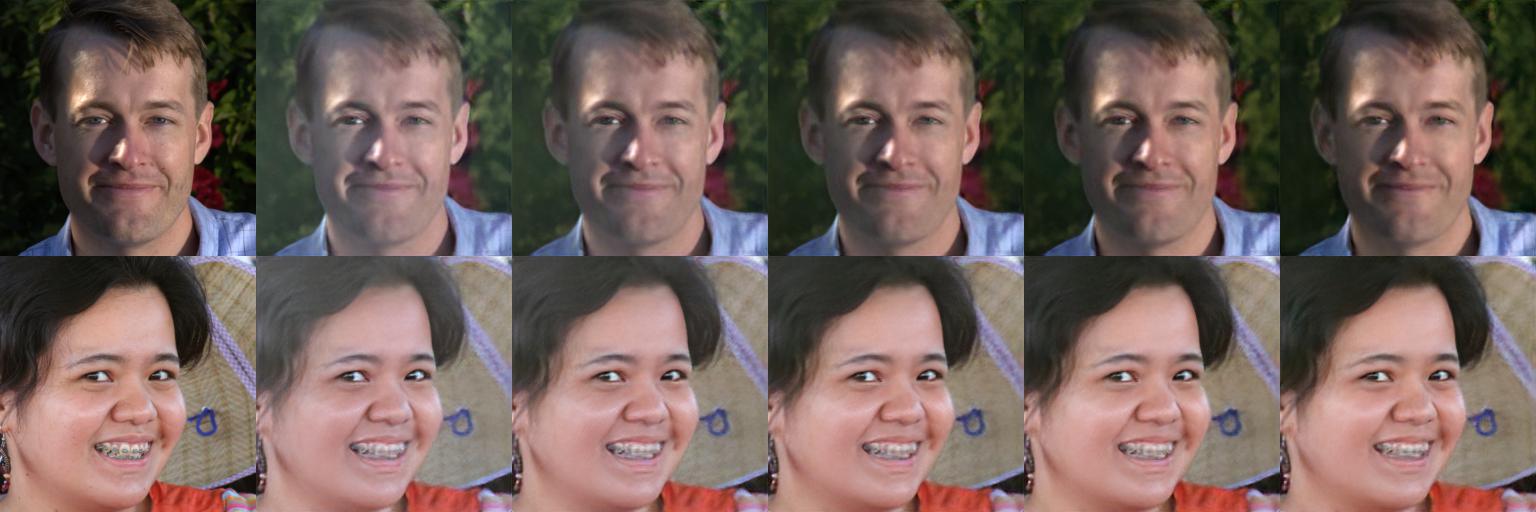} \styleganmsecaption \label{fig:styleganmse} \end{figure*}
\begin{figure*}[p]\ContinuedFloat \centering \styleganmsefigure{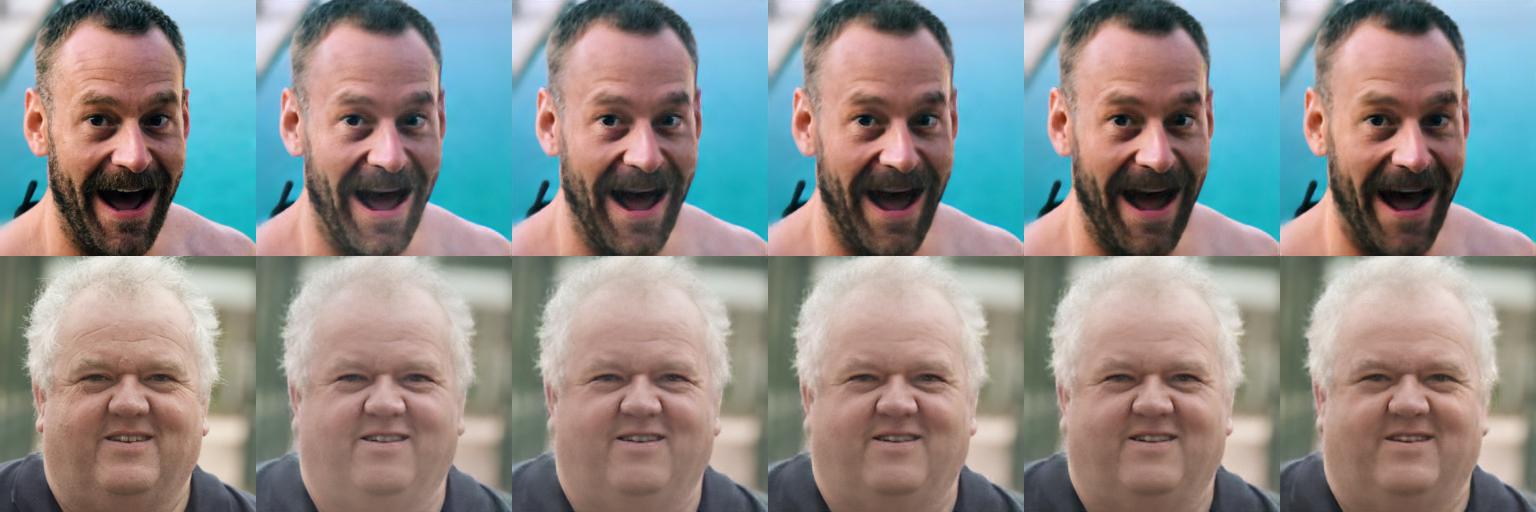} \styleganmsefigure{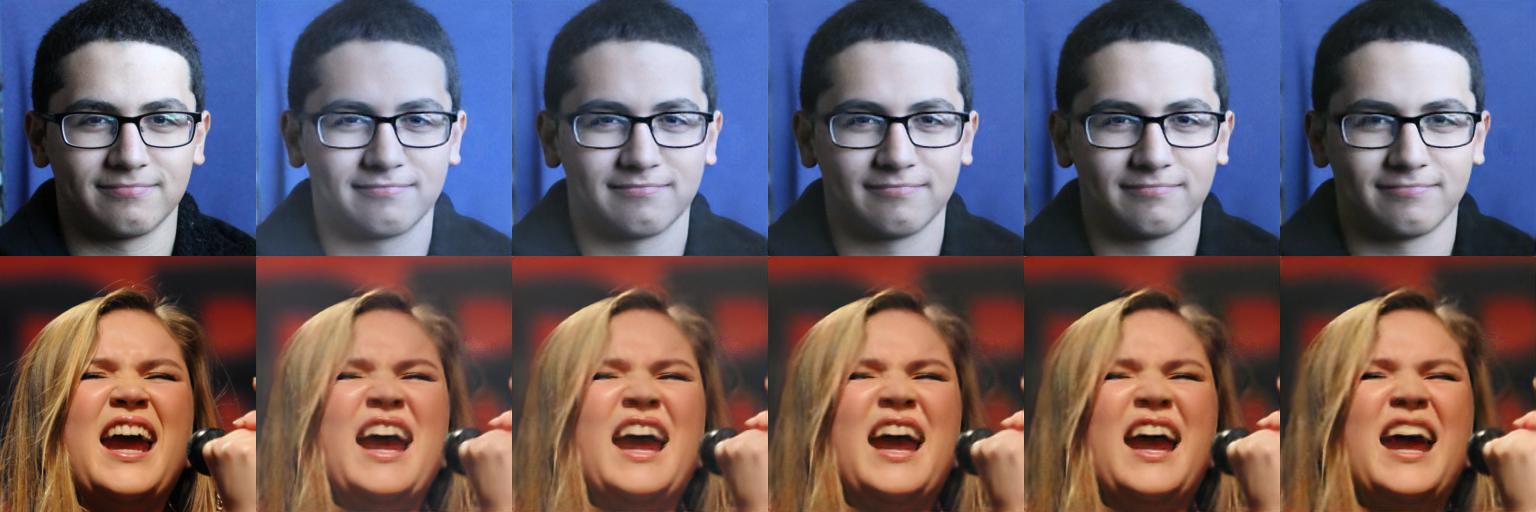} \styleganmsefigure{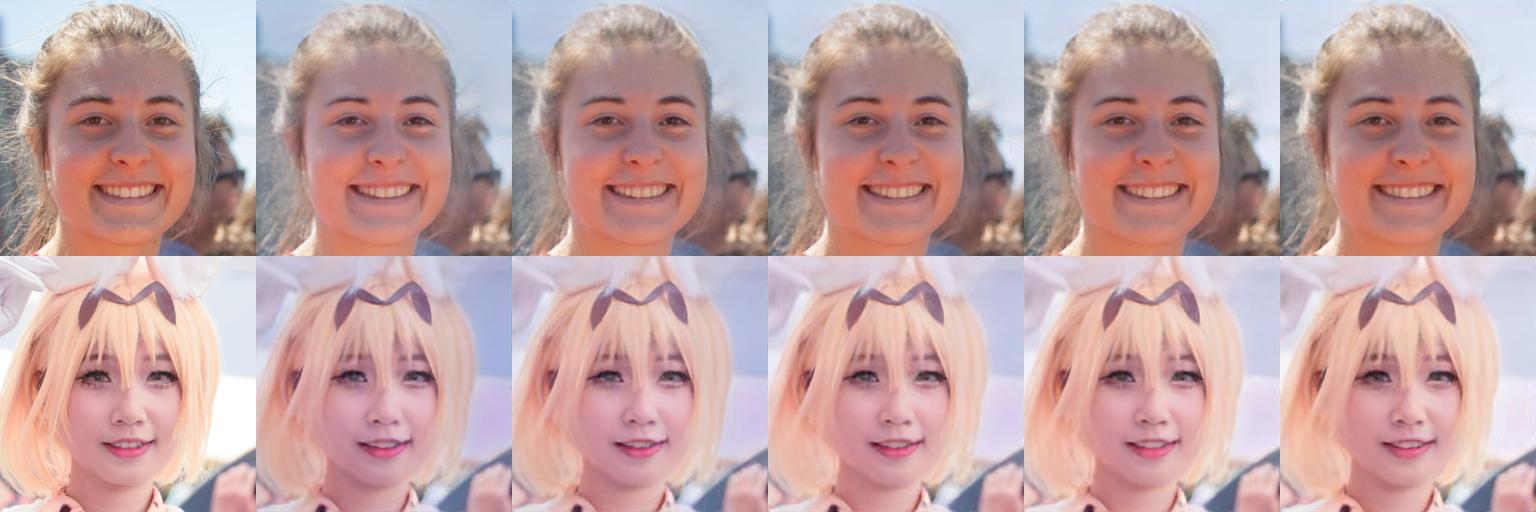} \styleganmsefigure{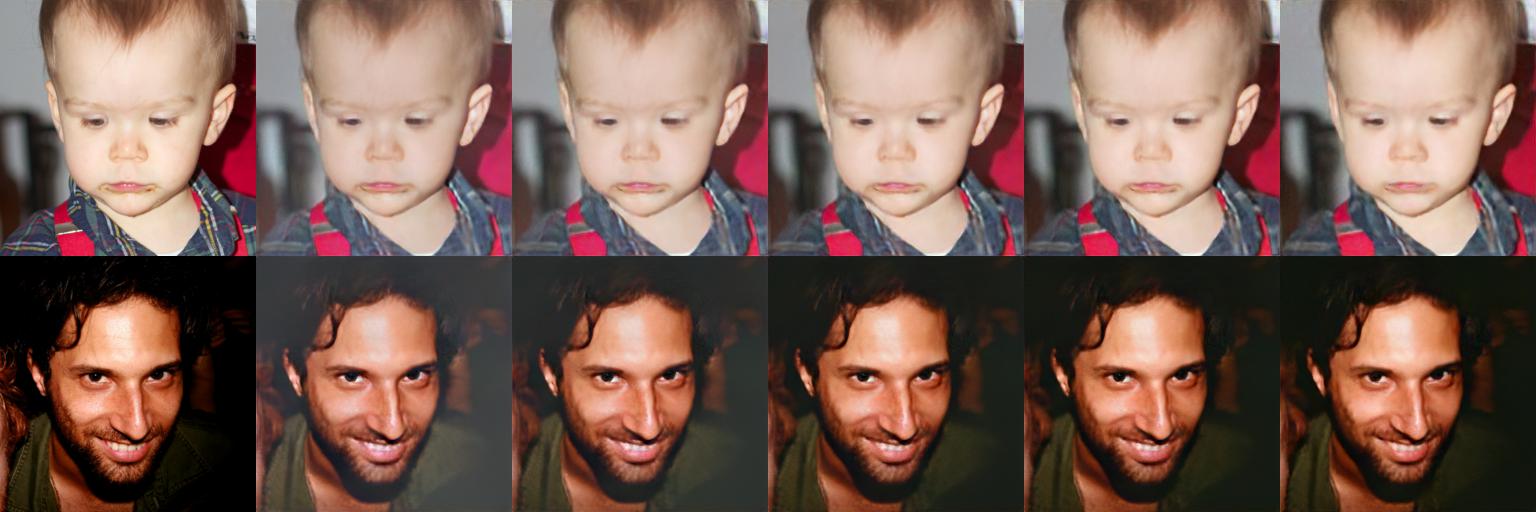} \styleganmsefigure{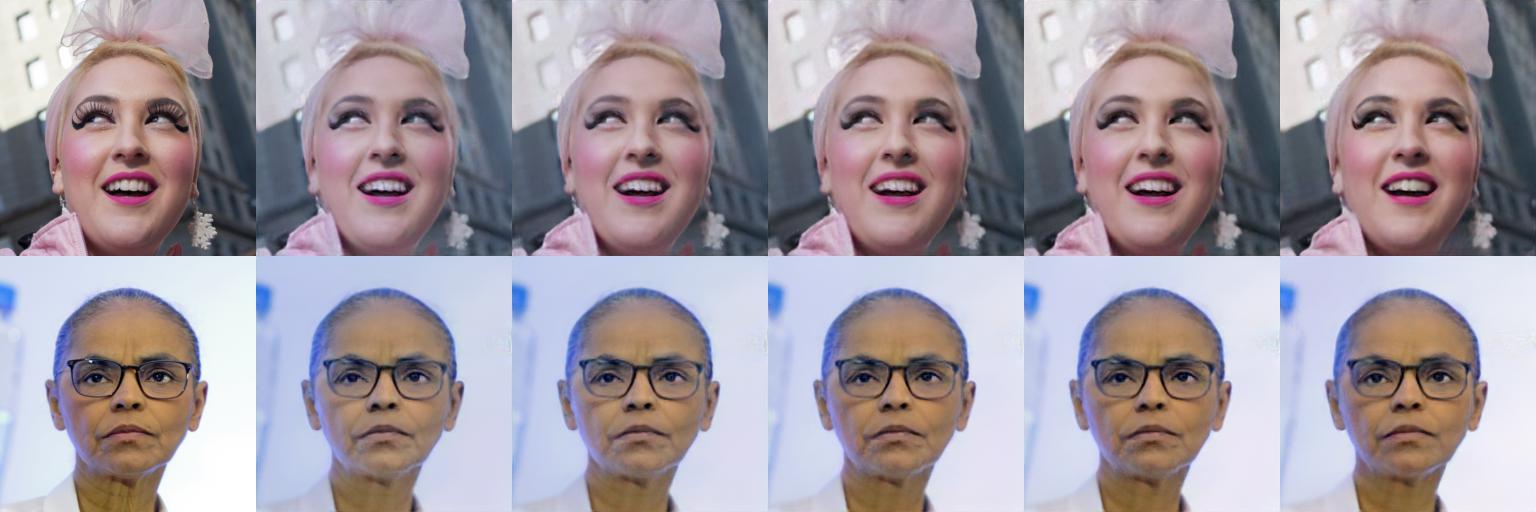} \styleganmsecaption \end{figure*}
\begin{figure*}[p]\ContinuedFloat \centering \styleganmsefigure{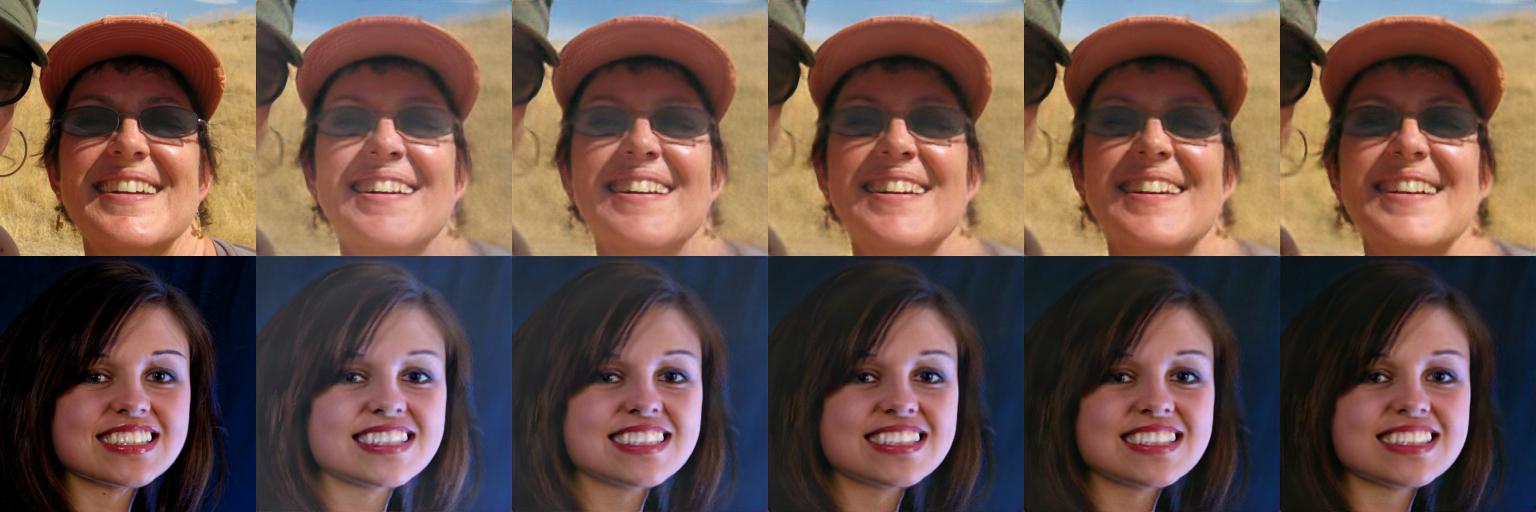} \styleganmsefigure{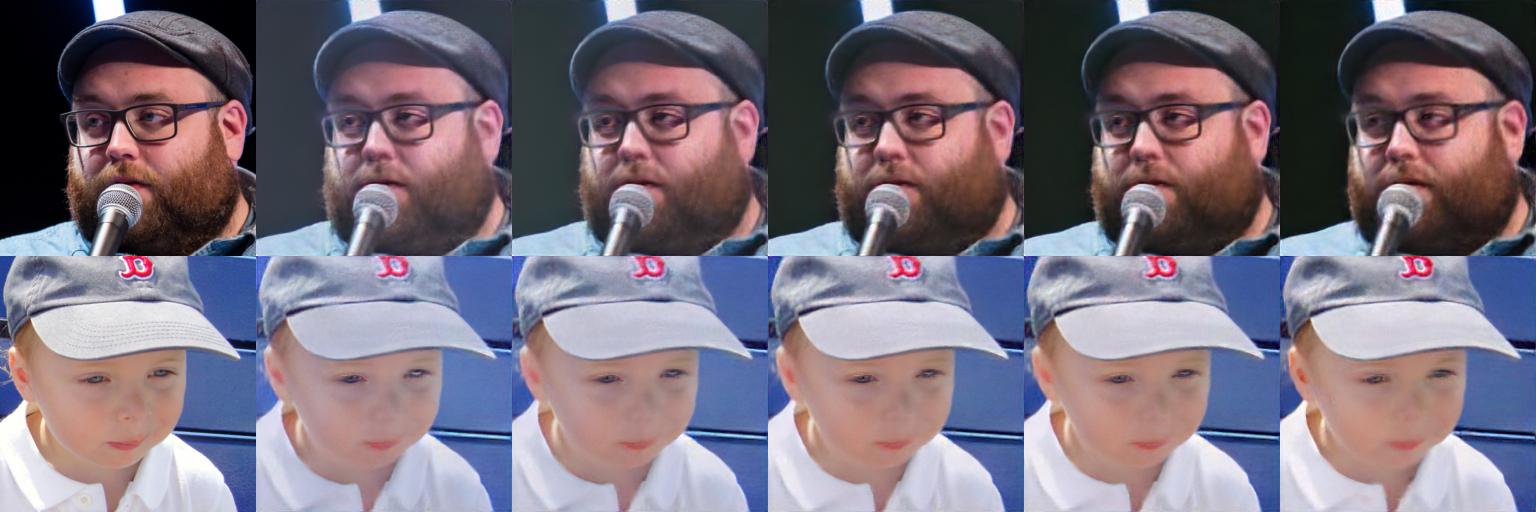} \styleganmsefigure{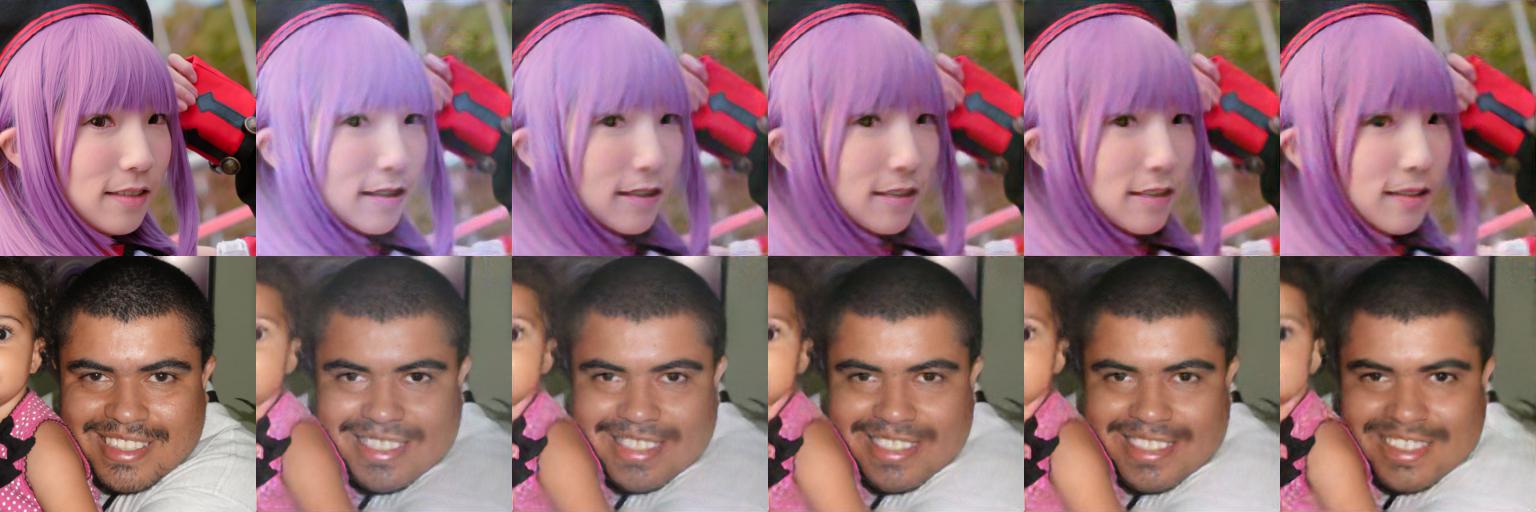} \styleganmsefigure{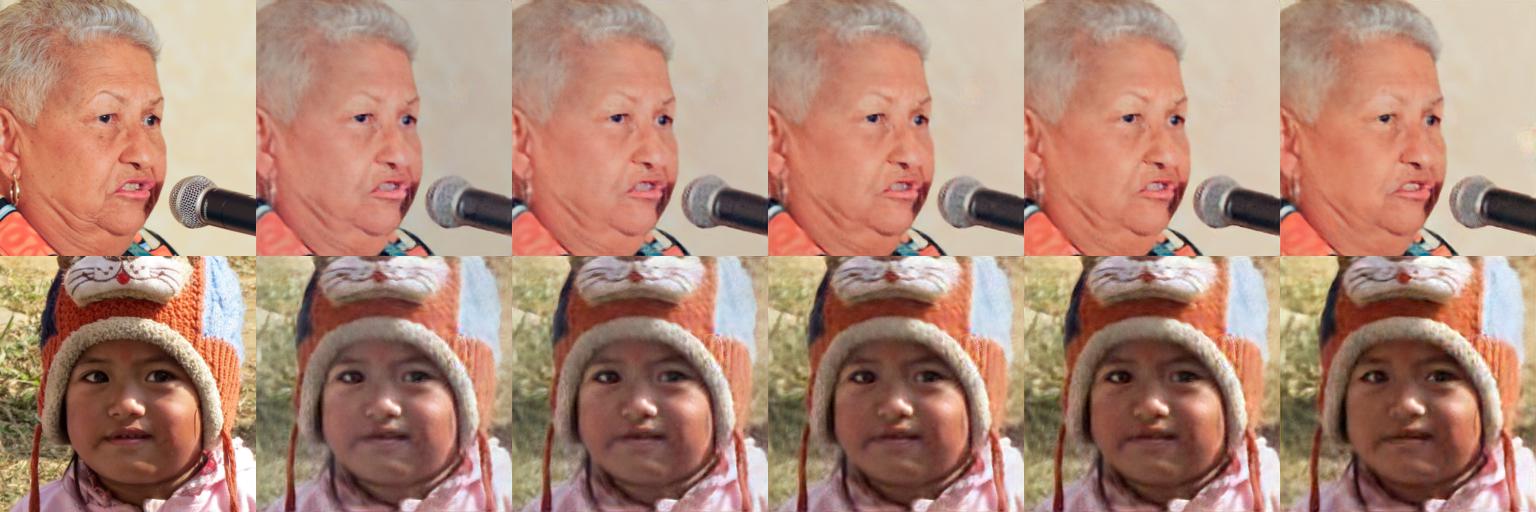} \styleganmsefigure{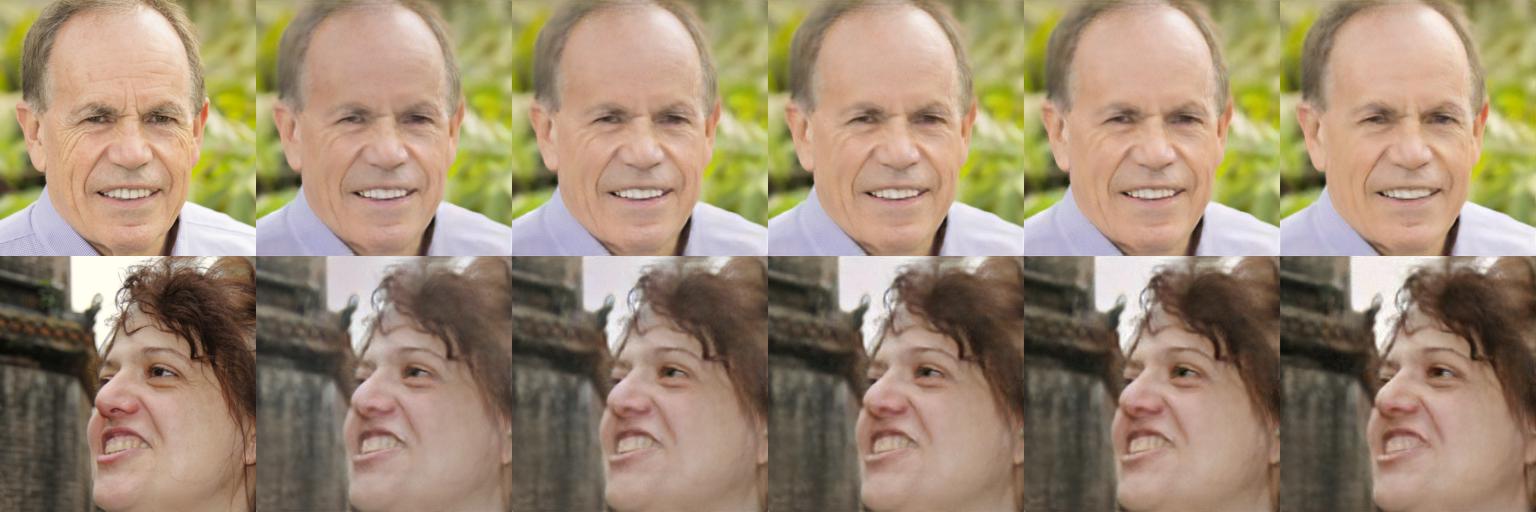} \styleganmsecaption \end{figure*}

\end{document}